%% file: main.tex
\documentclass[10pt,twocolumn,letterpaper]{article}

\usepackage[pagenumbers]{cvpr} % To force page numbers, e.g. for an arXiv version

\usepackage{booktabs} % For formal tables

\usepackage[ruled]{algorithm2e} % For algorithms
\usepackage{soul} 

\usepackage{commath} 
\usepackage{float}
\usepackage{pifont} % for ding

\usepackage{multirow}

\SetAlFnt{\small}
\SetAlCapFnt{\small}
\SetAlCapNameFnt{\small}
\SetAlCapHSkip{0pt}
\usepackage{amsmath}

\usepackage[pagebackref,breaklinks,colorlinks]{hyperref}

\newcommand{\shortcite}[1]{\cite{#1}}

\input{macros}

\makeatletter
\apptocmd\@maketitle{{%
\input{resources/figures/teaser2}
{}\par}}{}{}
\makeatother

\begin{document}

% Title portion
\title{
Stitch it in Time: GAN-Based Facial Editing of Real Videos}

\newcommand\myfigure{%
\input{resources/figures/teaser2}
}

\author{Rotem Tzaban \and Ron Mokady \and Rinon Gal \and Amit H. Bermano \and Daniel Cohen-Or \and \\ The Blavatnik School of Computer Science, Tel Aviv University}

\maketitle

\begin{abstract}

\input{abstract}
\end{abstract}

% \begin{teaserfigure}
%     \centering
%     \input{resources/figures/teaser}
%     \label{fig:teaser}
% \end{teaserfigure}

\input{intro}

\input{related}

\input{method}

\input{results}

\input{conclusion}

\newpage

\bibliographystyle{ieee_fullname}
\bibliography{egbib}

\end{document}

%% file: macros.tex
\usepackage{comment}
\usepackage{multirow,bigdelim}
\usepackage{lipsum}
\usepackage{array}
\usepackage{wrapfig}

\usepackage{adjustbox}
\usepackage[percent]{overpic}
\usepackage{makecell}
\usepackage[normalem]{ulem}

\usepackage{blindtext}

\usepackage{xcolor}
\usepackage{soul}

\newcolumntype{C}[1]{>{\centering\let\newline\\\arraybackslash\hspace{0pt}}m{#1}}

\newif\ifdraft
\drafttrue
\ifdraft

\definecolor{darkg}{rgb}{0,0.4,0}
\definecolor{lgreen}{rgb}{0,0.6,0}
\definecolor{amethyst}{rgb}{0.6, 0.4, 0.8}
\definecolor{orange}{rgb}{0.93,0.48,0.03}
\definecolor{blueviolet}{rgb}{0.54,0.16,0.88}
\definecolor{purple}{rgb}{0.5,0,0.5}

\newcommand{\w}{$\mathcal{W}$\xspace}
\newcommand{\wplus}{$\mathcal{W}+$\xspace}

\newcommand{\dcc}[1]{{\color{red}[\textbf{DC:} #1]}}
\newcommand{\rgc}[1]{{\color{purple}[\textbf{RG:} #1]}}
\newcommand{\rmc}[1]{{\color{darkg}[\textbf{RM:} #1]}}
\newcommand{\rzc}[1]{{\color{yellow}[\textbf{RZ:} #1]}}

\else
\newcommand{\dcc}[1]{}
\newcommand{\rgc}[1]{}
\newcommand{\rmc}[1]{}
\newcommand{\rzc}[1]{}

\fi

\def\naive{na\"{\i}ve\xspace}

\makeatletter
\DeclareRobustCommand\onedot{\futurelet\@let@token\@onedot}
\def\@onedot{\ifx\@let@token.\else.\null\fi\xspace}

\def\eg{\emph{e.g}\onedot}

\def\ie{\emph{i.e}\onedot}

\def\etal{\emph{et al}\onedot}
\makeatother
\usepackage[bottom]{footmisc}
\makeatletter
\def\blfootnote{\xdef\@thefnmark{}\@footnotetext}
\makeatother

\usepackage[capitalize, sort]{cleveref}
\crefname{section}{Sec.}{Secs.}
\Crefname{section}{Section}{Sections}
\Crefname{table}{Table}{Tables}
\crefname{table}{Tab.}{Tabs.}

%% file: resources/figures/teaser2.tex
    \vspace{-0.3615cm}
    \centering
    \setlength{\tabcolsep}{0.5pt}
    {
    \begin{tabular}{ccccccccccc}
        
        \vspace{-0.0615cm}
        
        % Identity 1

        \raisebox{0.12in}{\rotatebox{90}{Original}} &
        \includegraphics[width=0.095\textwidth]{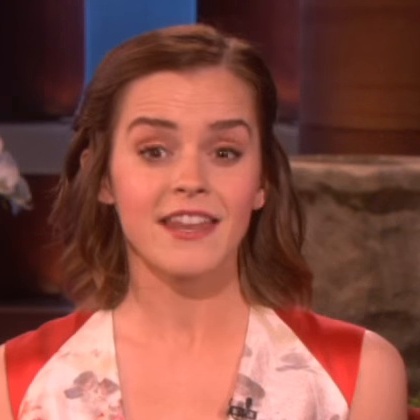} &
        \includegraphics[width=0.095\textwidth]{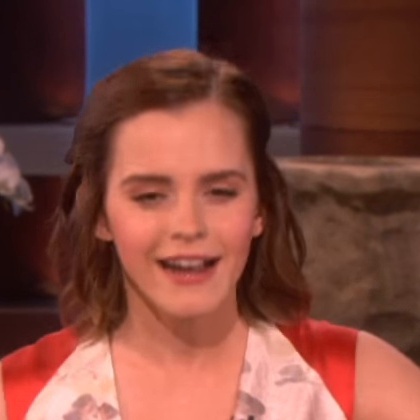} &
        \includegraphics[width=0.095\textwidth]{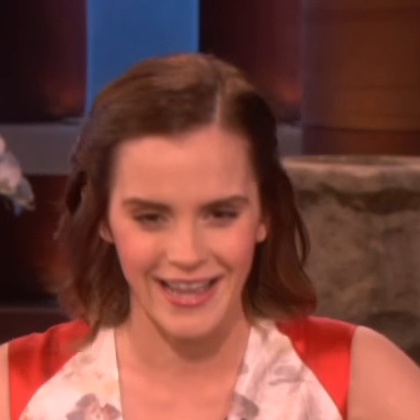} &
        \includegraphics[width=0.095\textwidth]{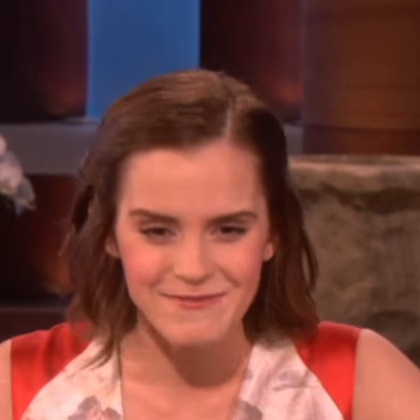} &
        \includegraphics[width=0.095\textwidth]{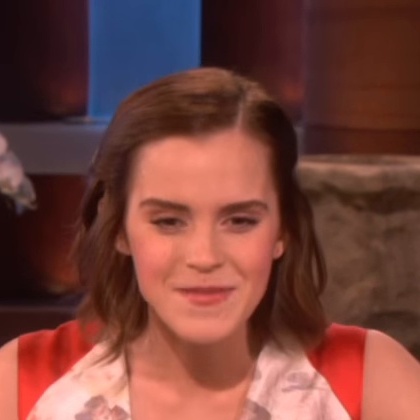} &
        \includegraphics[width=0.095\textwidth]{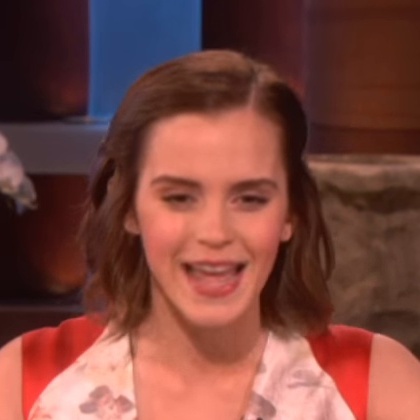} &
        \includegraphics[width=0.095\textwidth]{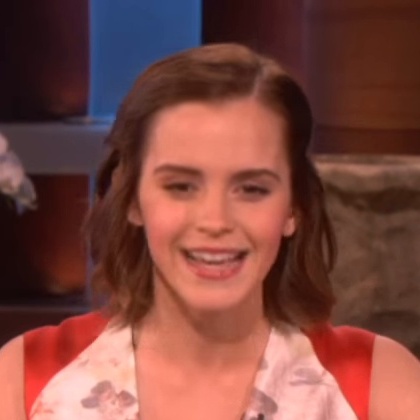} &
        \includegraphics[width=0.095\textwidth]{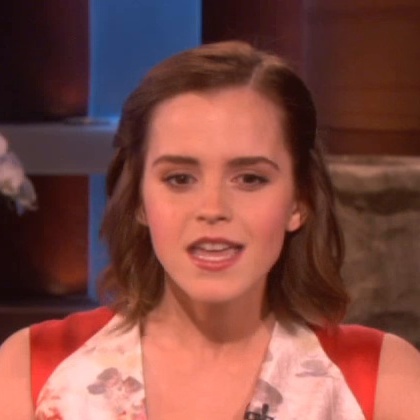} &
        \includegraphics[width=0.095\textwidth]{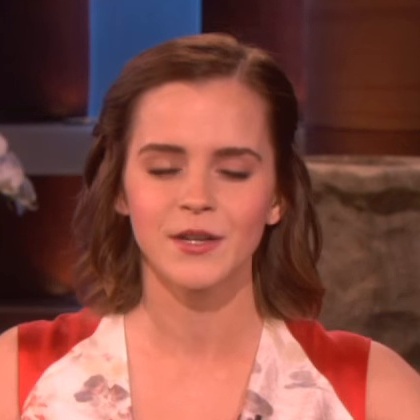} &
        \includegraphics[width=0.095\textwidth]{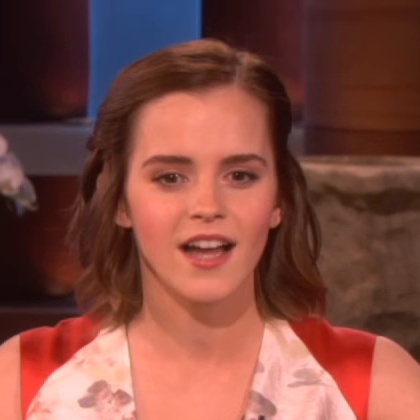} \\
        
        \raisebox{0.12in}{\rotatebox{90}{+Young}} &
        \includegraphics[width=0.095\textwidth]{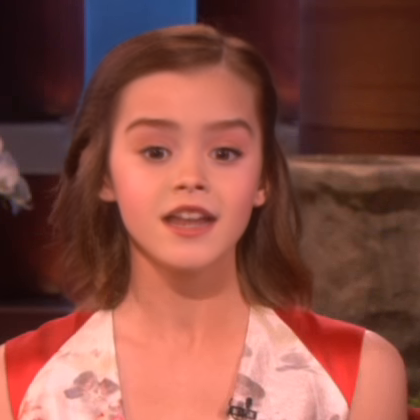} &
        \includegraphics[width=0.095\textwidth]{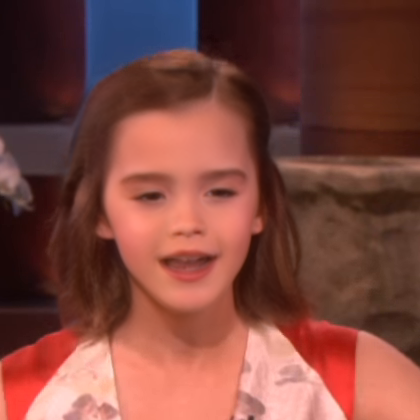} &
        \includegraphics[width=0.095\textwidth]{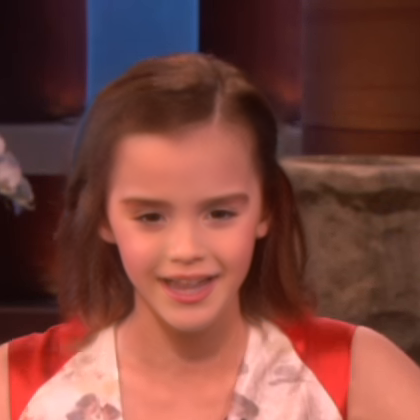} &
        \includegraphics[width=0.095\textwidth]{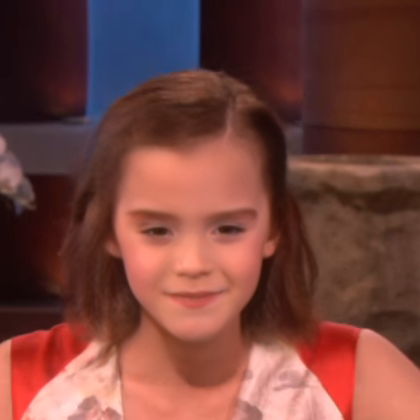} &
        \includegraphics[width=0.095\textwidth]{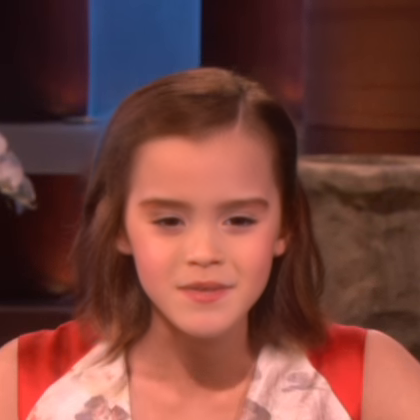} &
        \includegraphics[width=0.095\textwidth]{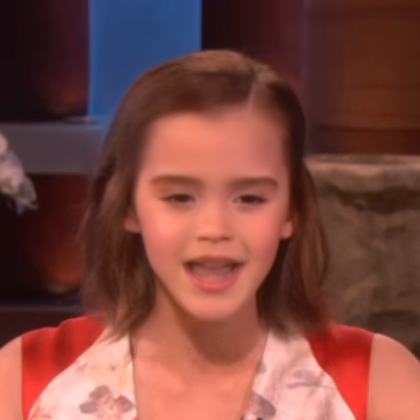} &
        \includegraphics[width=0.095\textwidth]{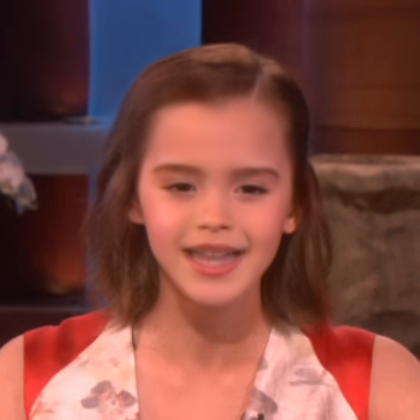} &
        \includegraphics[width=0.095\textwidth]{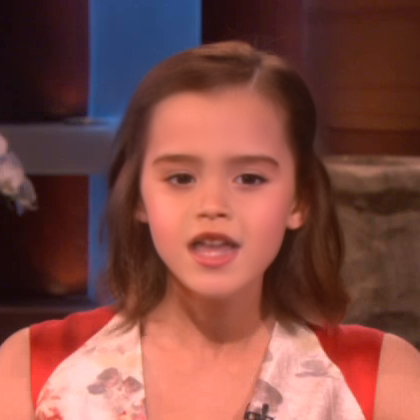} &
        \includegraphics[width=0.095\textwidth]{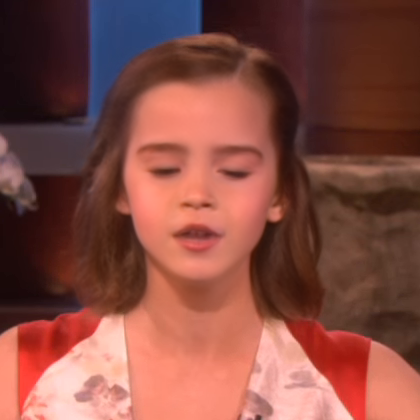} &
        \includegraphics[width=0.095\textwidth]{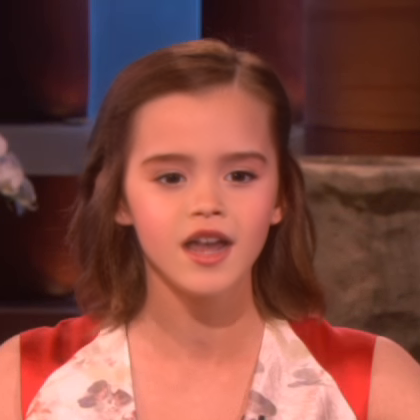} \\

        \raisebox{0.09in}{\rotatebox{90}{+Angry}} &
        \includegraphics[width=0.095\textwidth]{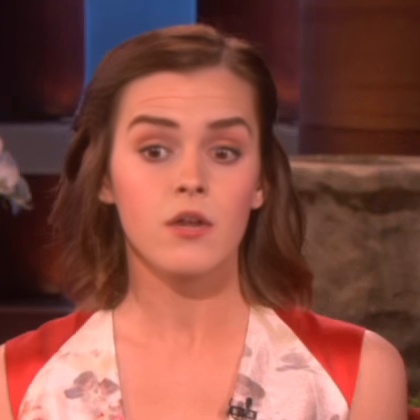} &
        \includegraphics[width=0.095\textwidth]{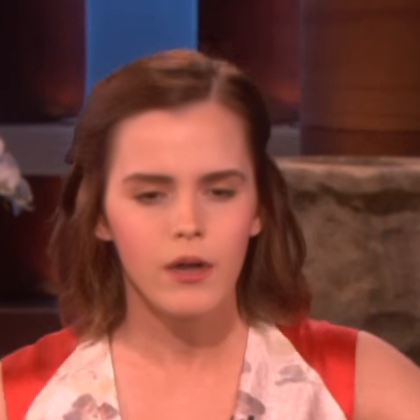} &
        \includegraphics[width=0.095\textwidth]{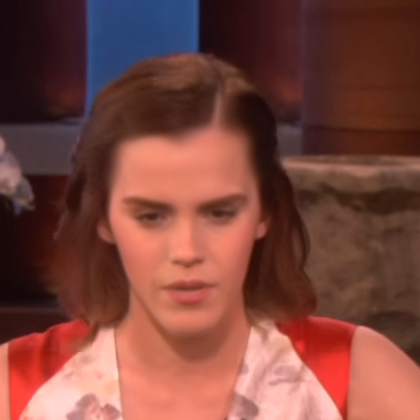} &
        \includegraphics[width=0.095\textwidth]{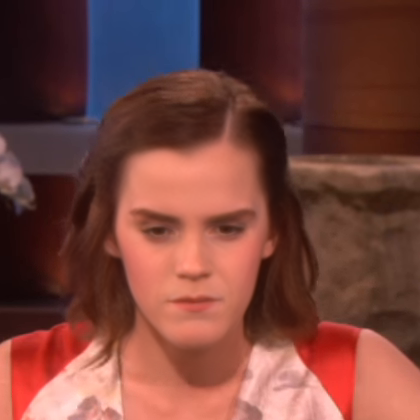} &
        \includegraphics[width=0.095\textwidth]{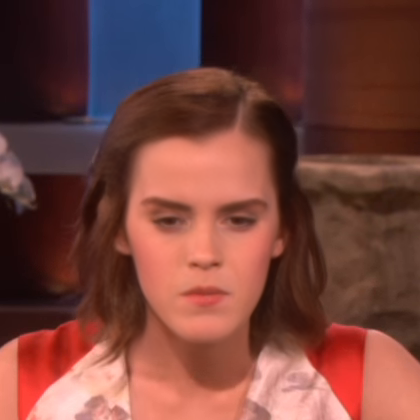} &
        \includegraphics[width=0.095\textwidth]{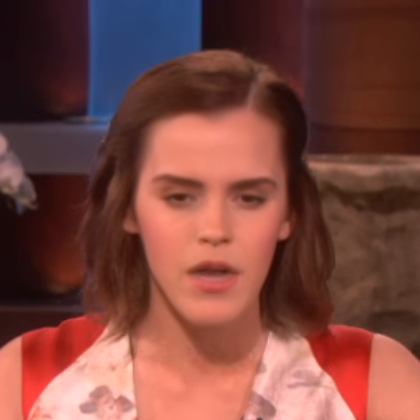} &
        \includegraphics[width=0.095\textwidth]{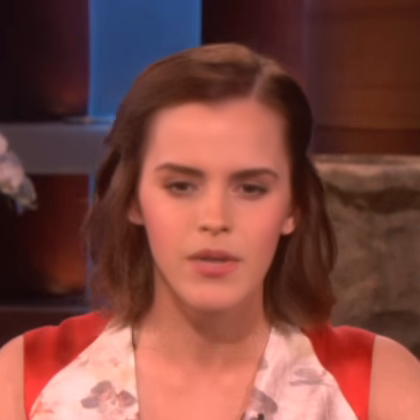} &
        \includegraphics[width=0.095\textwidth]{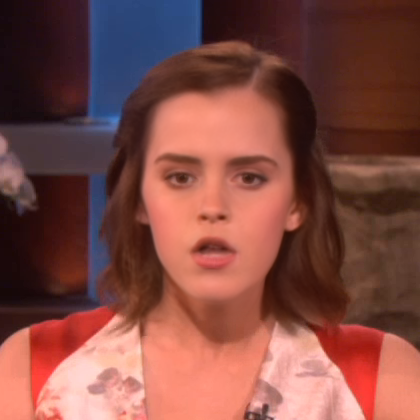} &
        \includegraphics[width=0.095\textwidth]{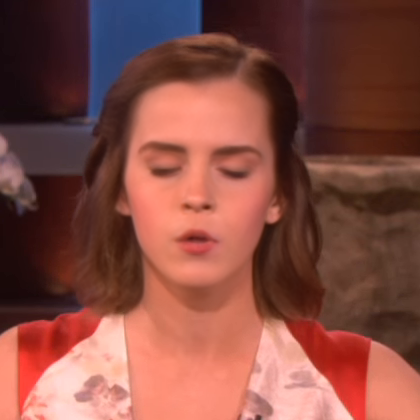} &
        \includegraphics[width=0.095\textwidth]{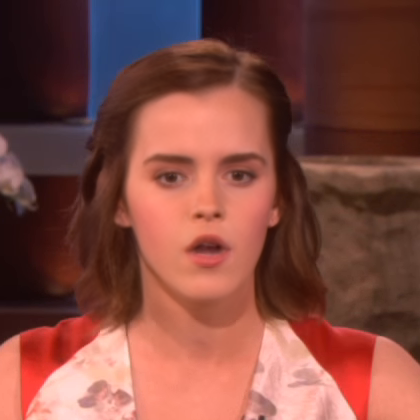} \\

        % \\

        %& Frame 1 & Frame 2 & Frame 3 & Frame 4 & Frame 1 & Frame 2 & Frame 3 & Frame 4
        
    \end{tabular}
    
    }
    % \vspace{-0.225cm}
     \captionof{figure}{Video editing using our proposed pipeline. Our framework can successfully apply consistent semantic manipulations to challenging talking-head videos, without requiring any temporal components or losses.}
    \vspace{0.325cm}
    \label{fig:teaser}

%% file: abstract.tex
% The ability of Generative Adversarial Networks (GANs) to encode rich semantics within their latent space has been widely adopted for facial image editing. However, replicating their success with videos has proven challenging. Training video GANs is difficult, owing to a lack of high quality datasets. More crucially, videos introduce another dimension of realism that needs to be maintained - temporal consistency.
% When editing videos, the source is already coherent, and so the challenge in maintaining temporal realism is largely \textit{virtual}, arising in part due to careless treatment of individual components in the editing pipeline. We propose to leverage the natural alignment of StyleGAN-based models and the tendency of neural networks to learn low frequency functions, and demonstrate that they already provide a strongly consistent prior. We draw on these insights and propose a framework for semantic editing of faces in videos, demonstrating significant improvements over the current state-of-the-art, \ron{producing meaningful editing while retaining the temporal consistency. Unlike current works, we utilize challenging high quality talking heads videos with their original frame rate, showing our method applicable for arbitrary real videos. Code and videos will be published.}

The ability of Generative Adversarial Networks to encode rich semantics within their latent space has been widely adopted for facial image editing. However, replicating their success with videos has proven challenging. Sets of high-quality facial videos are lacking, and working with videos introduces a fundamental barrier to overcome --- \textit{temporal coherency}. We propose that this barrier is largely artificial. The source video is already temporally coherent, and deviations from this state arise in part due to careless treatment of individual components in the editing pipeline. We leverage the natural alignment of StyleGAN and the tendency of neural networks to learn low frequency functions, and demonstrate that they provide a strongly consistent prior. We draw on these insights and propose a framework for semantic editing of faces in videos, demonstrating significant improvements over the current state-of-the-art. Our method produces meaningful face manipulations, maintains a higher degree of temporal consistency, and can be applied to challenging, high quality, talking head videos which current methods struggle with. Our code and results are available at \url{https://stitch-time.github.io/}

%% file: intro.tex
\section{Introduction}

The advent of Generative Adversarial Networks (GANs) has brought with it a renaissance in the field of content creation and manipulation, allowing users to modify photographs in intuitive ways. In particular, the highly disentangled latent space of StyleGAN \cite{karras2019style,karras2020analyzing} has been widely adapted for realistic editing of facial images. However, these semantic editing tools have been mostly restricted to images, as the editing of videos imposes an additional challenge -- maintaining \textit{temporal coherency}. Any manipulation of the video must be propagated consistently across all video frames. Prior work suggests tackling this challenge by training a GAN for video synthesis~\cite{skorokhodov2021stylegan,yan2021videogpt,yu2022generating}. However, with a lack of high quality video datasets and the complications arising from an additional data dimension, video-GANs have so far been unable to match the quality of their single-image counterparts. 

Instead, we propose to meet this challenge by using the latent-editing techniques commonly employed with an off-the-shelf, non-temporal StyleGAN model.
%, and without additional supervision.
We highlight a fundamental assumption about the video editing process: the initial video is already consistent. In contrast to synthesis works, we do not need to \textit{create} temporal consistency, but only \textit{maintain} it. Building on this intuition, we revisit the building blocks of recent StyleGAN-based editing pipelines, identify the points where temporal inconsistencies may arise, and propose that in many cases these inconsistencies can be mitigated simply through a careful choice of tools. 
%\ron{By exploiting the natural smoothness of neural networks, we successfully perform temporally consistent semantic editing.}

We begin our investigation by identifying two types of temporal inconsistencies: local -- where the transitions between adjacent frames are not smooth and display considerable jitter, and global -- where inaccuracies in the GAN editing process, such as changes in identity, build up over time.
We consider the recently proposed PTI~\cite{roich2021pivotal}; A two-step approach to inversion which first finds a 'pivot' -- an initial latent code that can be fed through the generator to produce an approximation of the input image. Then, the generator's weights are fine-tuned so that the specific 'pivot' code can better reproduce the target. PTI provides strong global consistency, keeping the identity aligned with the target video. However, our investigation reveals that it fares poorly on the local benchmark and produces inversions which behave inconsistently under editing operations.

At this point, we make two key observations: The generator is a highly parametric neural function, which are known to be predisposed to learning low frequency functions. As such, a small change in its inputs (the latent codes) is likely to induce only a minor variation in the generated images. Moreover, it has been shown that style-based models maintain incredible alignment under fine-tuning, particularly when transitioning to nearby domains \cite{wu2021stylealign,pinkney2020resolution,gal2021stylegannada}. As such, if the generator produces consistent editing for a set of smoothly changing latent codes - we expect any fine-tuned generator to be similarly predisposed towards temporal consistency. 

With these intuitions in mind, we propose that PTI's local inconsistency arises at the first step of the process -- finding the 'pivots'. More specifically, the employed optimization-based inversion is inconsistent. Highly similar frames can be encoded into different regions of the latent space, even when using the same initialization and random noise seed. On the other hand, encoder-based inversion methods utilize highly parametrized networks, and are therefore also biased to low-frequency representations. As such, an encoder is likely to provide slowly changing latents when its input only undergoes a minor change - such as when observing two adjacent video frames.

We merge the two approaches: an encoder for discovering locally consistent pivots, and generator fine-tuning to promote global consistency, and demonstrate that they already provide a strongly consistent prior.
Nevertheless, they are not sufficient for editing real videos. As StyleGAN cannot operate over the entire frame, we need to stitch the edited crop back to the original video. However, inversion and editing methods typically corrupt the background extensively, making their results difficult to blend into the original frame. For this purpose, we design a novel `stitching-tuning' operation that further tunes the generator to provide spatially-consistent transitions. By doing so, we achieve realistic blending while retaining the editing effects.

We demonstrate that our proposed editing pipeline can seamlessly apply latent-based semantic modifications to faces in real videos. Although we employ only non-temporal models, we can successfully edit challenging talking head videos with considerable movement and complex backgrounds, which current methods fail to tackle. In \cref{fig:teaser}, we show several frames extracted from a video edited using our method. These demonstrate our ability to alter in-the-wild scenes and maintain temporal coherence. Through a detailed ablation study, we validate each of the suggested components and demonstrate its contribution to both realism and consistency. All videos are available as part of our supplementary materials.

\input{resources/figures/pipeline}
\input{resources/figures/pipeline_example}

%% file: resources/figures/pipeline.tex
\vspace{-0.1cm}
\begin{figure*}[tb]
    \centering
    \setlength{\belowcaptionskip}{-5pt}
    \includegraphics[width=0.97\linewidth]{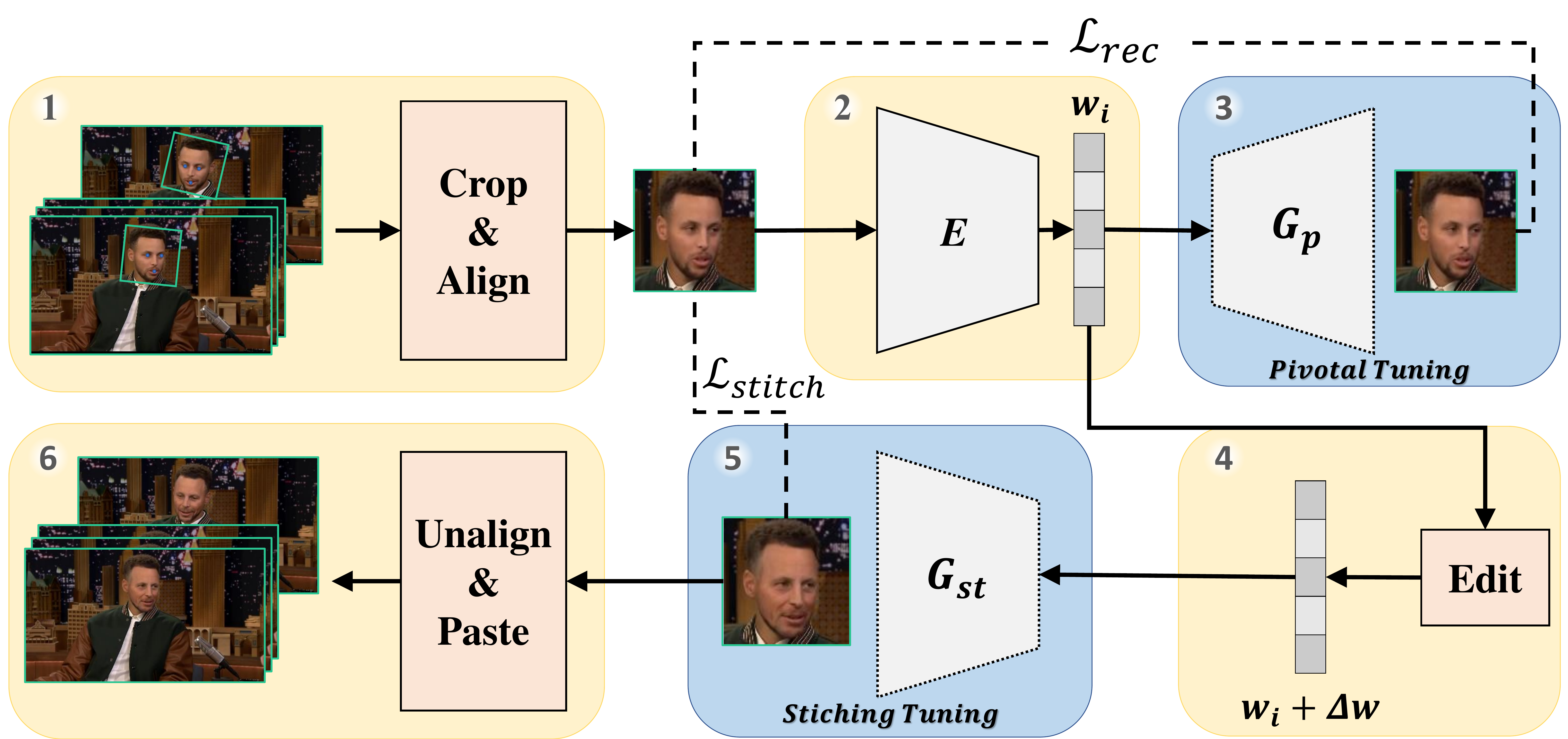}
    % \vspace{-0.25cm}
    \caption{
    Our full video editing pipeline contains $6$ steps. $(1)$ Videos are split into individual frames. The face in each frame is cropped and aligned. $(2)$ Each cropped face is inverted into the latent space of a pre-trained StyleGAN2 model, using a pre-trained e4e encoder. $(3)$ The generator is fine-tuned using PTI across all video frames in parallel, correcting for inaccuracies in the initial inversion and restoring global coherence. $(4)$ All frames are edited by manipulating their pivot latent codes linearly, using a fixed direction and step-size. $(5)$ We fine-tune the generator a second time, stitching the backgrounds and the edited faces together in a spatially-smooth manner. $(6)$ We reverse the alignment step and paste the modified face into the video.
    }
    % \vspace{-0.1cm}
    \label{fig:pipeline}
\end{figure*}

%% file: resources/figures/pipeline_example.tex
\begin{figure}
\setlength{\tabcolsep}{0.5pt}
    \centering
    { \small 
\begin{tabular}{ccccc}
Original & e4e Encoding & $+$PTI & $+$Editing & $+$Stitching \\
\raisebox{-.32\totalheight}{\includegraphics[width=0.2\columnwidth]{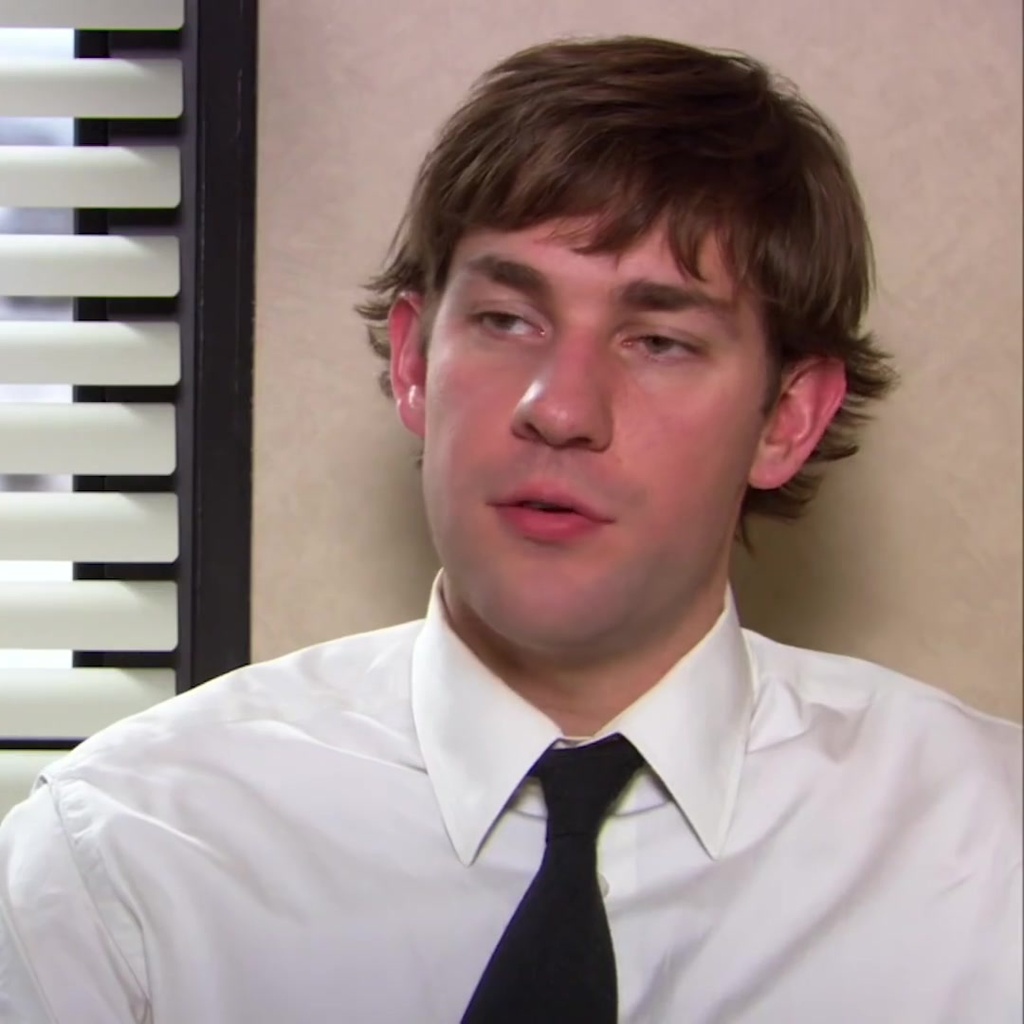}} &
\raisebox{-.32\totalheight}{\includegraphics[width=0.2\columnwidth]{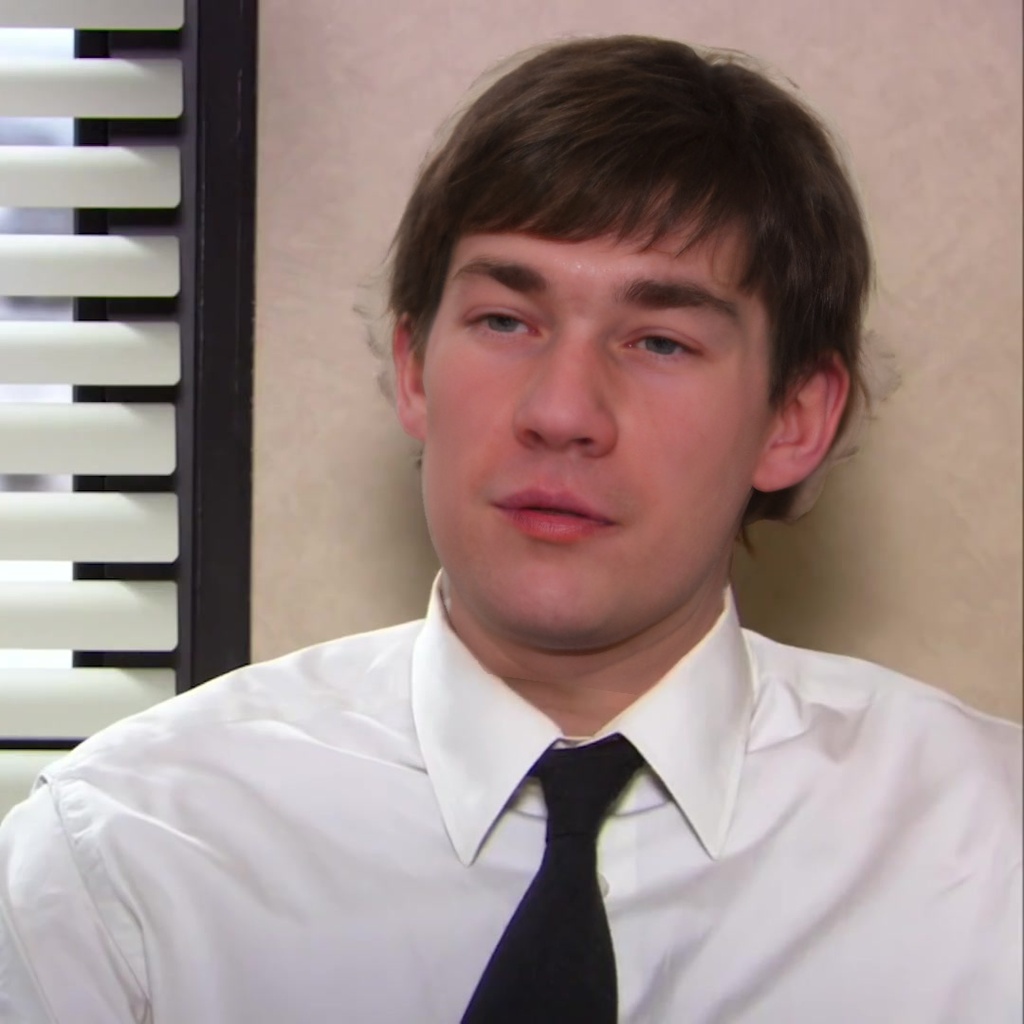}} &
\raisebox{-.32\totalheight}{\includegraphics[width=0.2\columnwidth]{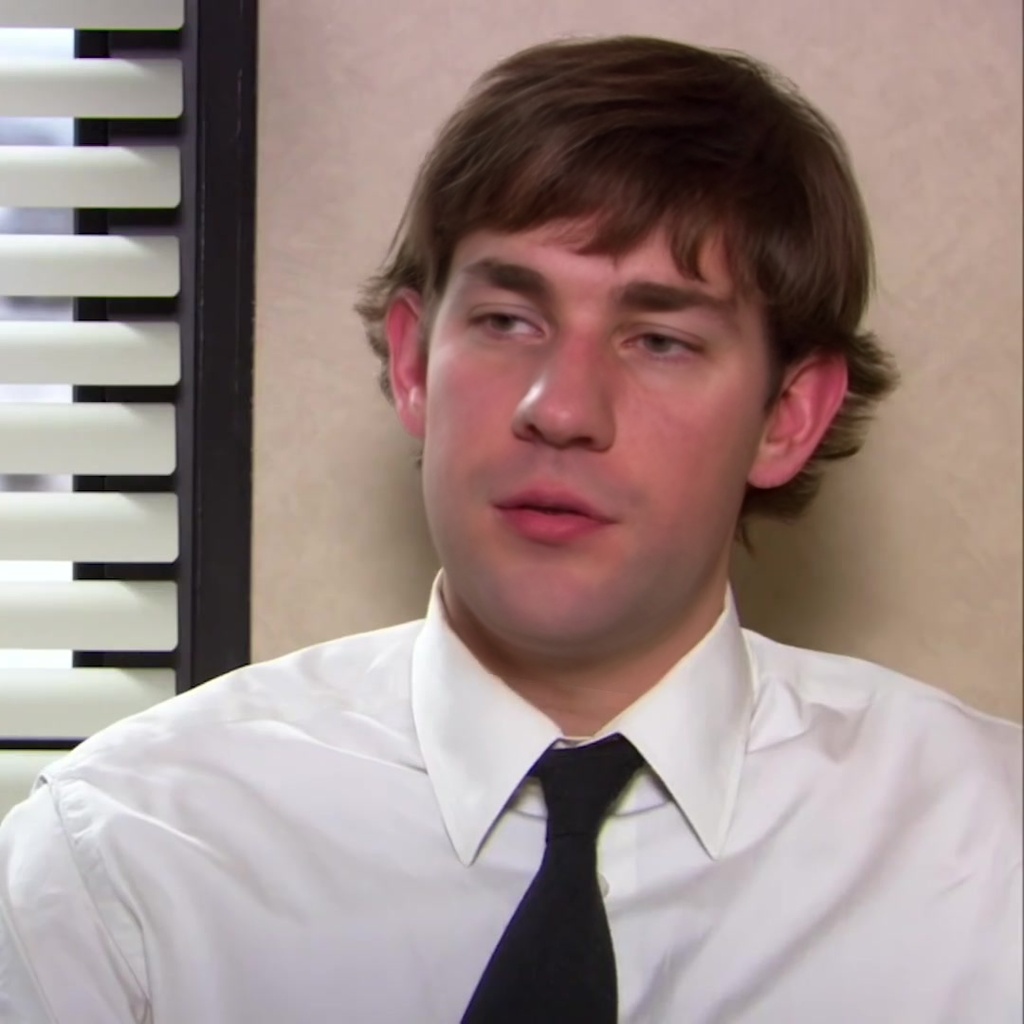}} &
\raisebox{-.32\totalheight}{\includegraphics[width=0.2\columnwidth]{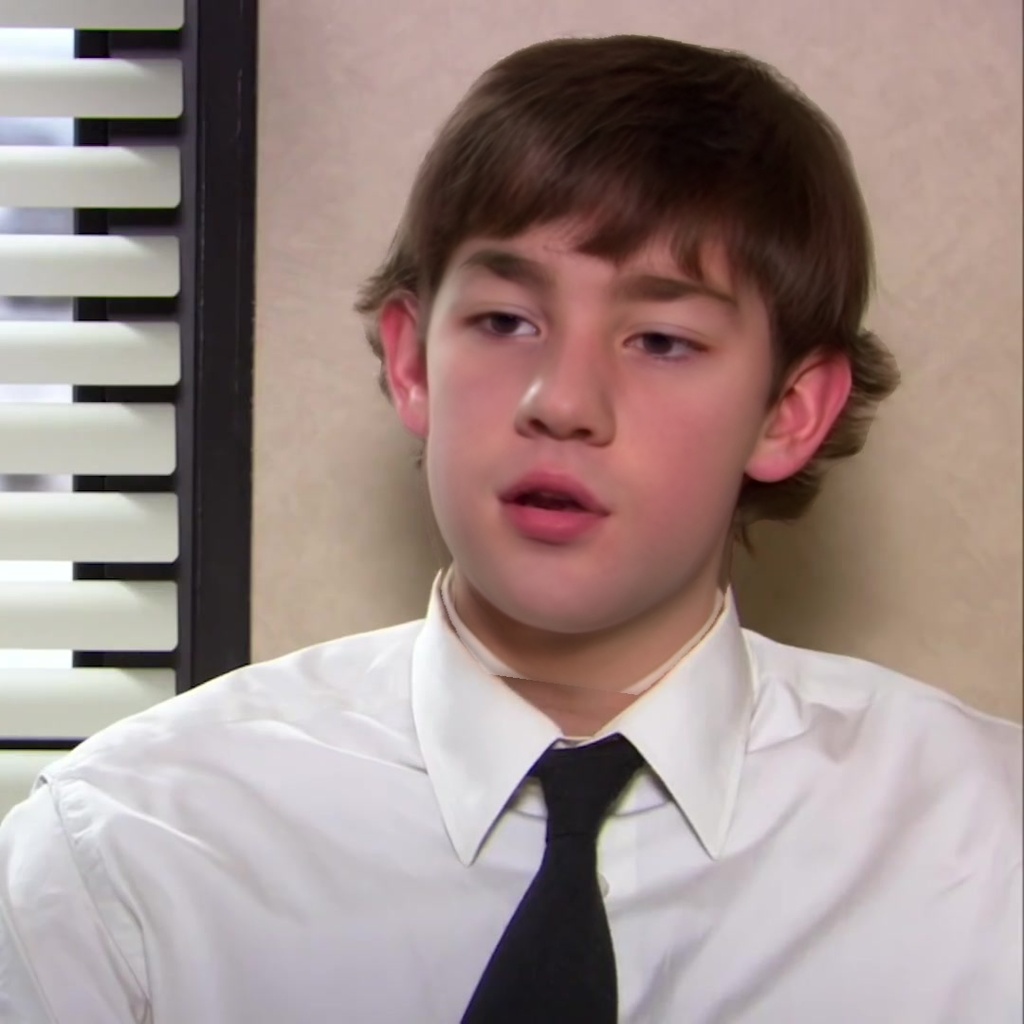}} &
\raisebox{-.32\totalheight}{\includegraphics[width=0.2\columnwidth]{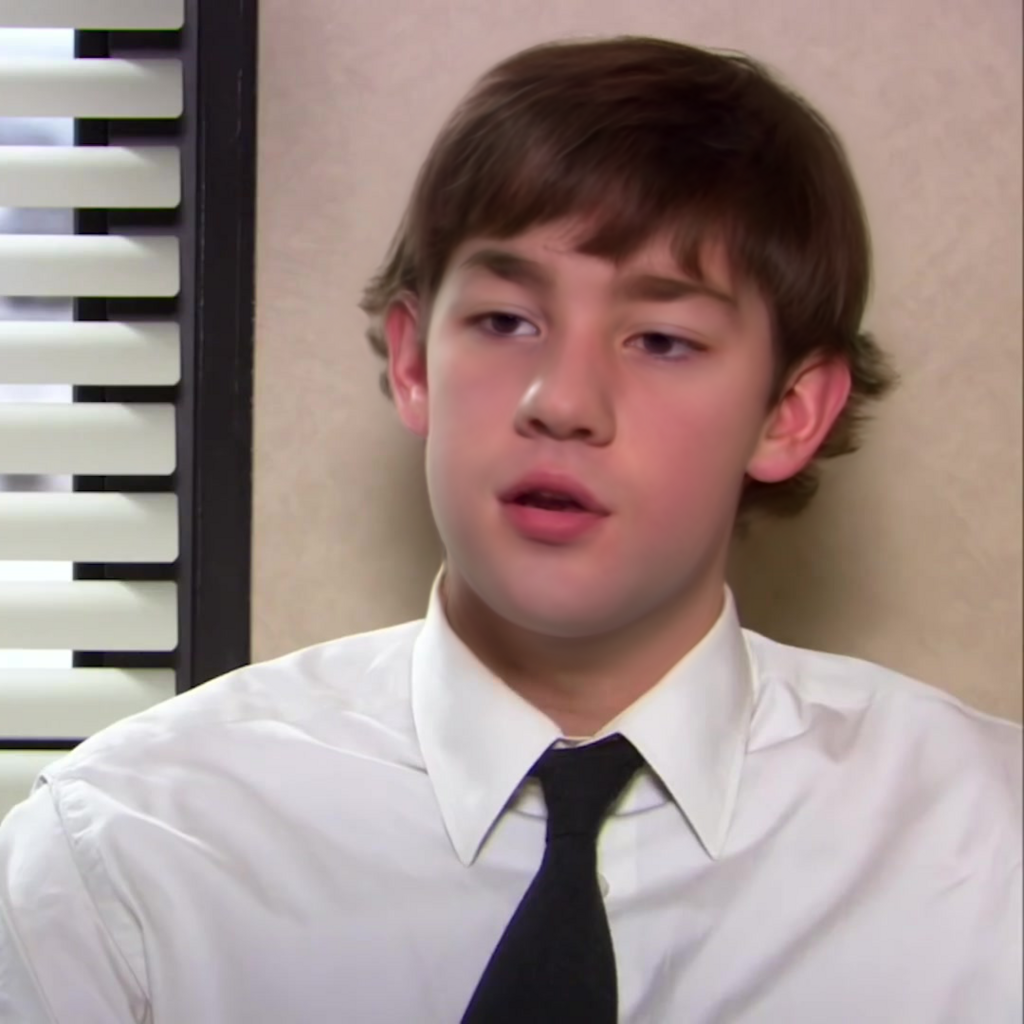}} \\

\raisebox{-.32\totalheight}{\includegraphics[width=0.2\columnwidth]{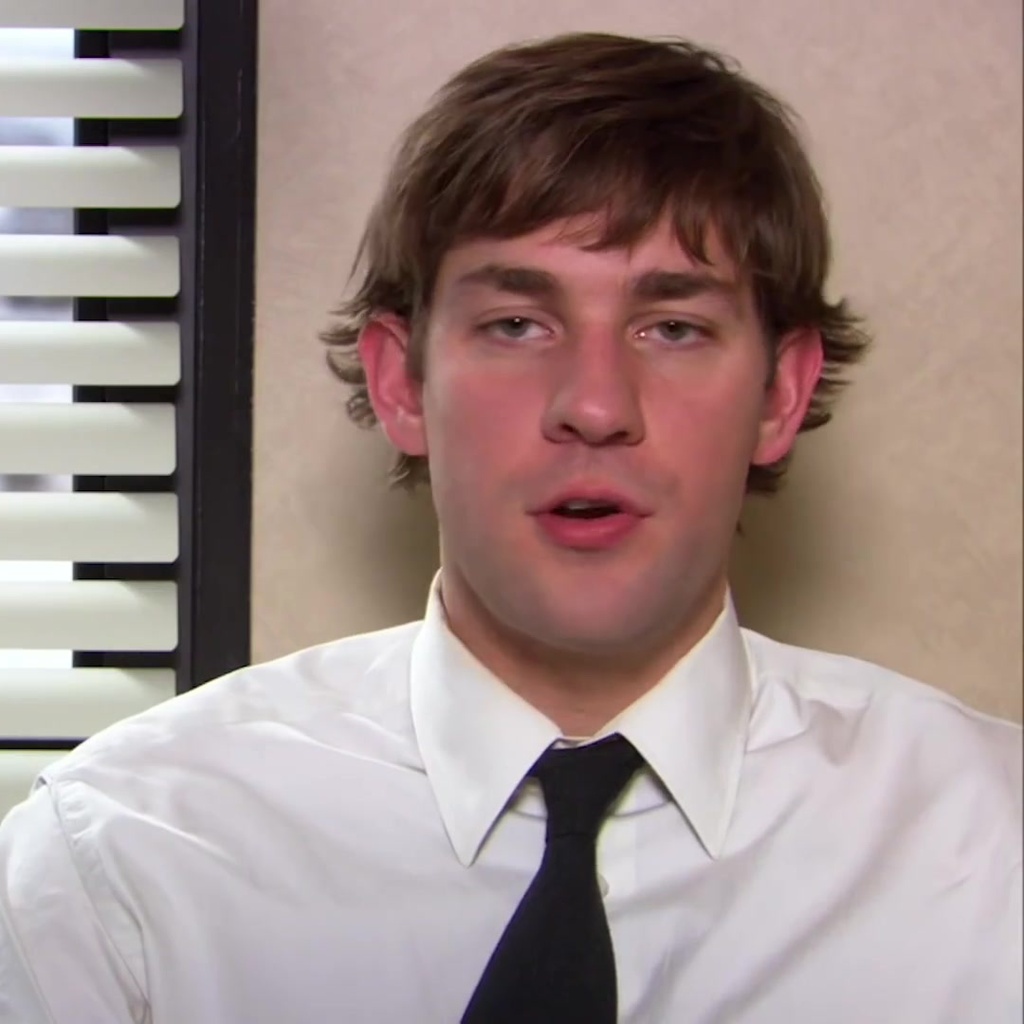}} &
\raisebox{-.32\totalheight}{\includegraphics[width=0.2\columnwidth]{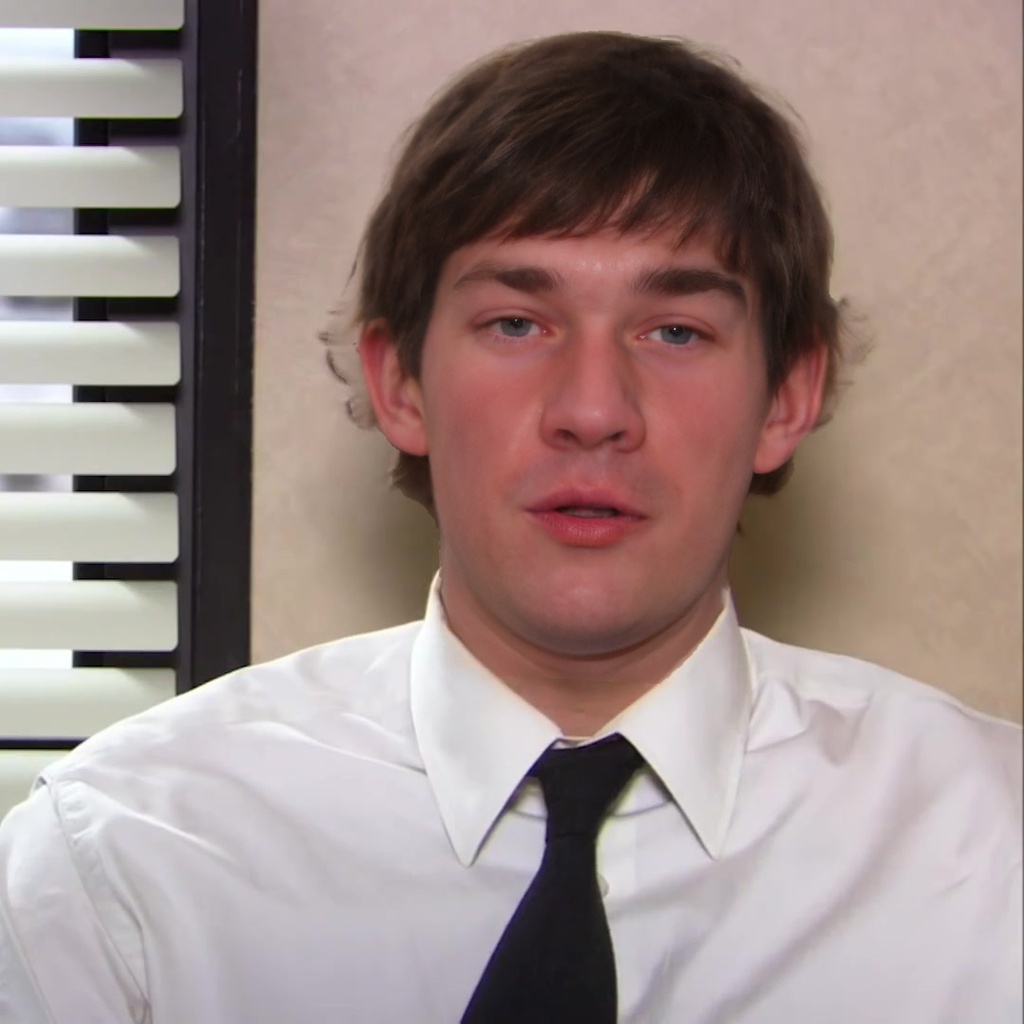}} &
\raisebox{-.32\totalheight}{\includegraphics[width=0.2\columnwidth]{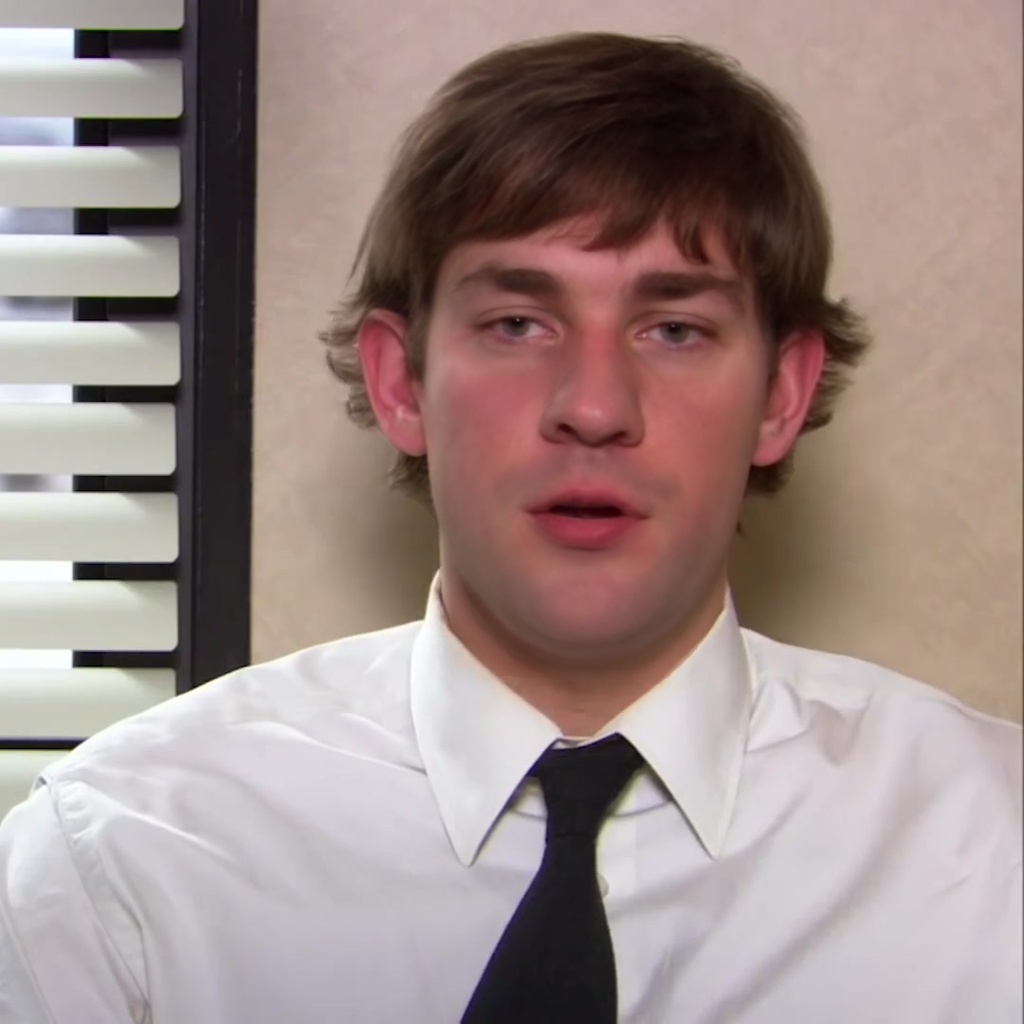}} &
\raisebox{-.32\totalheight}{\includegraphics[width=0.2\columnwidth]{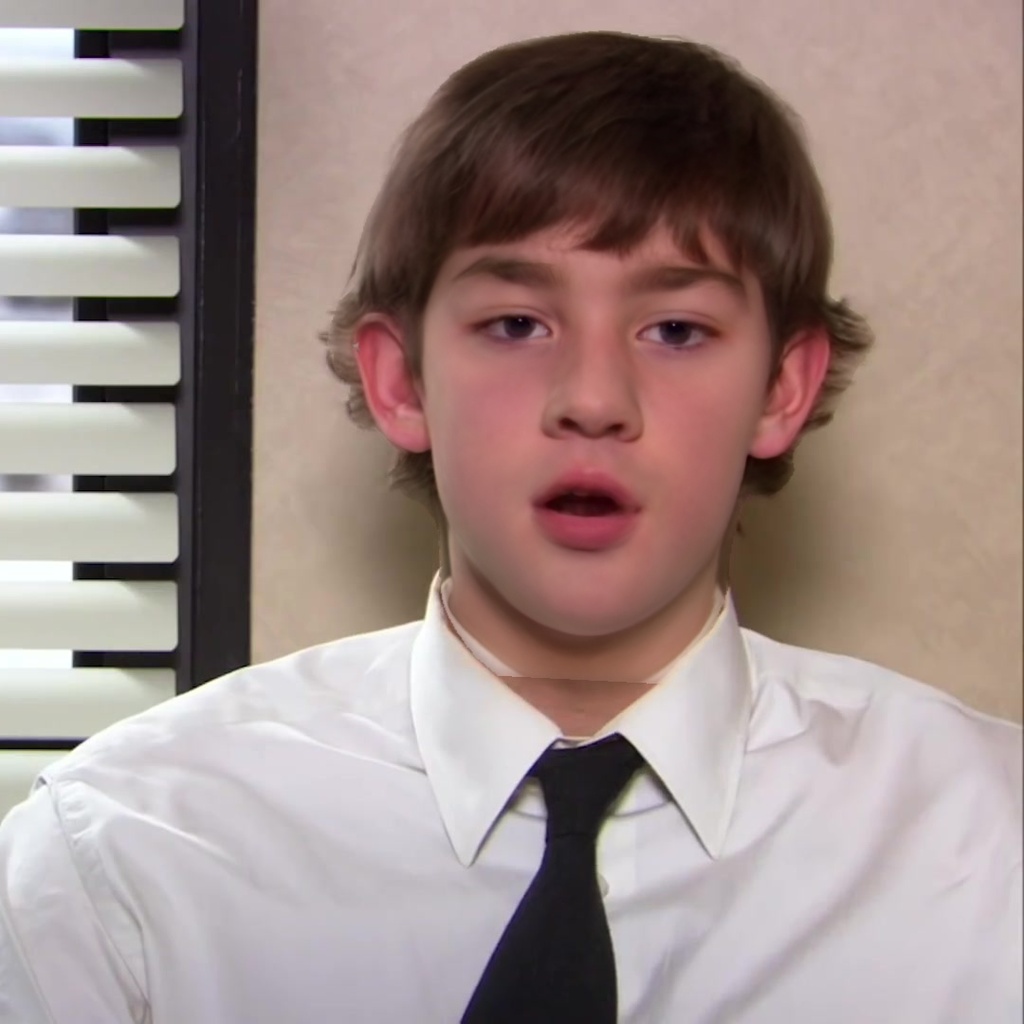}} &
\raisebox{-.32\totalheight}{\includegraphics[width=0.2\columnwidth]{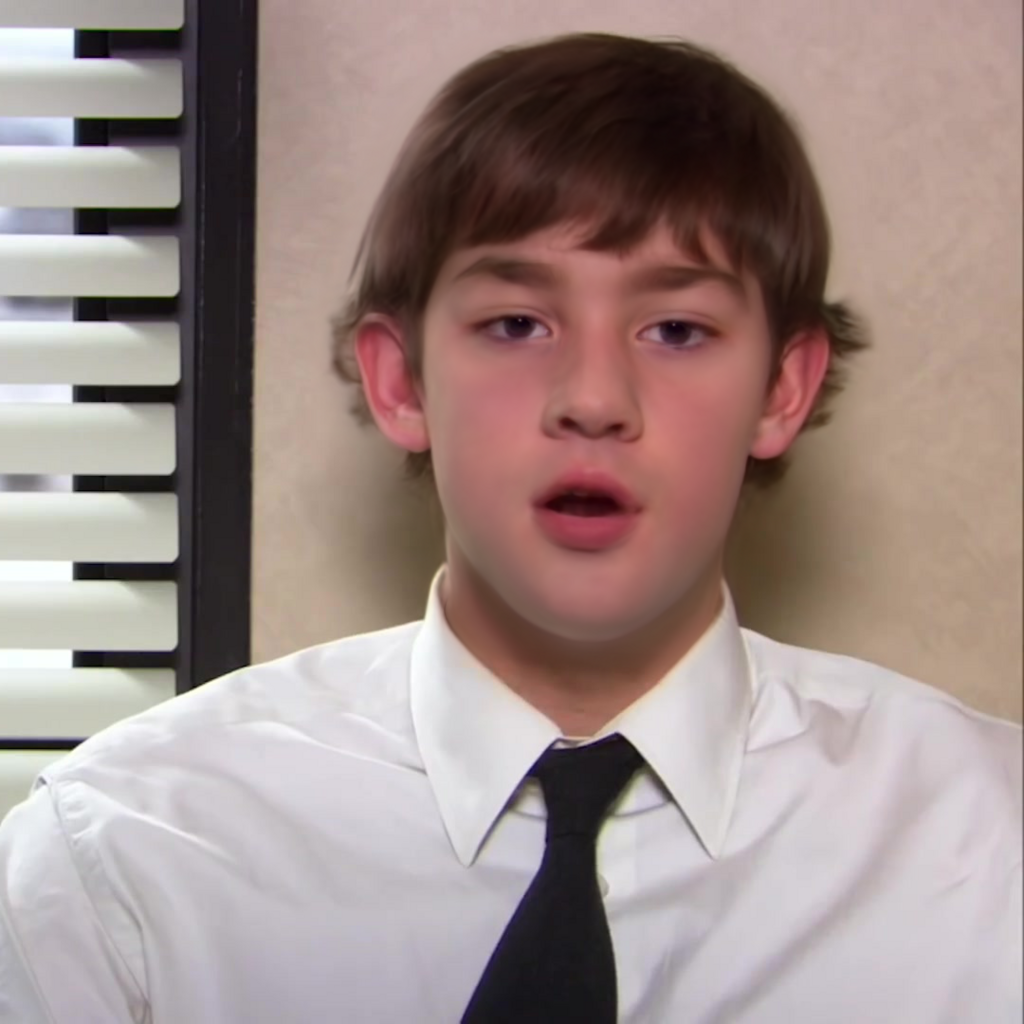}} \\

\raisebox{-.32\totalheight}{\includegraphics[width=0.2\columnwidth]{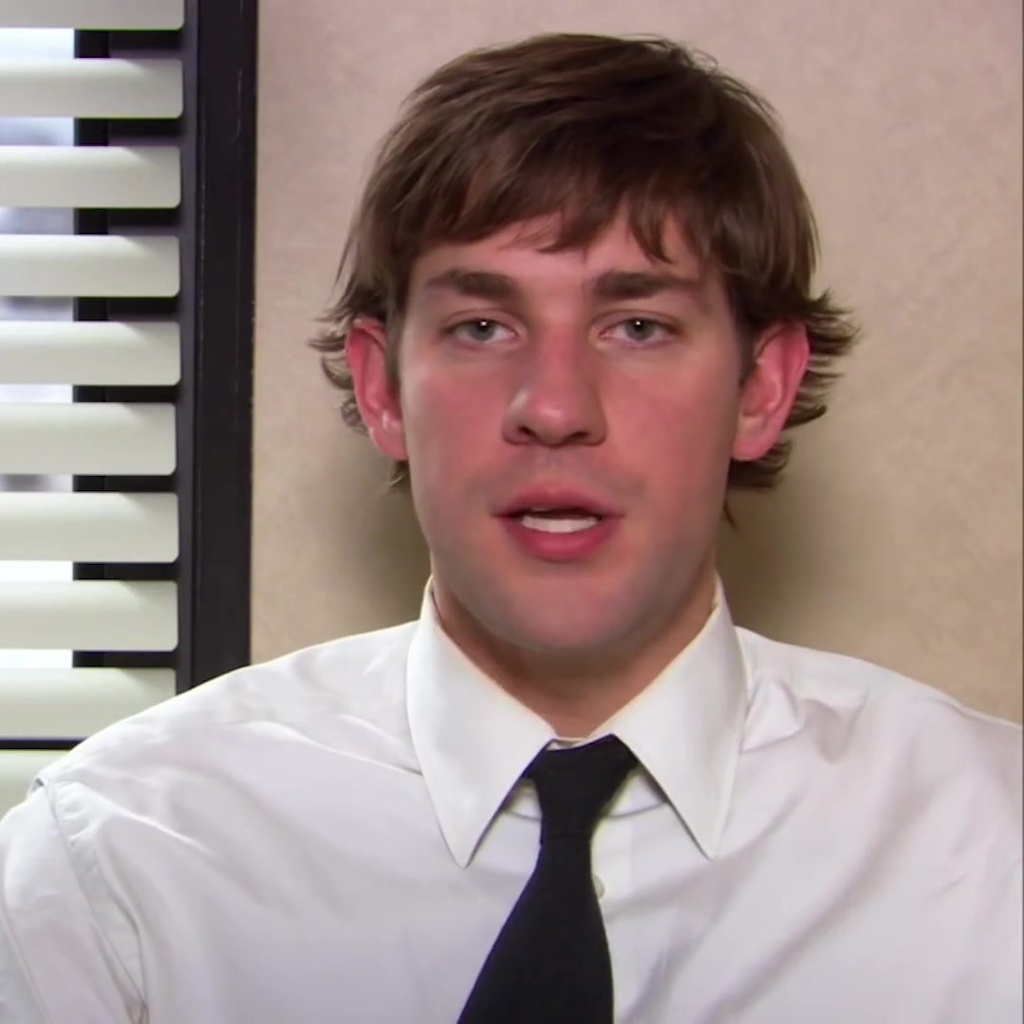}} &
\raisebox{-.32\totalheight}{\includegraphics[width=0.2\columnwidth]{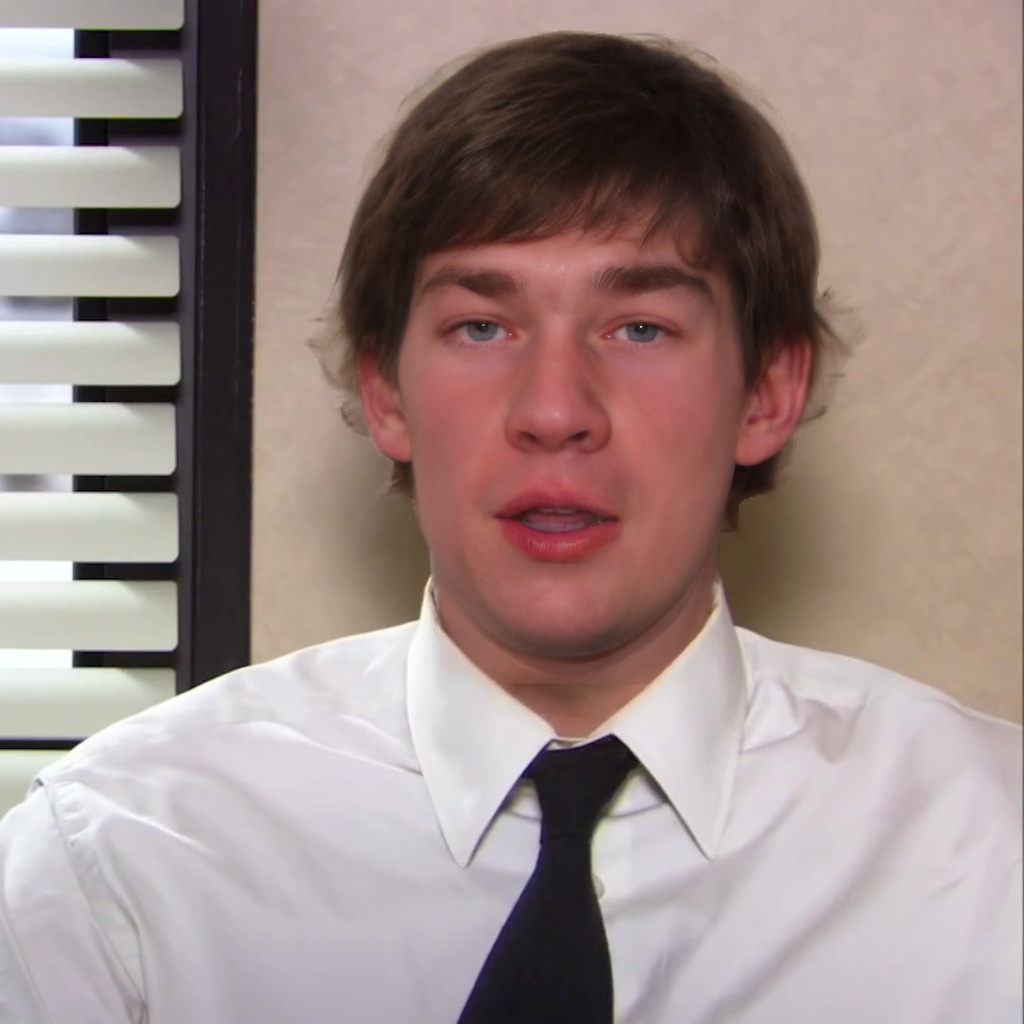}} &
\raisebox{-.32\totalheight}{\includegraphics[width=0.2\columnwidth]{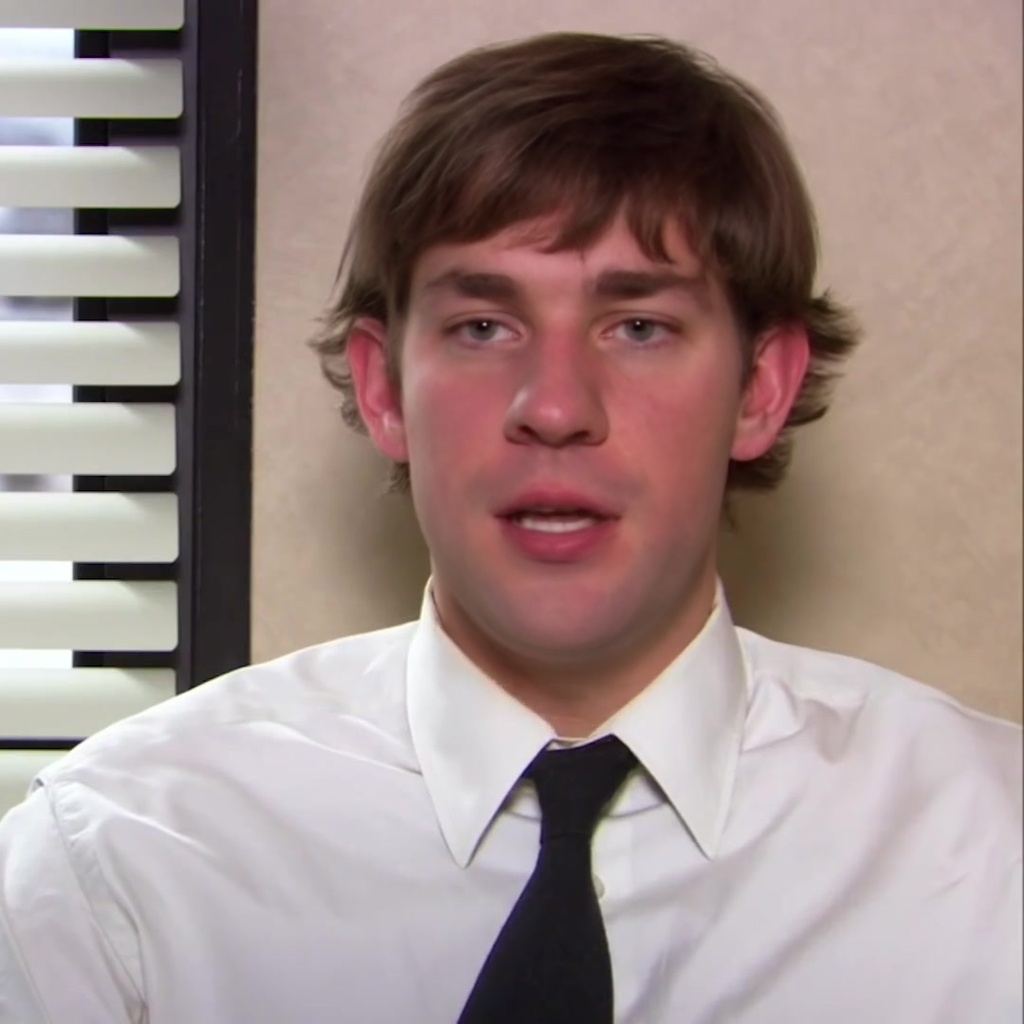}} &
\raisebox{-.32\totalheight}{\includegraphics[width=0.2\columnwidth]{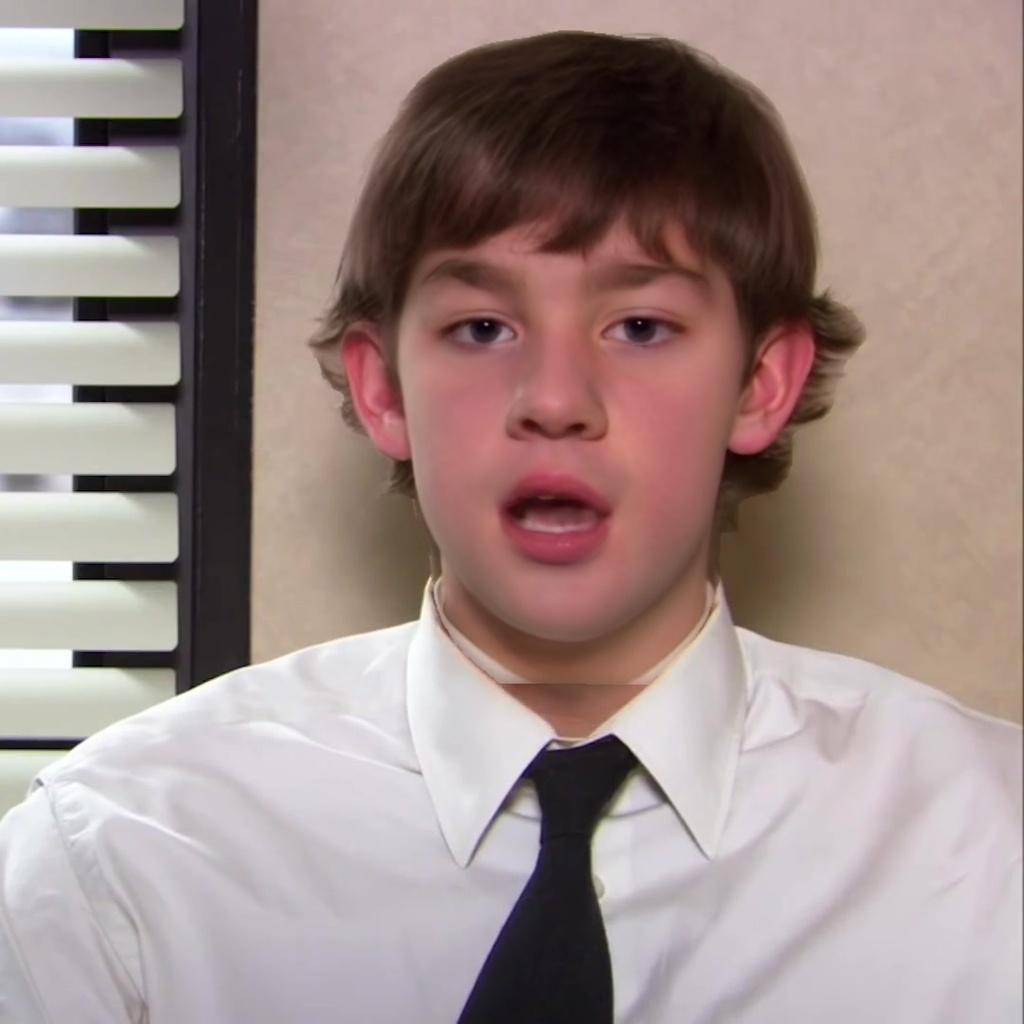}} &
\raisebox{-.32\totalheight}{\includegraphics[width=0.2\columnwidth]{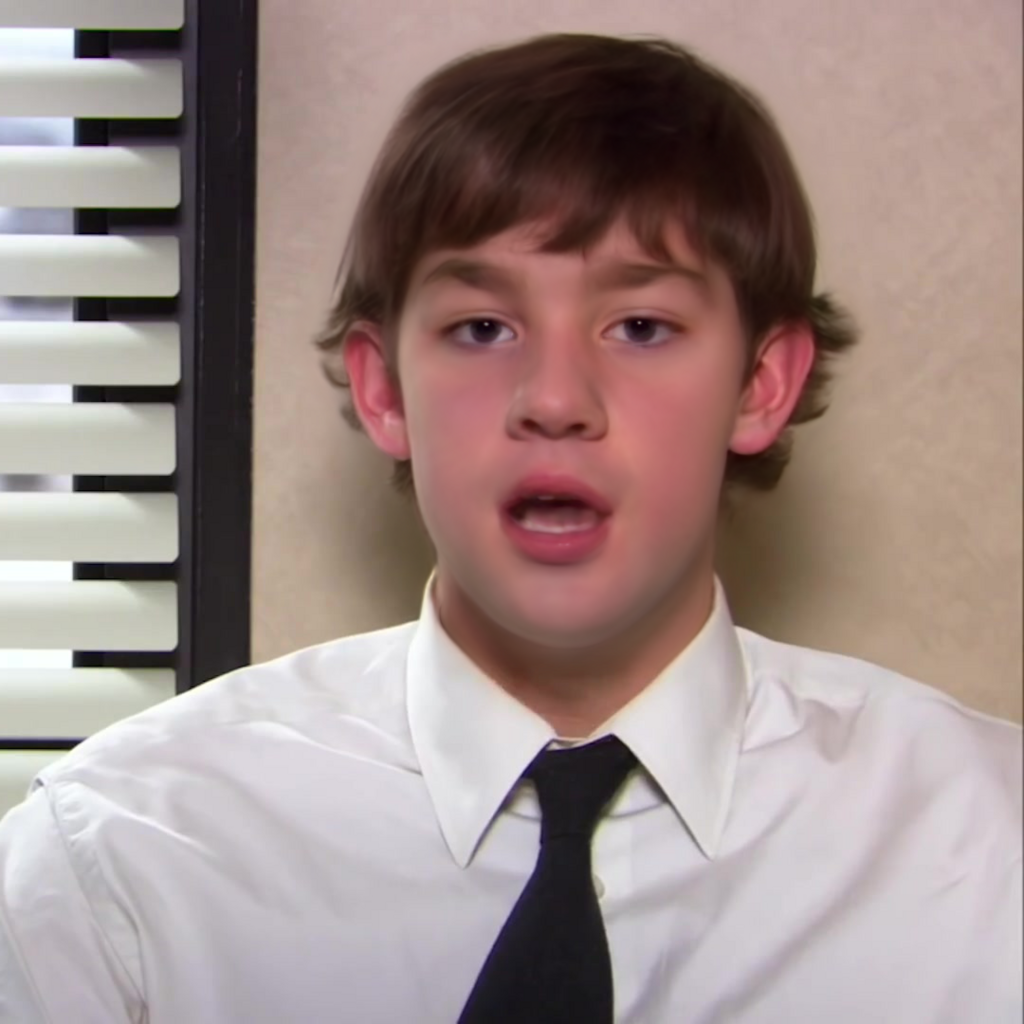}} \\

\end{tabular}
}

\caption{Visualization of our full editing pipeline. In the left column, we show three frames extracted from the source video. In the following columns, we show intermediate results of our pipeline over the same three frames. Left to right: the encoder-inversion step, the PTI fine-tuning step, the pivot editing step, and finally our stitching procedure. When not applying our stitching procedure, we use a segmentation-mask based blending procedure~\shortcite{yao2021latent}. Note in particular the neck region, which displays considerable artifacts after the editing step which are then eliminated through our stitching-tuning approach. }

\label{fig:pipeline_example}
\end{figure}

%% file: related.tex
\section{Background and Related Work}~\label{sec:rw}

\vspace{-12pt}

\paragraph{StyleGAN-based Editing}

StyleGAN~\cite{karras2019style, karras2020analyzing} employ a style-based architecture to generate high fidelity images from a semantically rich and highly structured latent space. Remarkably, StyleGAN enables realistic editing of images through simple latent code modifications.
Motivated by this, many methods have discovered meaningful latent directions, using various levels of supervision. These range from full-supervision~\cite{shen2020interpreting,abdal2020styleflow,denton2019detecting, goetschalckx2019ganalyze, tewari2020pie, tewari2020stylerig}, such as attributes labels or facial $3$D priors, to completely unsupervised and zero-shot approaches~\cite{harkonen2020ganspace,shen2020closedform,voynov2020unsupervised,wang2021a,patashnik2021styleclip,xia2021tedigan,gal2021stylegannada}.  

\paragraph{GAN Inversion}

%With the phenomenal success of latent-based editing, it was only natural that considerable effort would be invested into harnessing StyleGAN's latent space to edit real images. \ron{This is too long to my opinion}

However, applying these editing methods to real images requires one to first find the corresponding latent representation of the given image, a process referred to as \textit{GAN inversion} \cite{zhu2016generative,lipton2017precise,creswell2018inverting, yeh2017semantic, xia2021gan}. 
Multiple works have studied inversion in the context of StyleGAN.
% this critical problem of StyleGAN inversion, by
They either directly optimize the latent vector to reproduce a specific image ~\cite{abdal2019image2stylegan,abdal2020image2stylegan++,semantic2019bau,zhu2020improved,gu2020image, xu2021continuity, roich2021pivotal}, or train an efficient encoder over large collection of images~\cite{zhu2020domain, pidhorskyi2020adversarial,guan2020collaborative,richardson2020encoding,tov2021designing,alaluf2021restyle,kang2021gan,kim2021exploiting, alaluf2021hyperstyle}. Typically, direct optimization is more accurate, but encoders are faster at inference. Moreover, due to their nature as highly parametric neural function estimators, encoders display a smoother behavior, tending to produce more coherent results over similar inputs. We draw on these benefits in our work.

Earlier inversion methods produced code in one of two spaces: either in the native latent space of StyleGAN, denoted \w, or in the more expressive \wplus, where a distinct latent code is assigned to each of the generator's layers.
It has since been shown ~\cite{zhu2020improved, tov2021designing} that $\mathcal{W}$ exhibits a higher degree of editability --- latent codes in this space can be more easily manipulated while preserving a higher degree of realism. On the other hand, $\mathcal{W}$ offers poor expressiveness, resulting in inversions that are often inconsistent with the target identity.
Tov~et al.~\cite{tov2021designing} define this as the distortion-editability trade-off. They suggest that the two aspects can be balanced by designing an encoder that predicts codes in \wplus which reside close to $\mathcal{W}$. 
More recently, Roich~\etal~\shortcite{roich2021pivotal} demonstrated that one may side-step this trade-off. By fine-tuning the generator around an initial inversion code, dubbed the 'pivot', they achieve state-of-the-art reconstructions with a high level of editability. 
However, {\naive}ly applying PTI to a video can result in temporal inconsistencies, as the different pivots are not necessarily coherent. A different approach to improve inversion quality was proposed by Xu et al.\shortcite{xu2021continuity}, utilizing a sequence of frames instead of a single image. 

\paragraph{Video Generation using StyleGAN}

While most works have explored the use of StyleGAN in the image domain, a few recent works sought to bring its many benefits to the realm of video generation and manipulation.
On the generative front, 
Skorokhodov et al.~\shortcite{skorokhodov2021stylegan} suggest a style-based model which can produce short (e.g. $3$), coherent sequences of frames. However, their method requires temporal datasets, which currently exhibit lower quality and insufficient data.
Other works \cite{fox2021stylevideogan, tian2021good} train a second generator to produce temporally coherent latent codes for a pre-trained, fixed StyleGAN. They aim to disentangle spatial and temporal information. Tian et al.~\shortcite{tian2021good} train a temporal LSTM-based generator.
Taking a different approach, Fox et al.~\shortcite{fox2021stylevideogan} use only a single video. The produced latent sequences are later projected to arbitrary latent codes, resulting in animating random subjects. 
However, while offering solid initial results, these methods do not succeed in faithfully inverting a real video or editing it.

\paragraph{Video Semantic Editing}

Many approaches suggested the editing of facial attributes in images~\cite{lample2017fader, choi2018stargan, liu2019stgan, mokady2019masked, he2019attgan}. However, applying these at the frame level typically results in temporal inconsistencies, leading to unrealistic video manipulations.
To overcome this challenge, video-specific methods have been proposed. 
Duong et al.\shortcite{duong2019automatic} perform facial aging in video sequences using deep reinforcement learning. Closest to our work, Yao et al.~\shortcite{yao2021latent} propose to edit real video using StyleGAN by training a dedicated latent-code transformer to achieve more disentangled editing. These edits are applied as part of their proposed pipeline, which first includes a smoothed optical-flow based cropping-and-alignment step to reduce jitter. They then invert the frames using a \wplus encoder~\cite{richardson2020encoding}, perform the editing using their dedicated transformer, and stitch them back to the original video by employing Poisson blending using a segmentation mask.
In our work, we demonstrate that by building on tools that are already highly consistent, we can achieve improved editing in more challenging settings and with fewer visual artifacts -- without requiring any flow or time-based modules.

%% file: method.tex
\section{Method}

Given a real video and a semantic latent editing direction, we aim to produce an edited video. The outcome should preserve the fidelity of the original frames while modifying them in a temporally coherent manner, achieving meaningful and realistic editing.
To this purpose, we design a pipeline of six components: temporally consistent alignment, encoder-based inversion, generator tuning, editing, stitching tuning, and finally merging the results back into the original frame. In the following section, we describe in detail each core step, the tools used to implement it, and the motivation behind each choice. An overview of our full pipeline is presented in \cref{fig:pipeline}. In addition, a visualization of the state of the video at different stages of the editing pipeline is provided in \cref{fig:pipeline_example}.

\subsection{Alignment}\label{subsec:alignment} We employ a pre-trained StyleGAN2 model for face editing. This model, however, was trained on the FFHQ dataset, where each image was pre-processed. In particular, each image was cropped and aligned around the face. Inverting an image successfully into the latent space of the GAN thereby requires a similar crop-and-align phase. However, this pre-processing procedure consists of discrete steps (e.g. cropping) that are sensitive to the exact locations of extracted facial landmarks. This sensitivity can lead to the emergence of temporal inconsistencies, and thus we aim to reduce it. Inspired by the work of Fox et al.~\shortcite{fox2021stylevideogan}, we employ a gaussian lowpass filter over the landmarks. 
We find that this smoothing is sufficient to overcome any inconsistencies induced by the alignment step. 

\subsection{Inversion} To edit an aligned face, we must invert it into the latent space of the GAN. We do so using PTI \cite{roich2021pivotal}, a method which first discovers a 'pivot' latent code that approximately reconstructs the input image in the GAN's more editable regions, and then fine-tunes the \textit{generator} such that the same pivot code will produce a more accurate version of the target image. The goal of PTI was to overcome the distortion-editability trade-off \cite{tov2021designing} of inversion models, allowing for more accurate yet highly editable reconstructions. We argue that, beyond its original use, PTI can also assist in maintaining temporal coherency. The intuition here is that the original video which we wish to invert is itself temporally coherent. Therefore, if we can exactly reproduce each frame, we are guaranteed to have a coherent inversion.

In practice, however, we observe that PTI's editing performance is susceptible to inconsistencies in the pivots themselves. These manifest in two ways: First, if the pivots reside far from each other in the latent space, editing becomes less consistent. For example, the same face encoded in two different regions of the latent space might grow a different beard when the latent code is adjusted in the same direction. Second, if PTI has to 'fix' attributes (\eg a lack of beard) in the inversion, then further editing of the attribute will not typically account for this fix. As a consequence, if the attribute is inconsistent between pivots (a beard appearing in only some of the inverted frames), then PTI will only need to 'fix' it in some frames, and the resulting edit will differ around these frames.

We propose that both flaws can be corrected by simply replacing PTI's optimizer-based initial inversion (\ie finding the pivot) with an encoder-based version, and specifically e4e~\cite{tov2021designing}. As e4e is a deep neural network with many millions of parameters, it is inherently biased towards learning lower frequency representations~\cite{rahaman2019spectral}. We expect that this property will provide a strong smoothness bias, encouraging any coherent changes between images in consecutive frames to be mapped to coherently changing latent codes. In \cref{sec:quantitative}, we analyze this property and demonstrate that the use of an encoder does indeed outperform optimization-based inversions. Note that this property relies on the smoothness of transitions between input frames, and can hence be broken by inconsistent alignment, further motivating the changes of \cref{subsec:alignment}. To reduce memory and time requirements, we employ PTI around all pivot latent codes simultaneously (as opposed to a model for each frame).

Formally, given an $N$ frames source video $\{x_i\}^N_{i=1}$, we denote the cropped-and-aligned frames as $\{c_i\}^N_{i=1}$. We first use the e4e encoder $E$ to obtain their latent inversion $\{w_i\}^N_{i=1} = \{E(c_i)\}^N_{i=1}$. These latent vectors are then used as 'pivots' for PTI. Let $r_i = G(w_i; \theta)$ be the image generated from latent code $w_i$ with a generator $G$ parameterized by weights $\theta$, the PTI objective is defined as:
{\small
\begin{align}
\underset{\theta}{\min} \frac{1}{N}\sum^{N}_{i=1} ( \mathcal{L}_{\text{LPIPS}}(c_i, r_i) + \lambda^{\text{P}}_{L2}\mathcal{L}_{L2}(c_i, r_i)) + \lambda^{\text{P}}_R \mathcal{L}_{R}.
\end{align}
}
Where $\mathcal{L}_{\text{LPIPS}}$ is the LPIPS perceptual loss proposed by Zhang et al.~\shortcite{Zhang2018TheUE}, $\mathcal{L}_{L2}$ is pixel-wise MSE distance, and $\mathcal{L}_{R}$ is the locality regularization described by Roich et al.~\shortcite{roich2021pivotal}. $\lambda^{\text{P}}_{L2}, \lambda^{\text{P}}_R, $ are constants across all experiments.

\subsection{Editing}
Having acquired a set of temporally consistent inversions, we now turn to edit them. We demonstrate that our method works well with off-the-shelf linear editing techniques~\cite{shen2020interpreting,patashnik2021styleclip}.
We expect that, since StyleGAN itself is prone to low-frequency representations (and further motivated by path-length regularization), it will apply sufficiently consistent edits for nearby latent codes. In other words, if we edit temporally-smooth pivots using StyleGAN, we expect to generate a temporally smooth sequence. As we later demonstrate, this expectation aligns with our experimental results. Formally, given a semantic latent editing direction $\delta w$, we utilize the PTI-weights $\theta_p$ to obtain our edited frames $e_i = G(w_i+\delta w; \theta_p)$. 

\subsection{Stitching Tuning}

\input{resources/figures/stitching}

As a final step, we must inject the edited face $e_i$ back into the original video. Merely overwriting the original location of the cropped face is insufficient, as the editing process typically leads to changes in the background which are noticeable around the crop boundary. For instance, \cref{fig:pipeline_example,fig:ablation} demonstrate that artifacts emerge when stitching the crop {\naive}ly. Prior works~\cite{yao2021latent} elected to limit the modifications to the face area by using segmentation masks derived from the landmarks detected in the alignment step and performing Poisson blending. Such stitching, however, is still prone to inconsistencies around the border, leading to artifacts such as the appearance of ghostly outlines (see \cref{fig:comparison}). We propose to tackle this limitation through a novel tuning technique inspired by PTI, referred to as \textit{stitching tuning}. An overview of this technique is provided in \cref{fig:stitching}. For each edited frame, we designate a boundary at the edge of the segmentation mask. We then briefly fine-tune our generator around the \textit{edited pivot} with a dual objective. First, we aim to restore the boundary to its pre-inversion values, that is to blend it perfectly in the original frame. Second, we wish to retain our editing results, by requiring similarity to the edited frame in the area covered by the segmentation mask. This short tuning session can successfully restore the boundary and induce a smooth transition towards the center of the image, without affecting the edited face. 

Formally, we first use off-the-shelf pre-trained segmentation network \cite{yu2021bisenet} to produce segmentation masks $\{m_i\}^N_{i=1}$ for all frames. We then perform dilation on each to obtain a set of expand masks $\{m^d_i\}^N_{i=1}$. The boundary is defined as their element-wise \textit{xor} operator $\{b_i\}^N_{i=1} = \{m_i \oplus m^d_i\}^N_{i=1}$. Let $s_i = G(w_i+\delta w; \theta_{st})$ denote the outcome of the stitching tuning procedure, our first objective is blending the boundary to the original image:
\begin{align}
\mathcal{L}_{b,i} = \mathcal{L}_{L1}(s_i \odot b_i, x_i \odot b_i),
\end{align}
where $\odot$ is the element-wise multiplication. The second objective is preserving the editing result:
\begin{align}
\mathcal{L}_{m,i} = \mathcal{L}_{L1}(s_i \odot m_i, e_i \odot m_i).
\end{align}
These loss terms used to further optimize the generator weights $\theta_{st}$ for each frame:
\begin{align}
\underset{\theta_{st}}{\min} \mathcal{L}_{b,i} + \lambda_m\mathcal{L}_{m,i} , 
\end{align}
where the weights are initialized with the PTI weights $\theta_p$, and  $\lambda_m$ is constant across all experiments. Finally, we perform the inverted alignment and stitch each frame $s_i$ to the original frame $x_i$ using the dilated masks $\{m^d_i\}^N_{i=1}$. Gaussian blur might be applied to further smooth the edges of the masks.

\subsection{Implementation Details}

For all experiments, we set $\lambda^{\text{P}}_{L2} = 10$, $\lambda^{\text{P}}_R = 0.1 $ and $\lambda_m = 0.01$. When tuning a model with PTI we use a learning rate of $3e-5$ and train until each frame is observed 80 times. When tuning a model for stitching, we increase the learning rate to $3e-4$ and the number of training iterations to 100 per frame. For all other implementation details we follow PTI.

% This allows us to reduce editing times from approximately 5 hours per video to approximately 50 minutes on a single NVIDIA 2080 RTX. \rmc{The real reason is that it actually produce an even smoother result. Rinon, Let's add it to the main text}

%% file: resources/figures/stitching.tex
\vspace{-0.1cm}
\begin{figure}[tb]
    \centering
    \setlength{\belowcaptionskip}{-5pt}
    \includegraphics[width=0.99\linewidth]{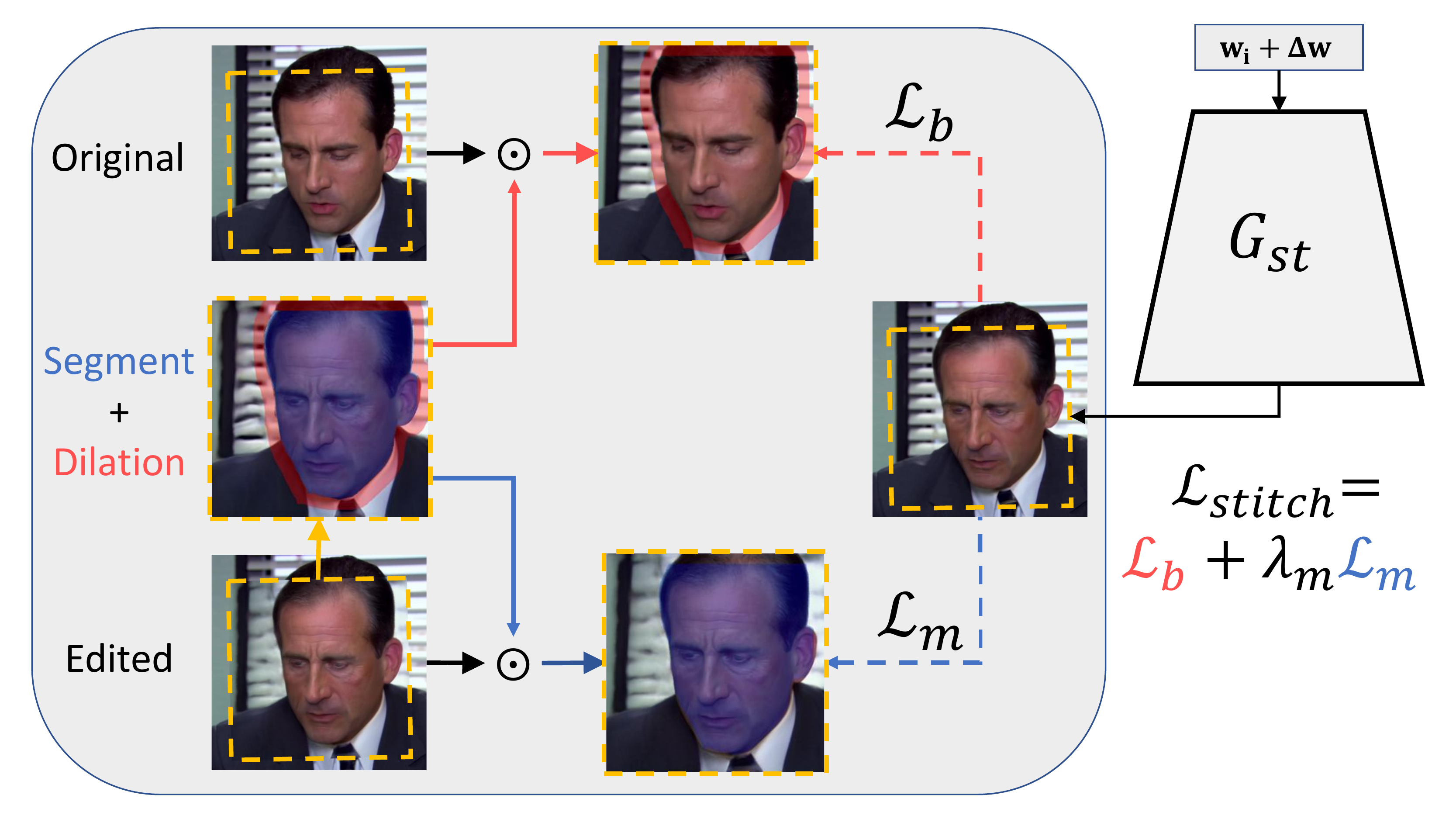}
    % \vspace{-0.25cm}
    \caption{
    Outline of our stitching-tuning method. We start with generating an edited image using a modified pivot code and segment the image using an off-the-shelf segmentation network \cite{yu2021bisenet}. The segmentation mask is dilated, creating a boundary region. We then fine-tune the generator so that the modified pivot will provide an image that is $(a)$ consistent with the original edit inside the face mask (blue), and $(b)$ consistent with the original background inside the boundary mask (red). We synthesize the final image using the tuned generator and paste it inside the dilated mask region (blue + red).
    }
    % \vspace{-0.2cm}
    \label{fig:stitching}
\end{figure}

%% file: results.tex
\section{Experiments}

We demonstrate the effectiveness of our method by applying it to a range of in-the-wild videos gathered from popular publicly available content. These include challenging scenes characterized by complex backgrounds and considerable movement.
% \st{Unless otherwise noted, all frames were inverted using a pre-trained e4e~\cite{tov2021designing} encoder into the \w space of the network.} \rmc{I believe we end inverting to W+.. anyway we discuss e4e in the method, it belongs to there to my opinion.} 
Individual frames were edited using InterFaceGAN~\cite{shen2020interpreting} and StyleCLIP~\cite{patashnik2021styleclip}. All experiments were conducted on a single NVIDIA RTX 2080. The total editing time for a single $300$ frames video is roughly $1.5$ hours.
The full videos for each setup can be found in our supplementary materials.

\subsection{Qualitative Results}

\input{resources/figures/ours}
\input{resources/figures/ours_harris}

%We conduct a qualitative evaluation of our method using a series of challenging, in-the-wild scenes from popular publicly available content. 

In \cref{fig:ours,fig:harris} we show key frames extracted from videos edited using our method. Our approach can handle challenging, highly detailed backgrounds, as well as considerable head movement and speech, all of which are beyond the scope of the current state-of-the-art. Moreover, by editing at the frame level, our method is inherently compatible with existing editing techniques, can support manipulations in multiple latent spaces, and can be easily applied for both spatially local (e.g., smile) and global (e.g., age) changes. 
% \abc{maybe add some specifics here? something like, note that the stitching maintains the straight lines of the shades in the back, and that the skin tone is preserved well through-out the progression of frames} \rgc{I don't think it's needed. We show this and talk about this in our comparisons / ablations where you can also actually see the effects.}

\input{resources/figures/comparison}

In \cref{fig:comparison} we compare our method to~\cite{yao2021latent}, the current state-of-the-art in semantic editing of faces in videos. 
%When working on their simpler scenes, our method maintains improved identity and avoids pitfalls like the appearance of artifacts around stitching regions. 
When evaluating their method on our more challenging scenes, we observe considerable quality degradation and loss of temporal coherence. Our method, in contrast, maintains high fidelity and consistent editing without relying on any explicit temporal smoothing.
Note in particular the blurry artifacts induced by \cite{yao2021latent} when blending the shirt. Moreover, the employment of optimization-based PTI results in the age changing between the different frames.

\input{resources/figures/ood}
In \cref{fig:ood} we showcase our ability to edit out-of-domain videos. Both the encoder and PTI can seamlessly adapt to animated faces. Furthermore, the alignment of fine-tuned StyleGAN models ensures we can re-use the same editing directions, as previously demonstrated by \cite{gal2021stylegannada}, \cite{alaluf2021hyperstyle} and \cite{zhu2021mind}.

\subsection{Quantitative Results}
\label{sec:quantitative}

We next evaluate our method quantitatively. As previously outlined, we expect that encoder-based methods will be smoother at the local level, avoiding the jitter induced by optimization techniques. On the other hand, we expect them to display considerable identity drift at the global scale - with the minor inversion inconsistencies between the frames building up over time. To validate our intuition and evaluate the temporal coherence of videos, we propose two novel metrics: temporally-local (TL-ID) and temporally-global (TG-ID) identity preservation.

In the first case, TL-ID, we aim to evaluate the video's consistency at the local level. We do so by employing an off-the-shelf identity detection network~\cite{deng2019arcface} to evaluate the identity similarity between pairs of adjacent video frames. To account for the effect of inconsistencies in the identity network itself, we normalize these identity preservation scores by the similarity score of each pair of frames in the original video. Finally, we average the normalized scores over the entire video, and then once again over a set of videos.
Higher TL-ID scores indicate that the method produces \textit{smooth} results, without considerable \textit{local} identity jitter.

Our second metric, TG-ID, employs the same identity detection network and averaging scheme to evaluate the similarity between all possible pairs of video frames, not necessarily adjacent. This metric aims to capture longer-range coherence and identify slow, but consistent identity drift.
For both metrics, a score of $1$ would indicate that the method successfully maintains the identity consistency of the original video.
Note that both metrics only compare frames \textit{within} a given video. As such, they do not measure the similarity of the inversions to those of the original video. Rather, we focus on quantifying the temporal consistency of these inversions and the editing operations they support.

We utilize both metrics to evaluate our method against the baseline PTI~\cite{roich2021pivotal}, as well as Latent Transformer~\cite{yao2021latent} which makes use of a pre-trained pSp~\cite{richardson2020encoding} encoder for inversion. The results are shown in \cref{tb:temporal_id}. As expected, encoder-based methods outperform the optimization-based method on local consistency, demonstrating that they do indeed provide a smooth prior. Moreover, combining an encoder with PTI results in local consistency which is just shy of $1$, showing that the proposed pipeline is sufficient to inherit almost all temporal consistency from the source video.

On the global front, PTI improves identity preservation over longer time spans. While editing is still susceptible to the inconsistencies of the local pivots, the encoder-based methods lead to an identity drift that eventually surpasses the local jitter. PTI, meanwhile, constantly re-aligns the identity to that of the source video, avoiding longer term drift. Notably, PTI does demonstrate some drop in global performance when compared to the local score. We hypothesize that this originates in the increased distance between pivots when optimizing around distant frames.
By leveraging PTI, our method can similarly maintain a high level of global consistency, and even outperform PTI thanks to more consistent pivot codes.

\input{resources/tables/temporal_id}

\subsection{Ablation Study}
\label{sec:ablation}

We further demonstrate the benefits of each component in our pipeline by conducting an ablation study. In \cref{fig:ablation} we show key frames from a video edited using our method when crucial steps of the pipeline are removed or replaced. We showcase the effects of replacing the encoder with an optimization method (w/o e4e), removing the PTI generator-tuning step, replacing the stitching-tuning step by {\naive}ly pasting the edited image inside the segmentation mask, and finally our full pipeline.

Without an encoder, the edited frames become inconsistent when the face undergoes considerable movement or changes in expression. Without PTI, frames are less faithful to the original video, stitching performance suffers, and identity changes over longer time periods.
Without stitching, artifacts appear around the hair and borders of the segmentation mask (\ie the edge of the face). Our full method maintains both local and long term consistency, and seamlessly melds the edited region into the original frame.

\input{resources/figures/ablation}

%% file: resources/figures/ours.tex
\begin{figure}
\setlength{\tabcolsep}{0.5pt}
    \centering
    { \small 
\begin{tabular}{lcccc}
\rotatebox[origin=t]{90}{Original} &
\raisebox{-.32\totalheight}{\includegraphics[width=0.24\columnwidth]{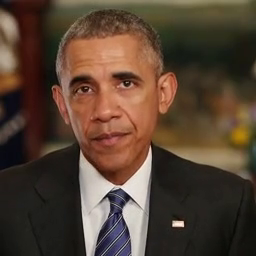}} &
\raisebox{-.32\totalheight}{\includegraphics[width=0.24\columnwidth]{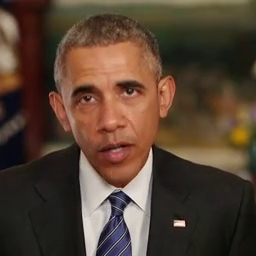}} &
\raisebox{-.32\totalheight}{\includegraphics[width=0.24\columnwidth]{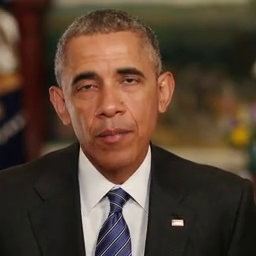}} & 
\raisebox{-.32\totalheight}{\includegraphics[width=0.24\columnwidth]{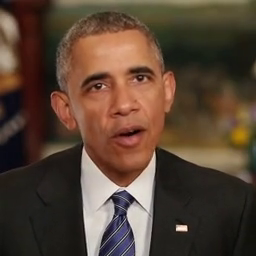}} \\
\noalign{\vskip .5mm}
\rotatebox[origin=t]{90}{$+$Smile} &
\raisebox{-.32\totalheight}{\includegraphics[width=0.24\columnwidth]{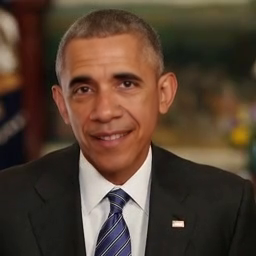}} &
\raisebox{-.32\totalheight}{\includegraphics[width=0.24\columnwidth]{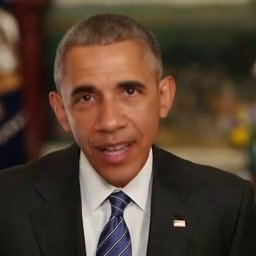}} &
\raisebox{-.32\totalheight}{\includegraphics[width=0.24\columnwidth]{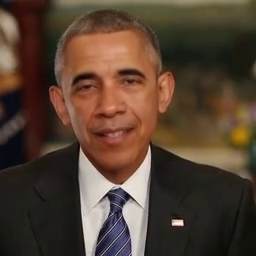}} &
\raisebox{-.32\totalheight}{\includegraphics[width=0.24\columnwidth]{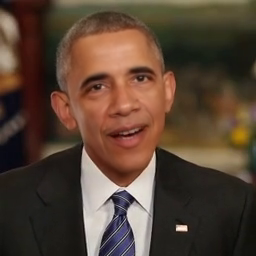}}\\
\noalign{\vskip .5mm}
\rotatebox[origin=t]{90}{$+$Gender} &
\raisebox{-.32\totalheight}{\includegraphics[width=0.24\columnwidth]{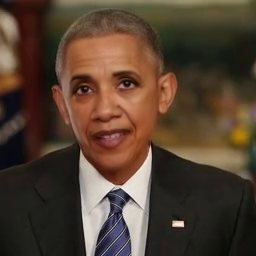}} &
\raisebox{-.32\totalheight}{\includegraphics[width=0.24\columnwidth]{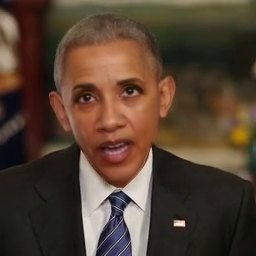}} &
\raisebox{-.32\totalheight}{\includegraphics[width=0.24\columnwidth]{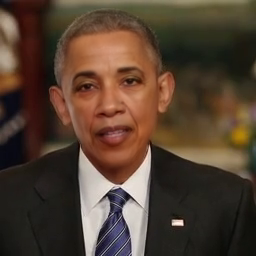}} &
\raisebox{-.32\totalheight}{\includegraphics[width=0.24\columnwidth]{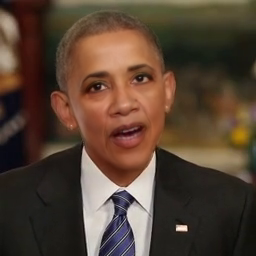}}\\
\noalign{\vskip .5mm}
\rotatebox[origin=t]{90}{$+$Old} &
\raisebox{-.32\totalheight}{\includegraphics[width=0.24\columnwidth]{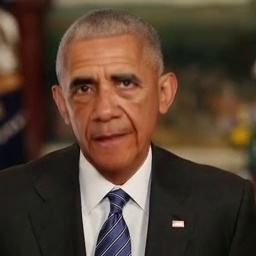}} &
\raisebox{-.32\totalheight}{\includegraphics[width=0.24\columnwidth]{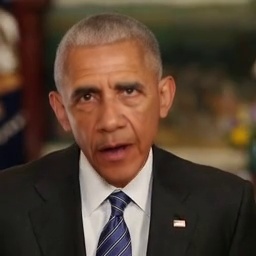}} &
\raisebox{-.32\totalheight}{\includegraphics[width=0.24\columnwidth]{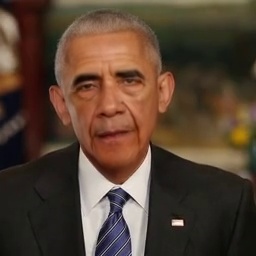}} &
\raisebox{-.32\totalheight}{\includegraphics[width=0.24\columnwidth]{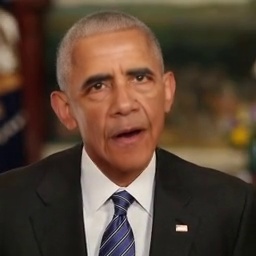}}\\
\noalign{\vskip .5mm}
\rotatebox[origin=t]{90}{$+$Young} &
\raisebox{-.32\totalheight}{\includegraphics[width=0.24\columnwidth]{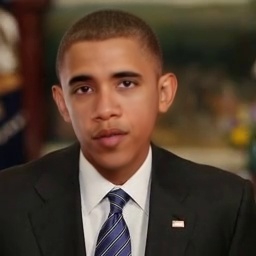}} &
\raisebox{-.32\totalheight}{\includegraphics[width=0.24\columnwidth]{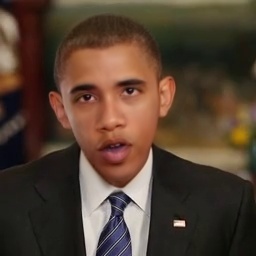}} &
\raisebox{-.32\totalheight}{\includegraphics[width=0.24\columnwidth]{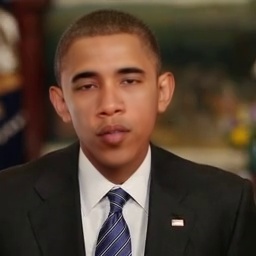}} &
\raisebox{-.32\totalheight}{\includegraphics[width=0.24\columnwidth]{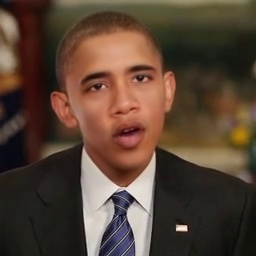}}\\
\end{tabular}
}
\caption{Multiple editing results over a single video. Our model preserves the original video details while enabling a range of semantic manipulations.}
\label{fig:ours}
\end{figure}

%% file: resources/figures/ours_harris.tex
\begin{figure*}
    \centering
    \setlength{\belowcaptionskip}{-6pt}
    \setlength{\tabcolsep}{0.5pt}
    {
    \begin{tabular}{ccccccccccc}
        
        \vspace{-0.0615cm}
        
        % Identity 1

        \raisebox{0.12in}{\rotatebox{90}{Original}} &
        \includegraphics[width=0.095\textwidth]{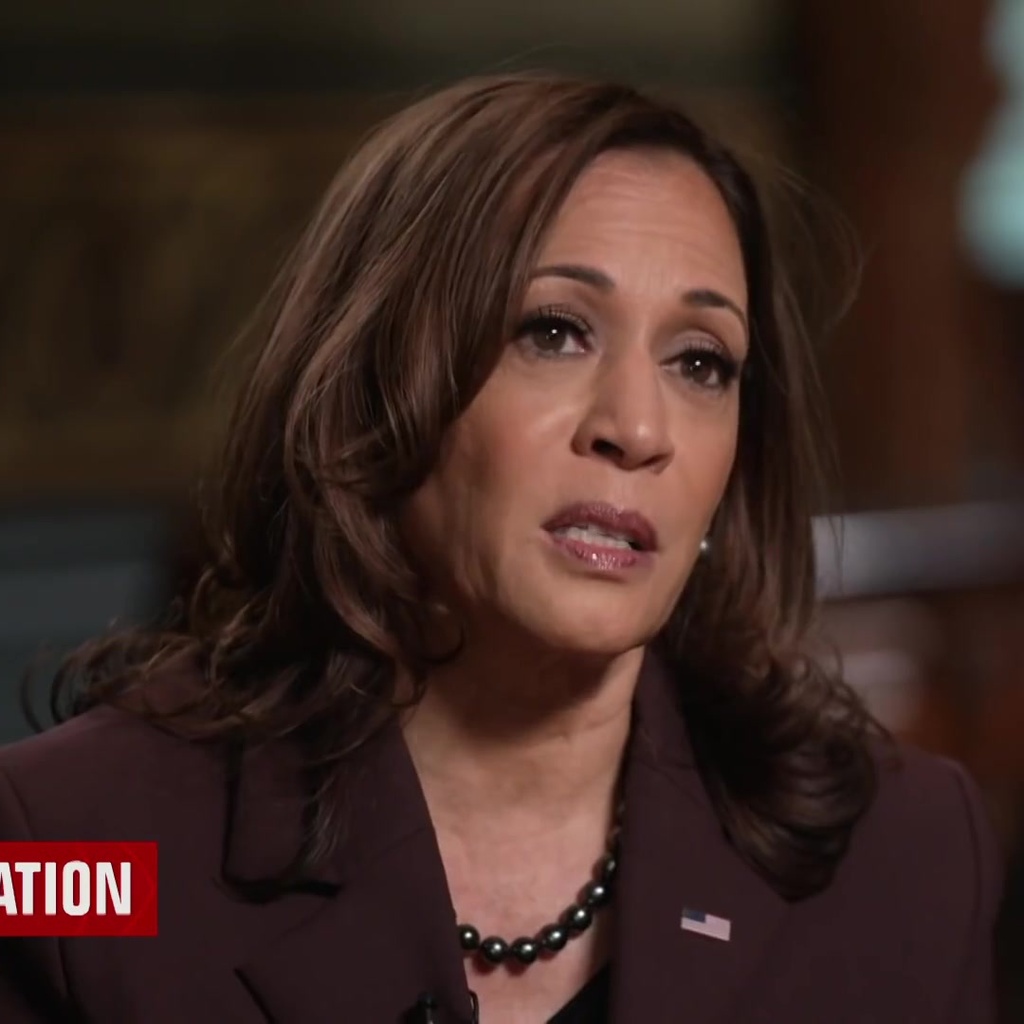} &
        \includegraphics[width=0.095\textwidth]{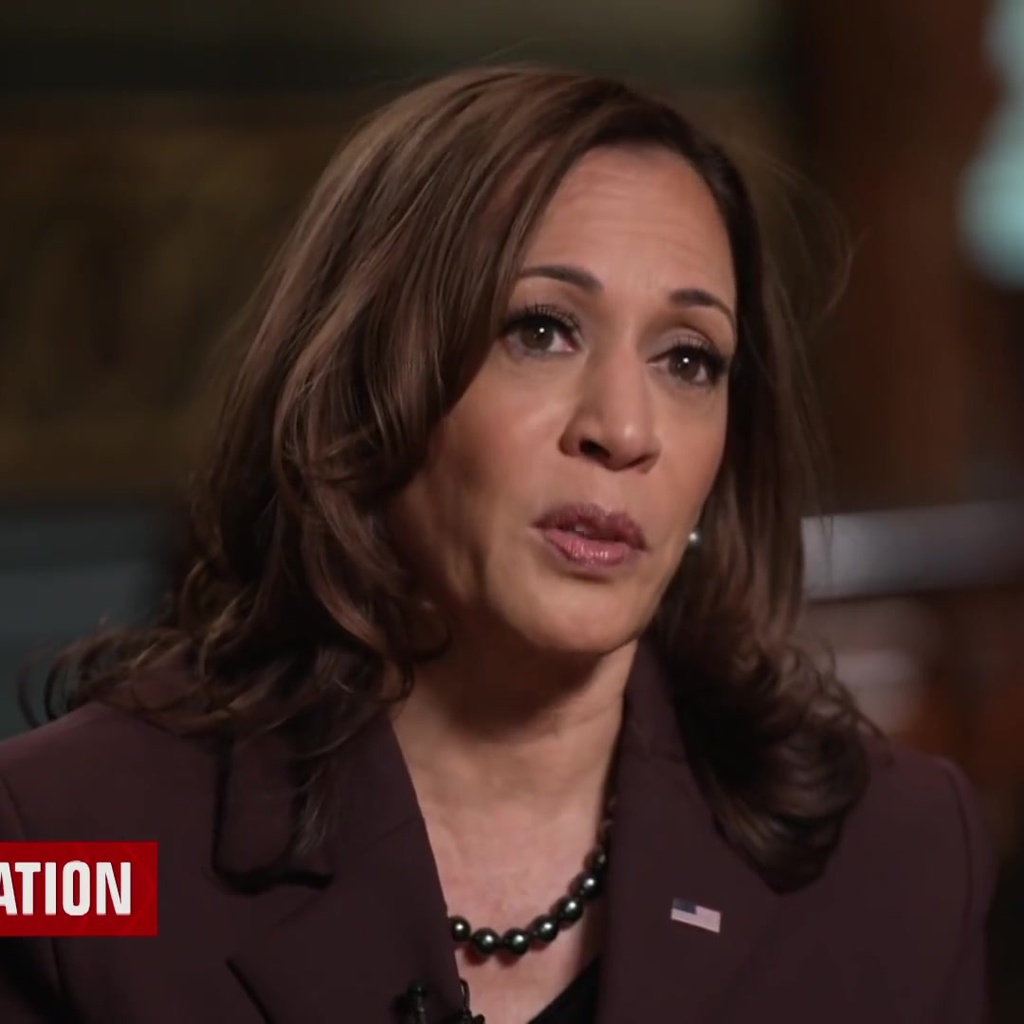} &
        \includegraphics[width=0.095\textwidth]{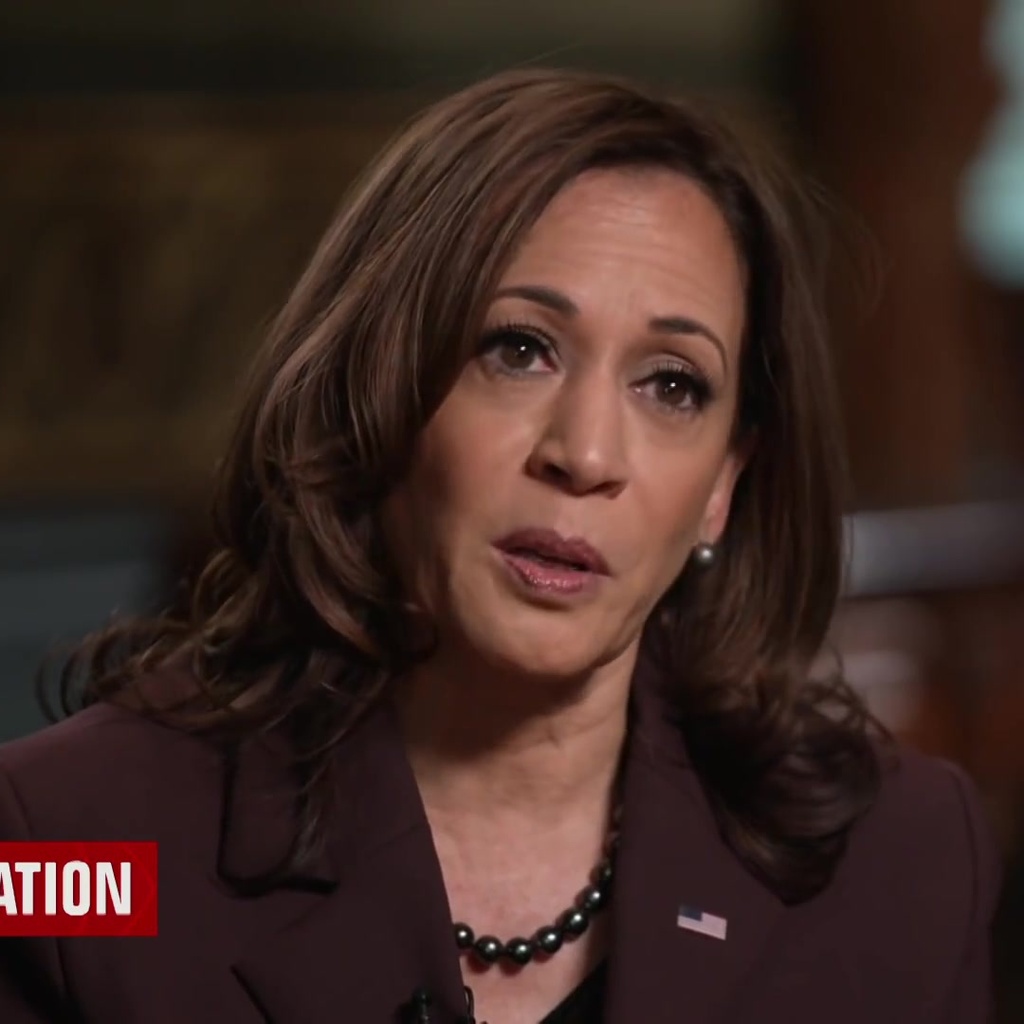} &
        \includegraphics[width=0.095\textwidth]{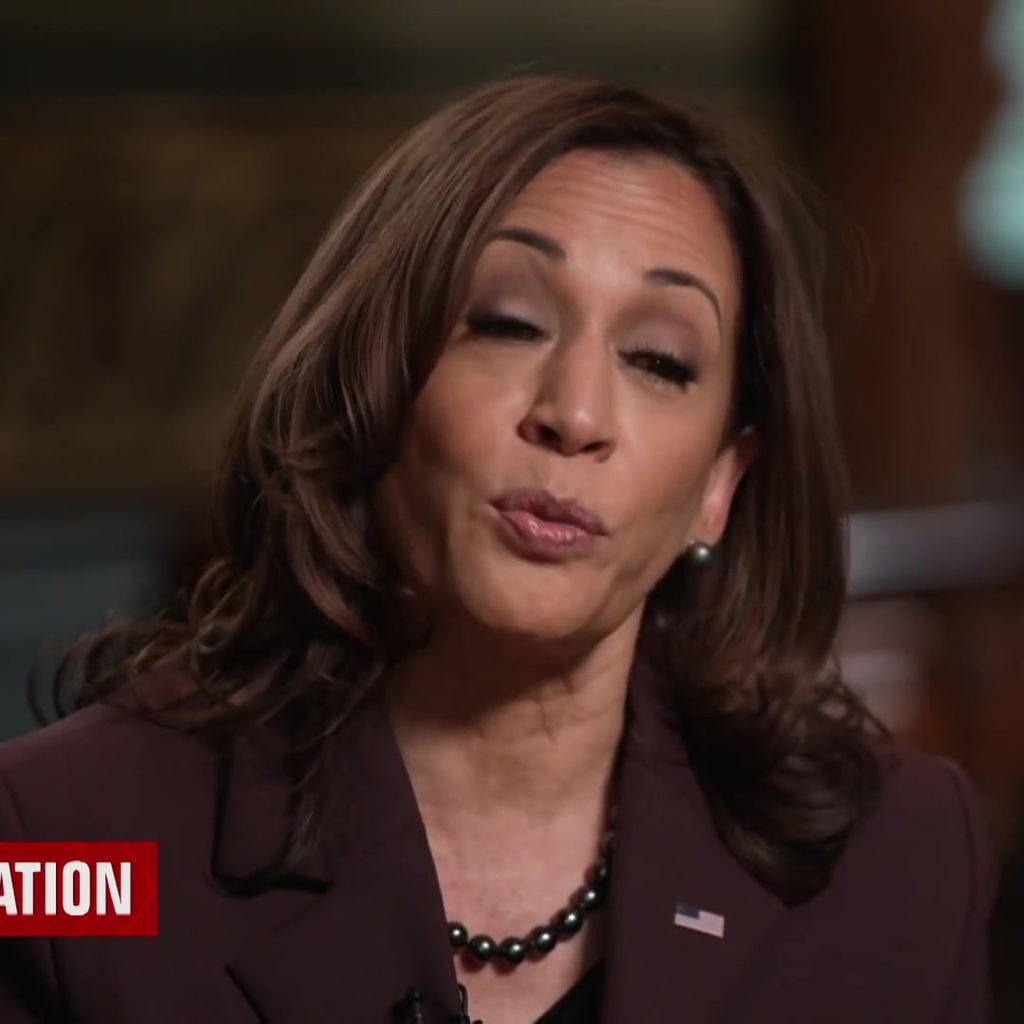} &
        \includegraphics[width=0.095\textwidth]{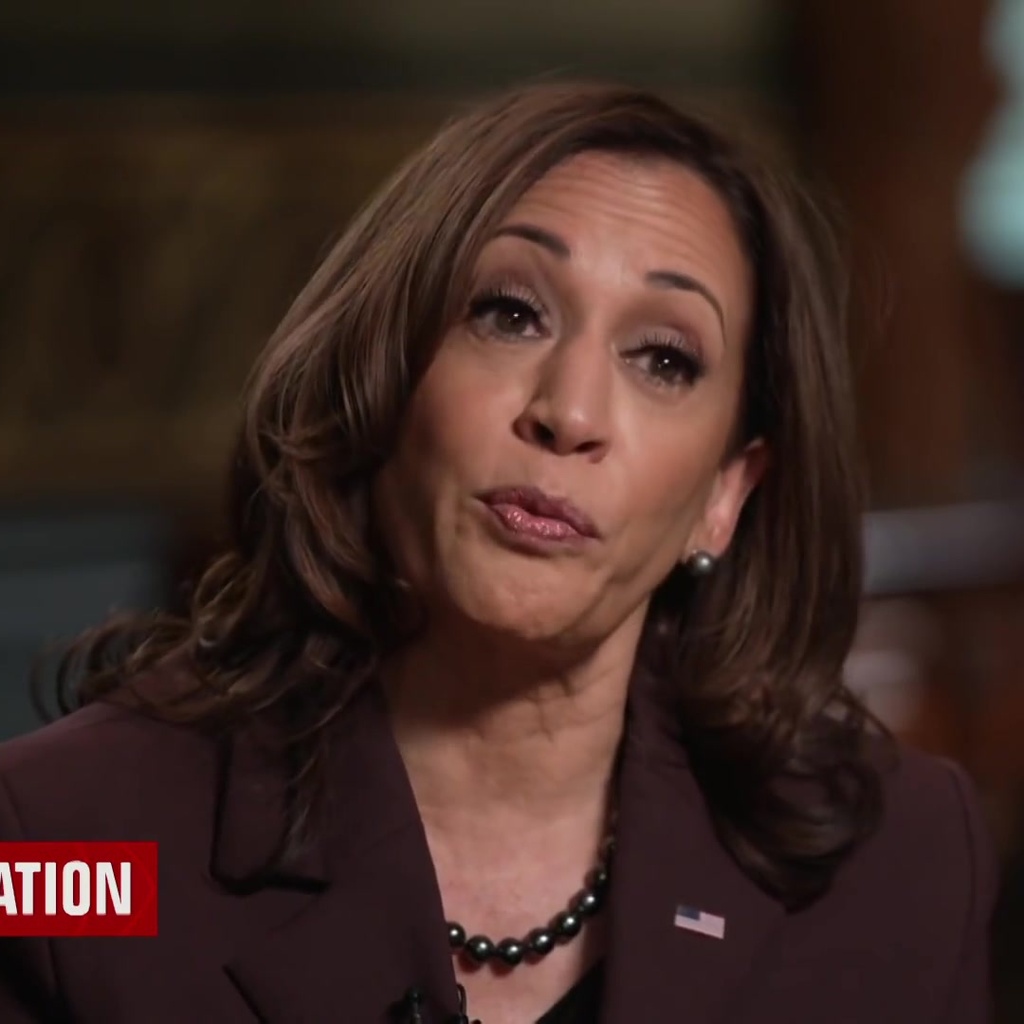} &
        \includegraphics[width=0.095\textwidth]{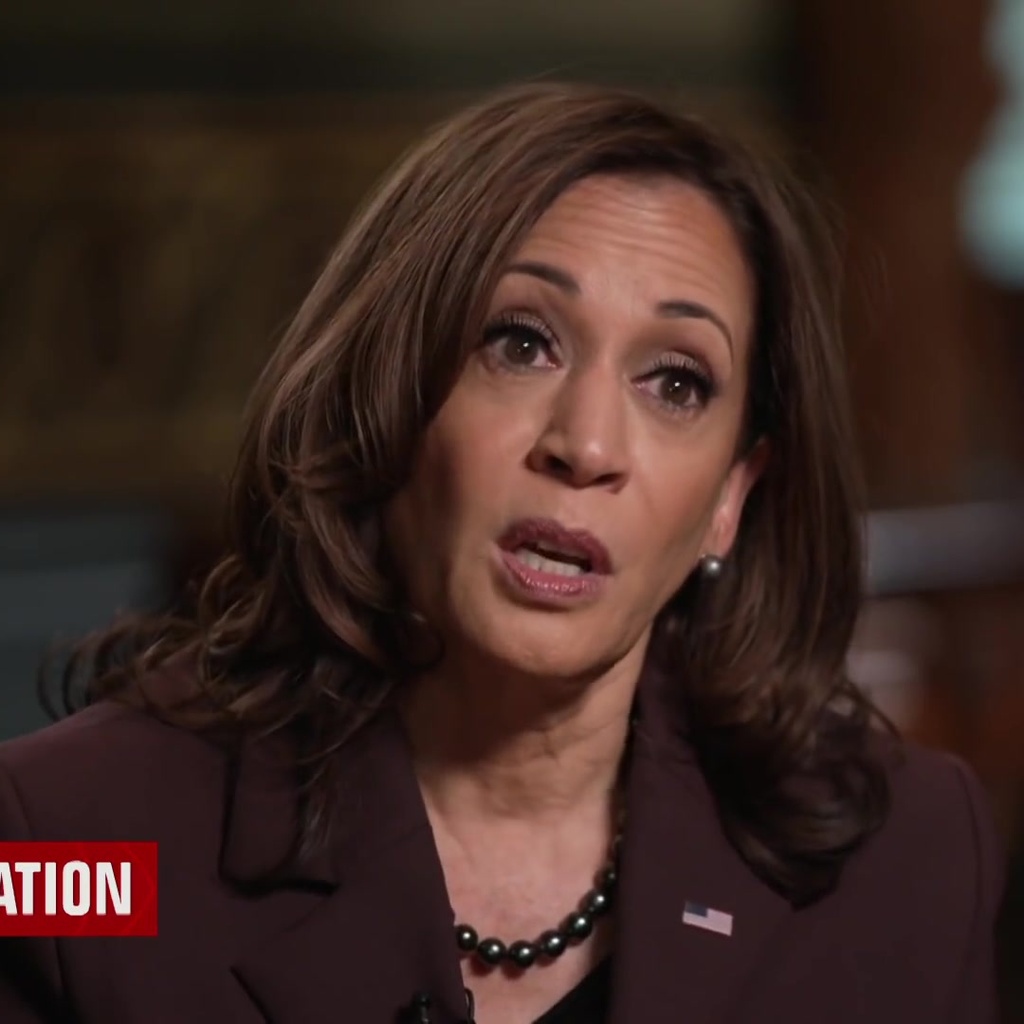} &
        \includegraphics[width=0.095\textwidth]{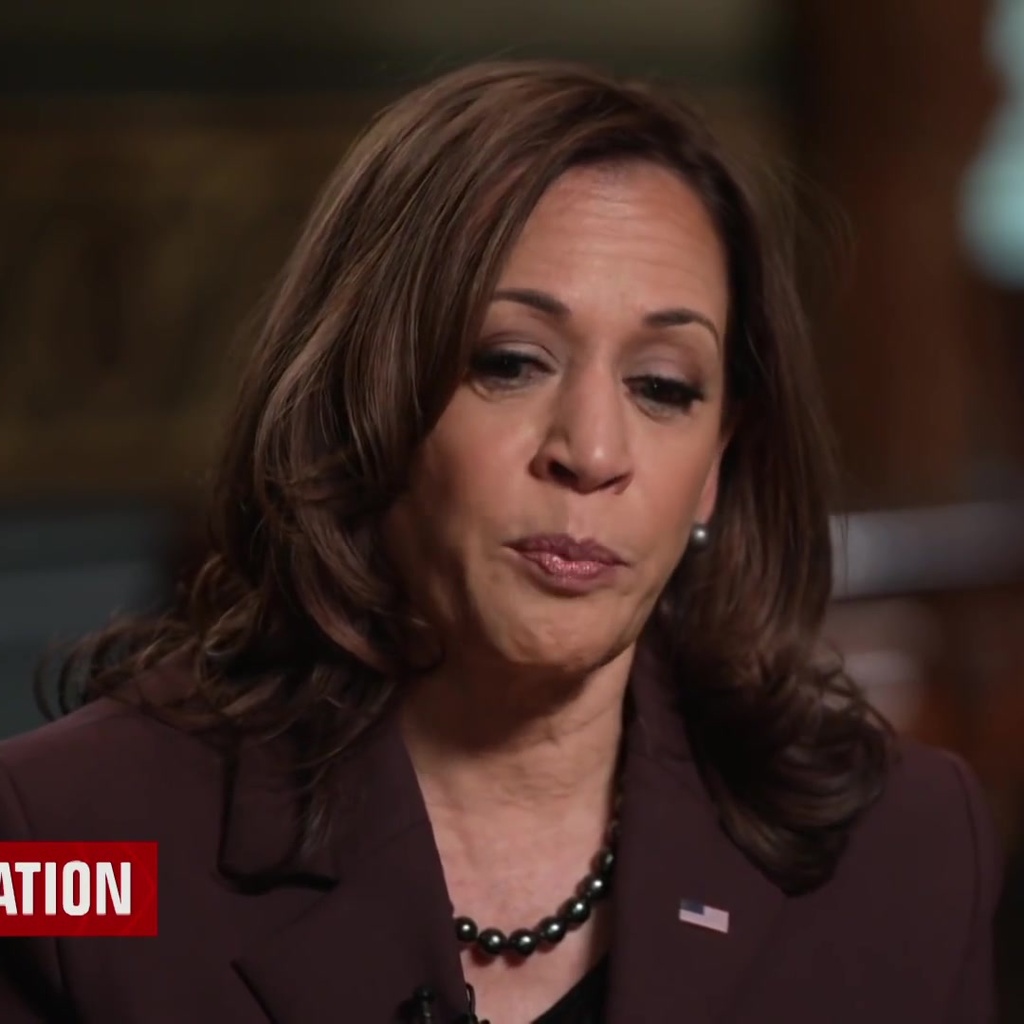} &
        \includegraphics[width=0.095\textwidth]{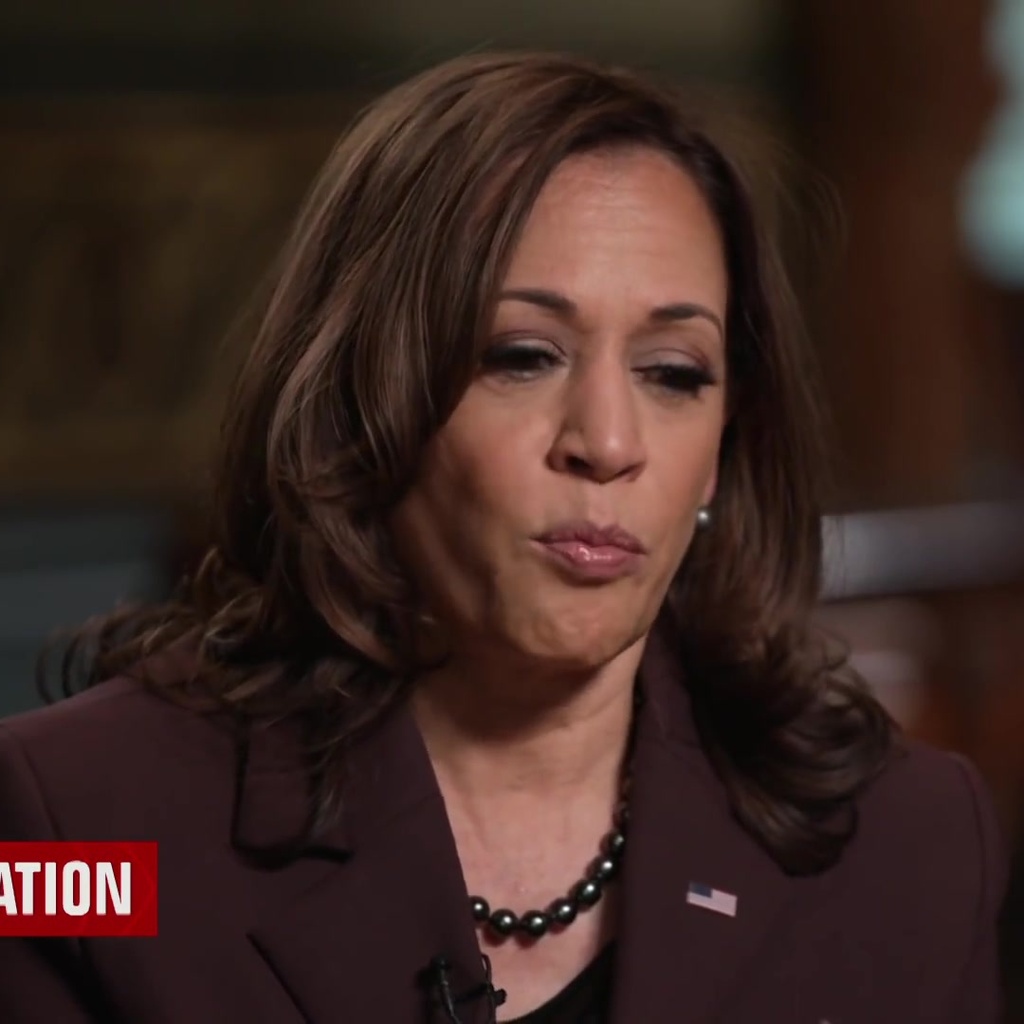} &
        \includegraphics[width=0.095\textwidth]{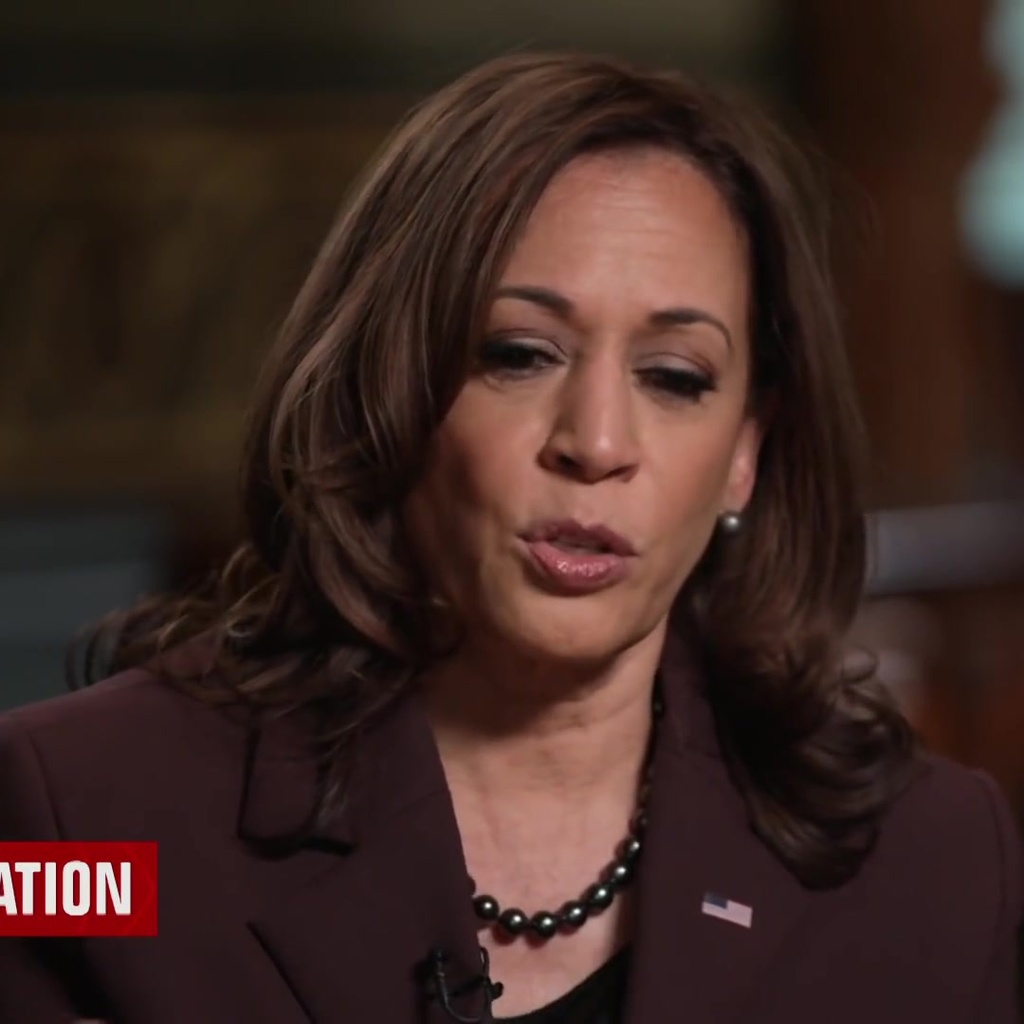} &
        \includegraphics[width=0.095\textwidth]{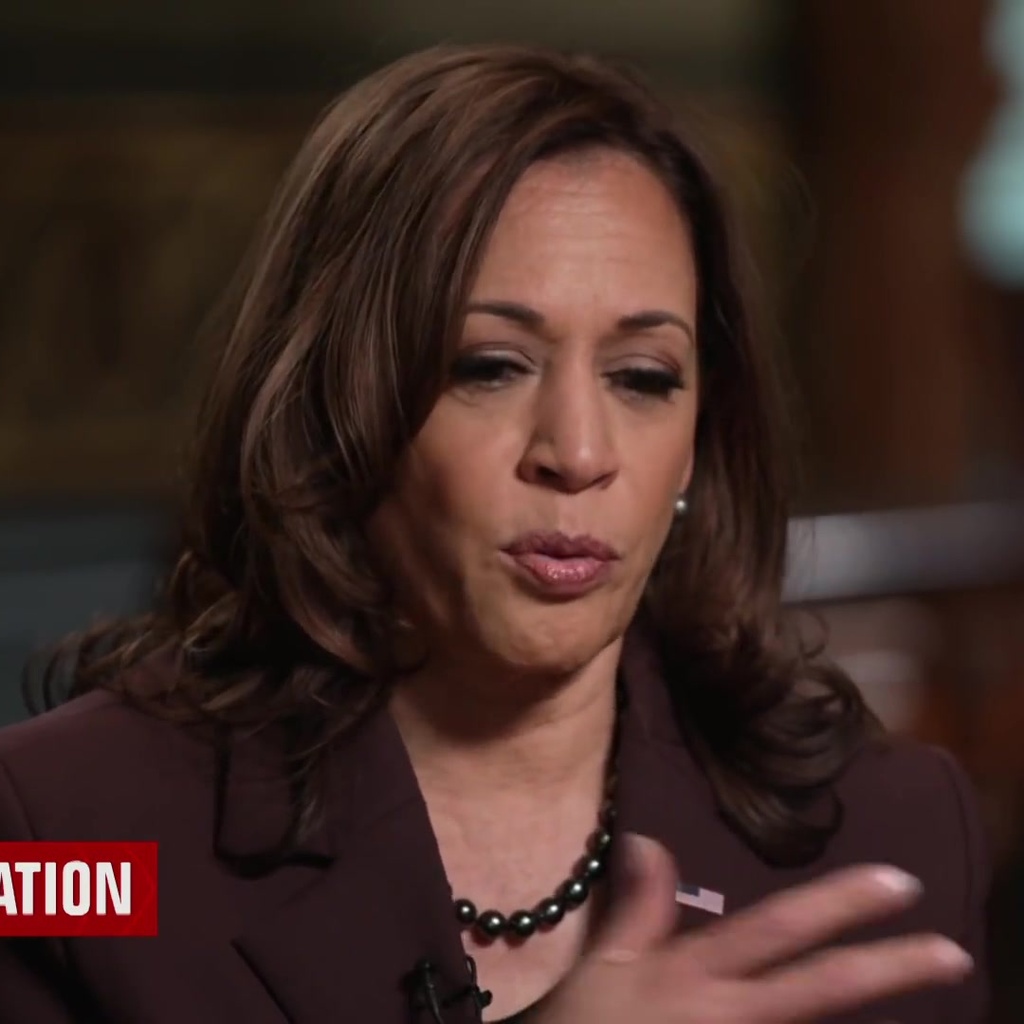} \\
        
        \raisebox{0.18in}{\rotatebox{90}{+Smile}} &
        \includegraphics[width=0.095\textwidth]{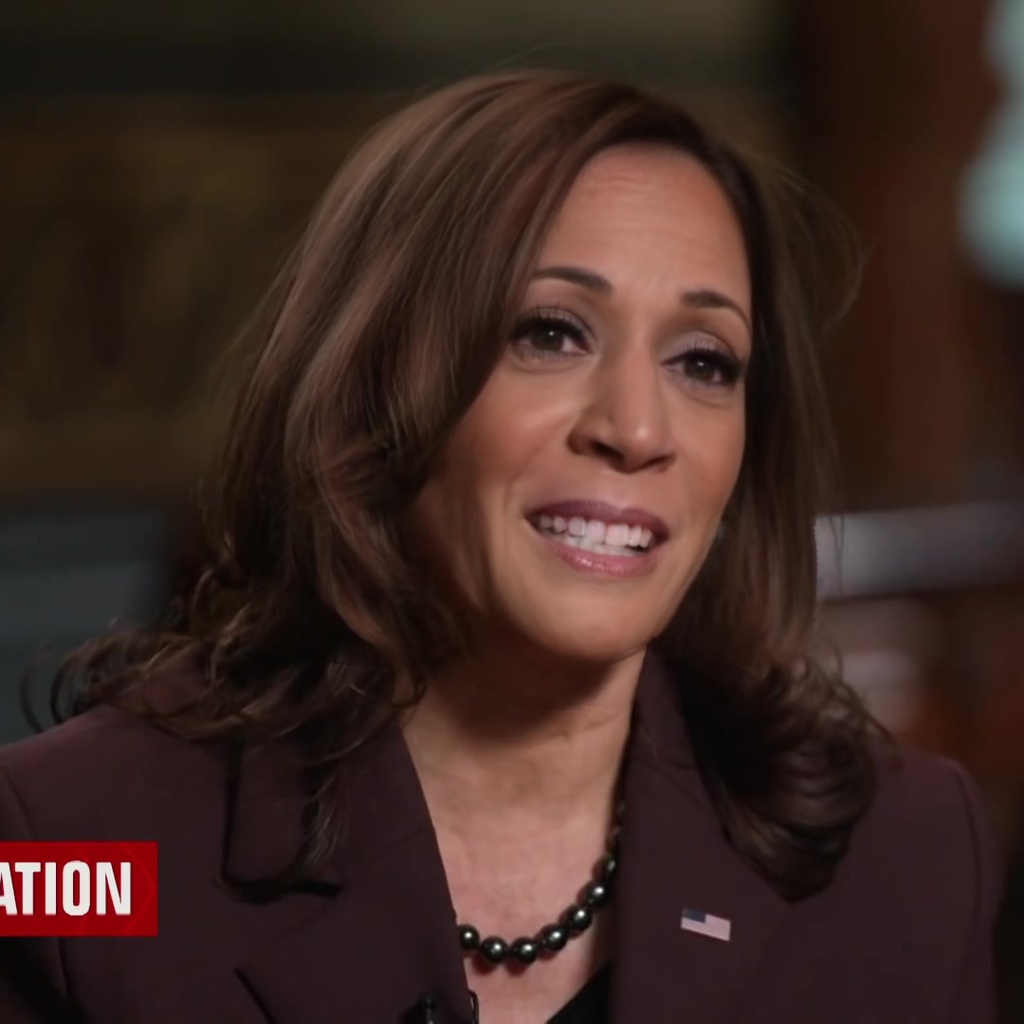} &
        \includegraphics[width=0.095\textwidth]{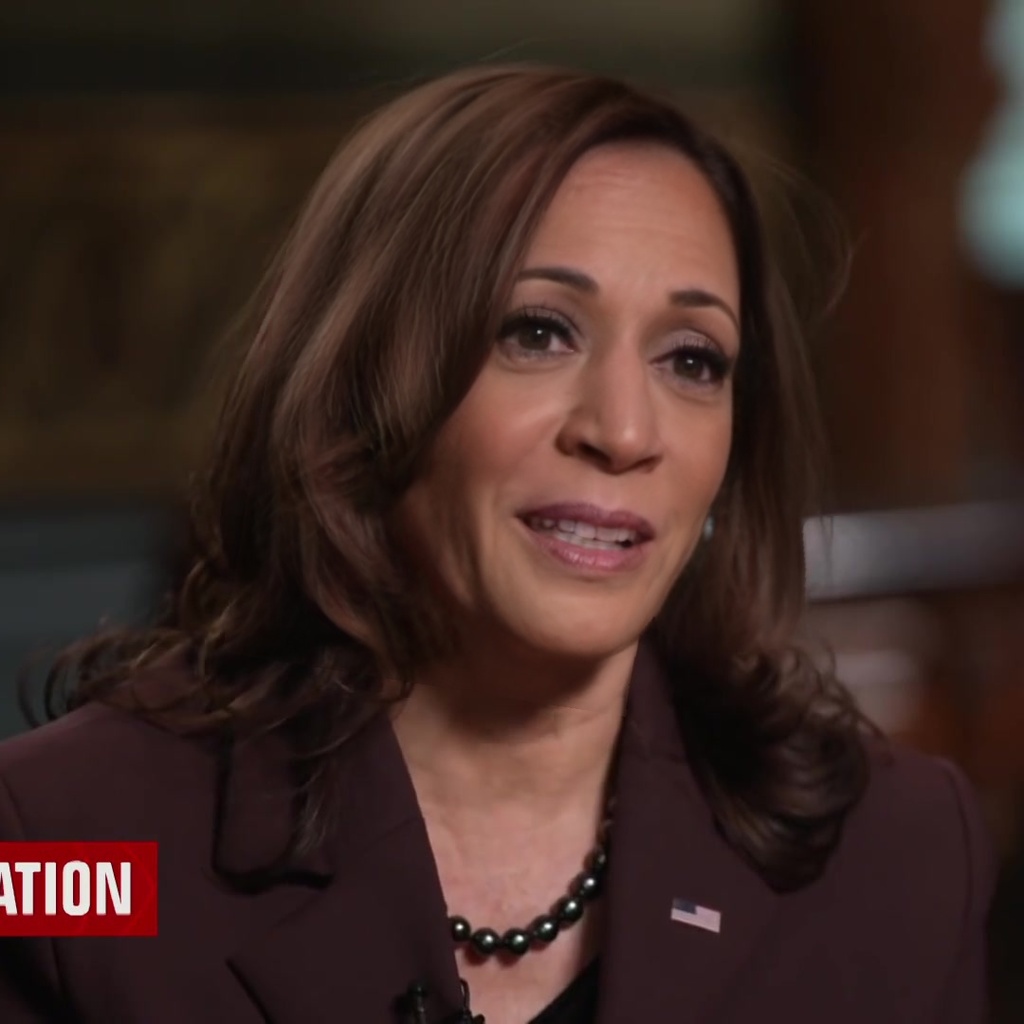} &
        \includegraphics[width=0.095\textwidth]{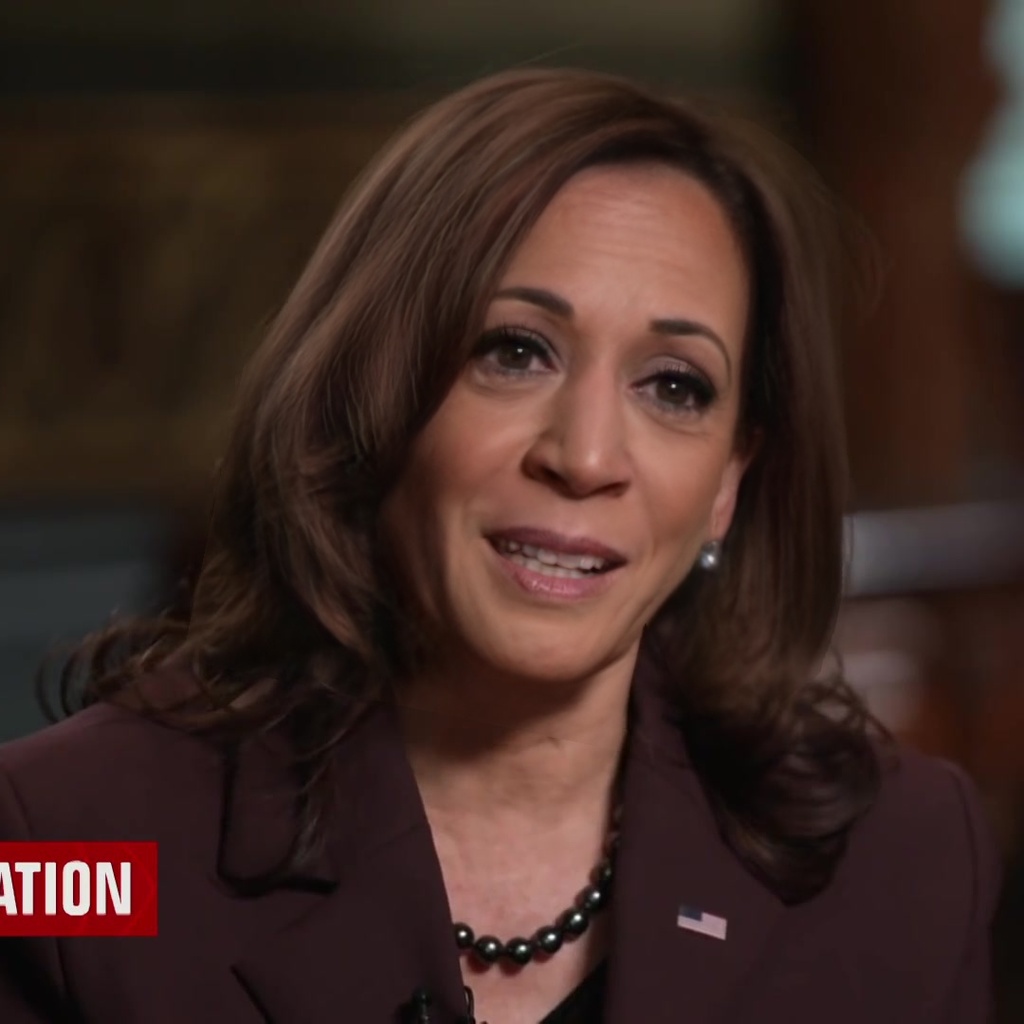} &
        \includegraphics[width=0.095\textwidth]{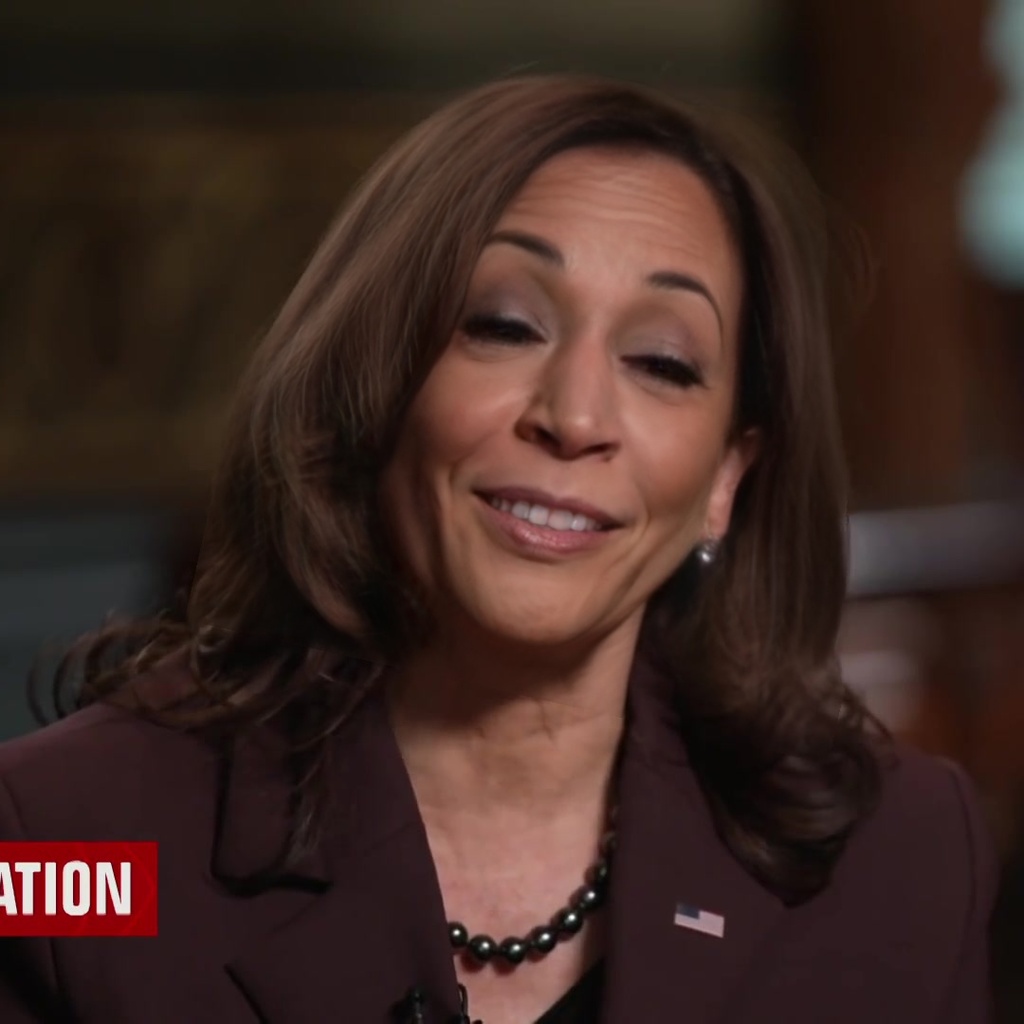} &
        \includegraphics[width=0.095\textwidth]{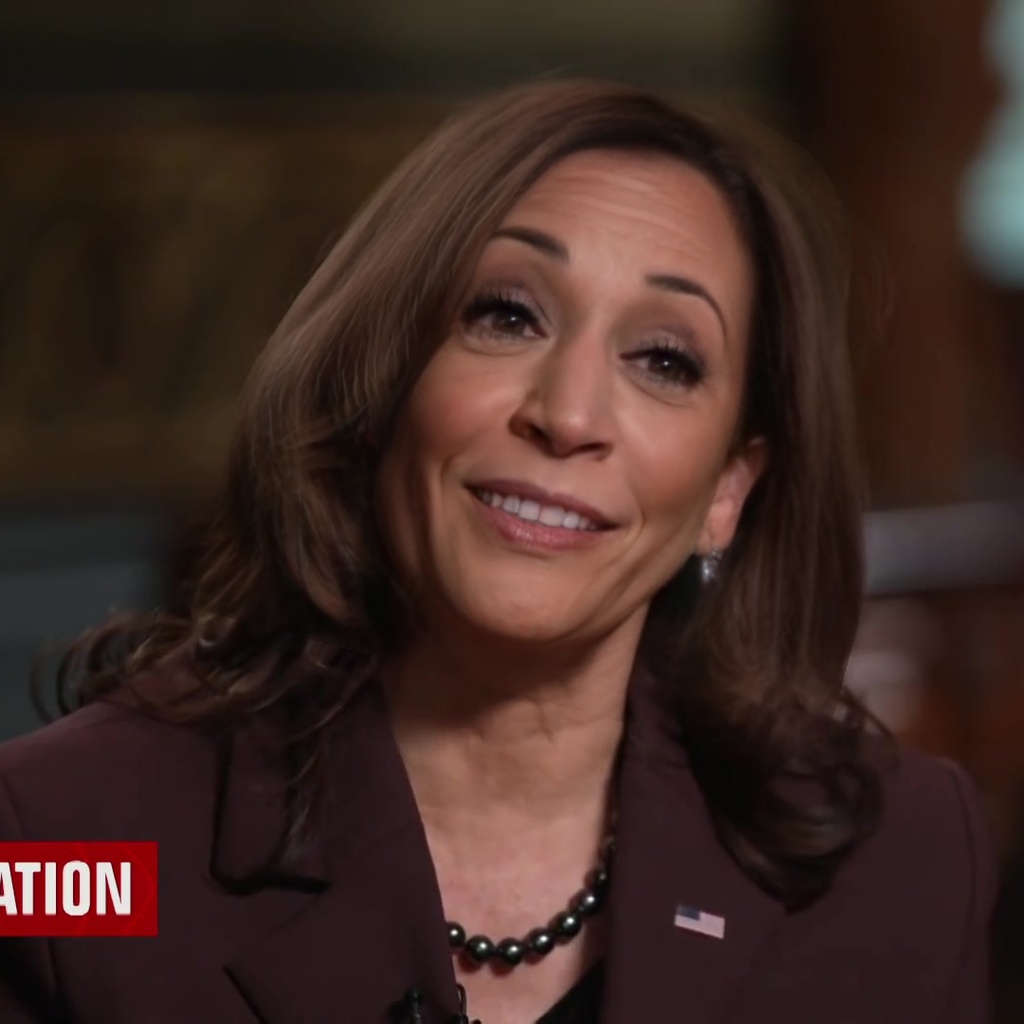} &
        \includegraphics[width=0.095\textwidth]{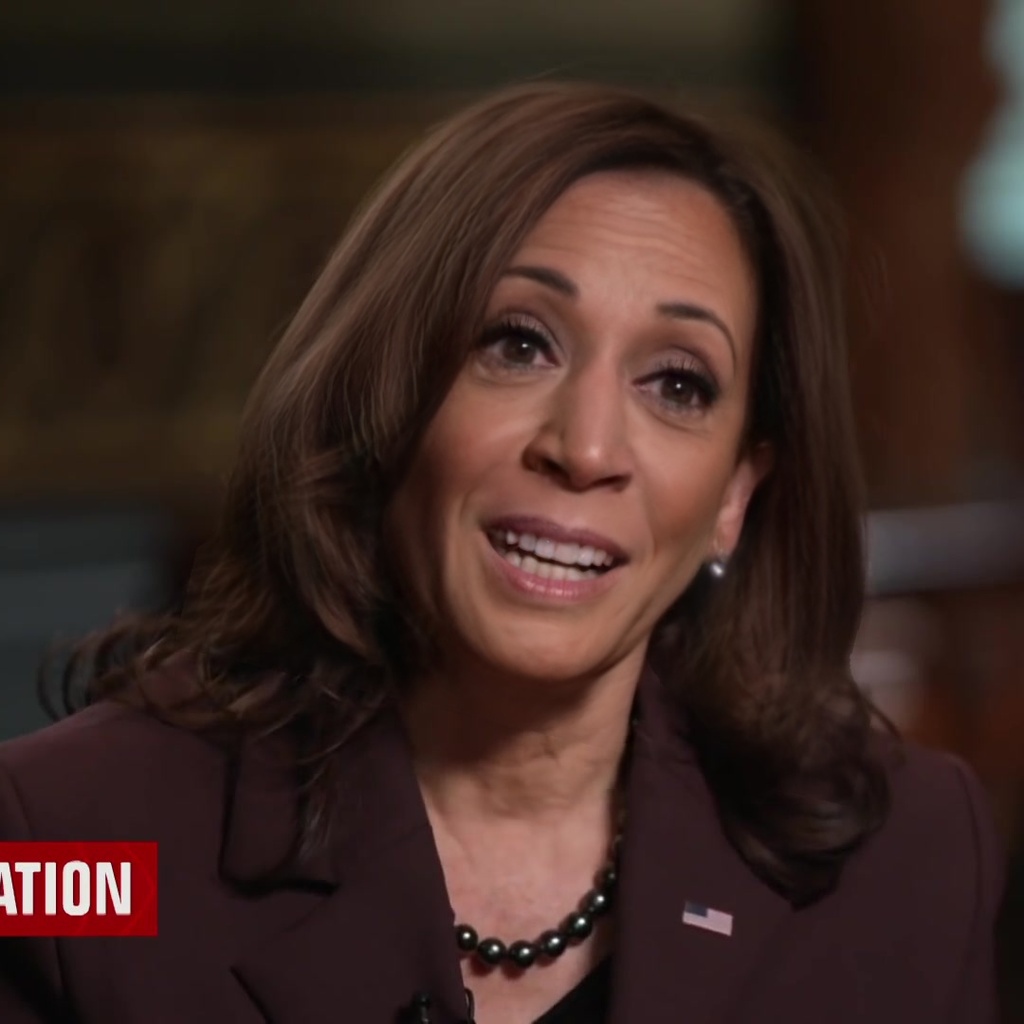} &
        \includegraphics[width=0.095\textwidth]{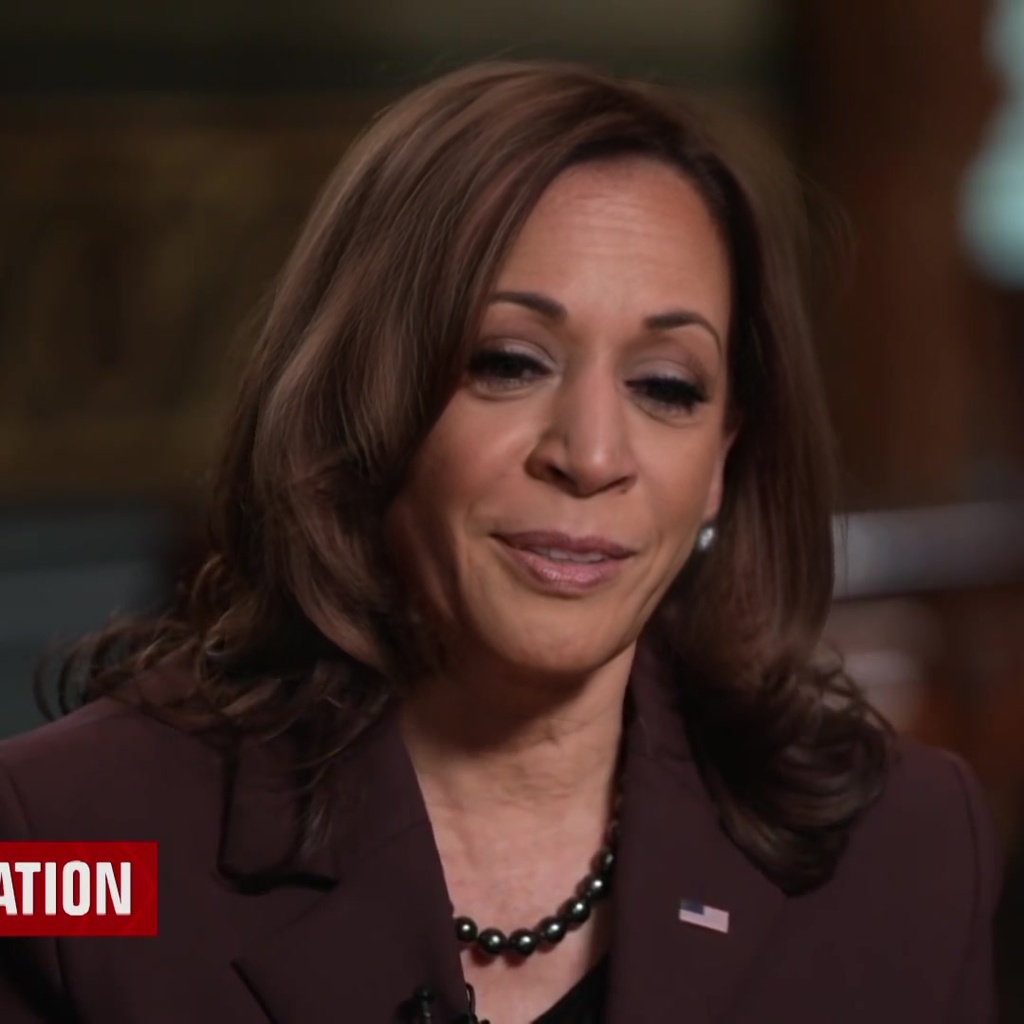} &
        \includegraphics[width=0.095\textwidth]{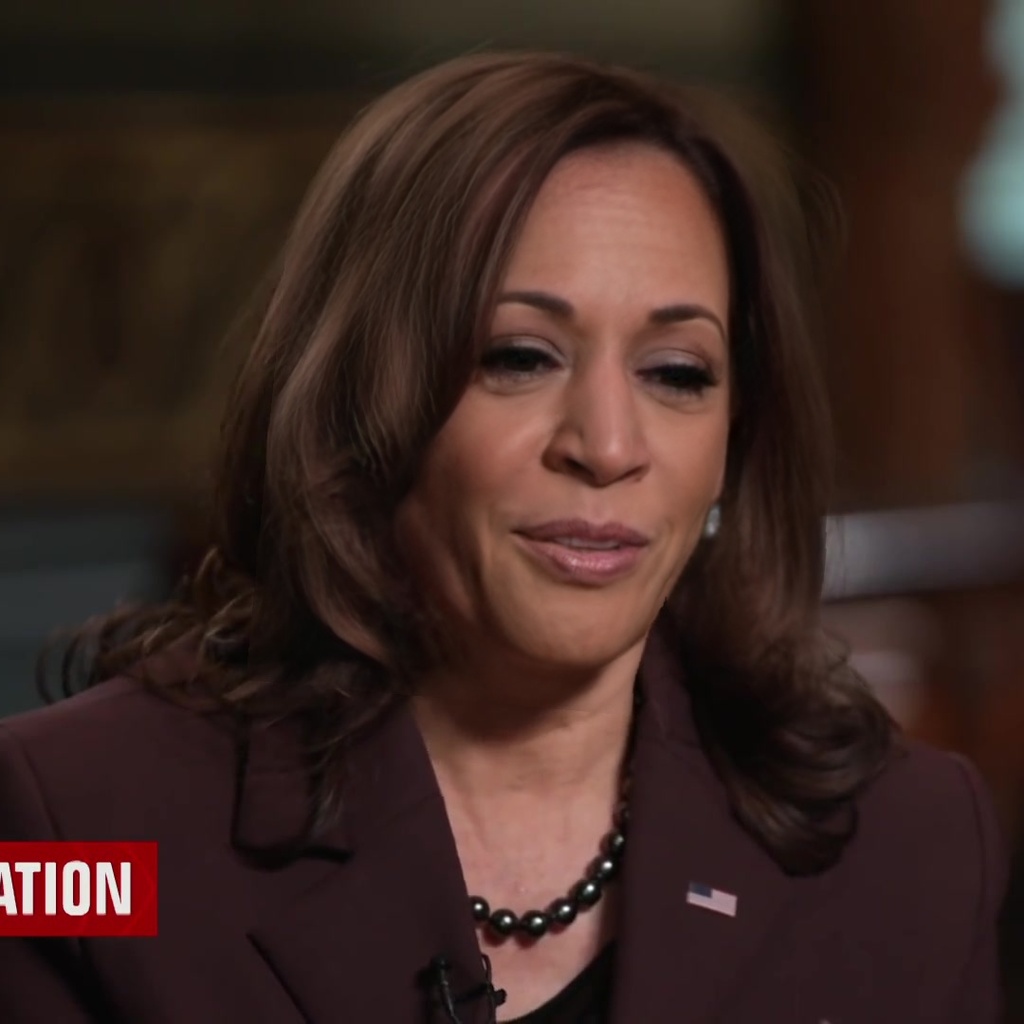} &
        \includegraphics[width=0.095\textwidth]{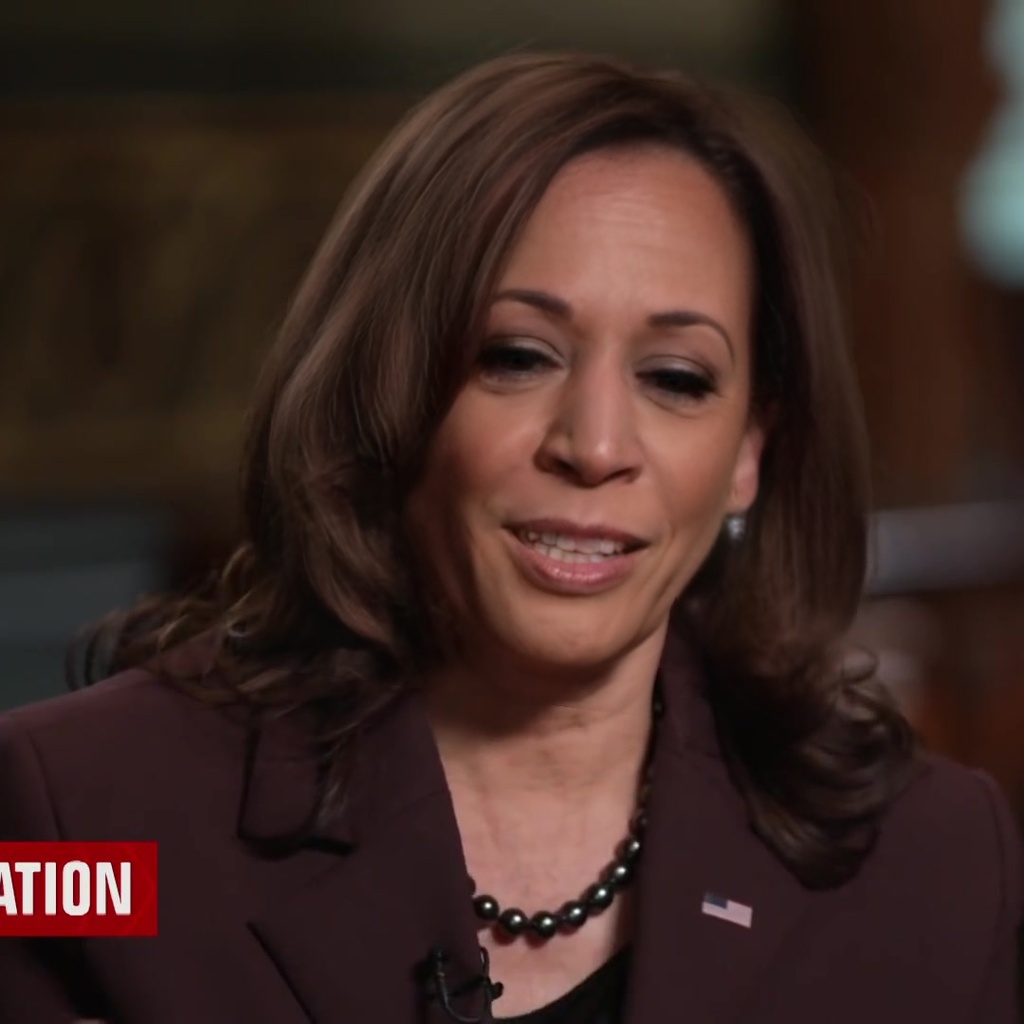} &
        \includegraphics[width=0.095\textwidth]{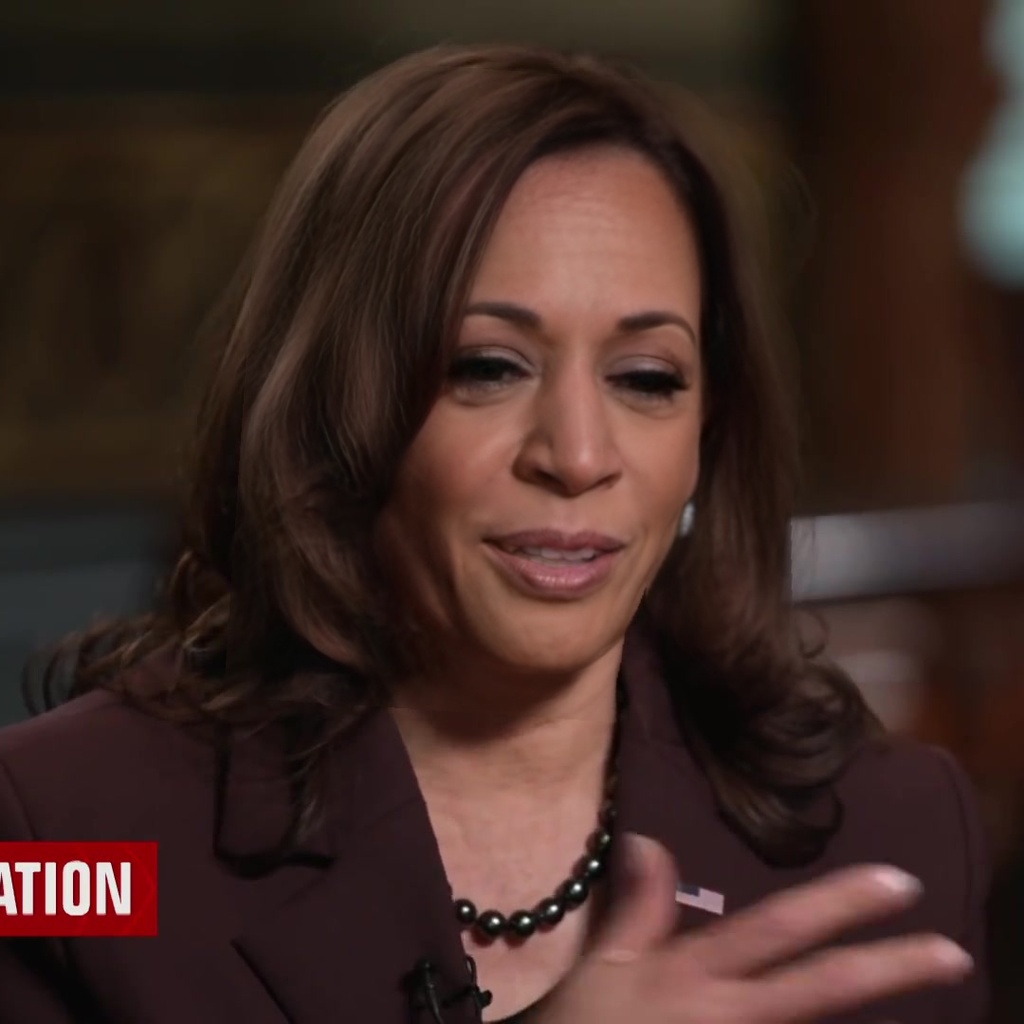} \\
        
        \raisebox{0.13in}{\rotatebox{90}{+Young}} &
        \includegraphics[width=0.095\textwidth]{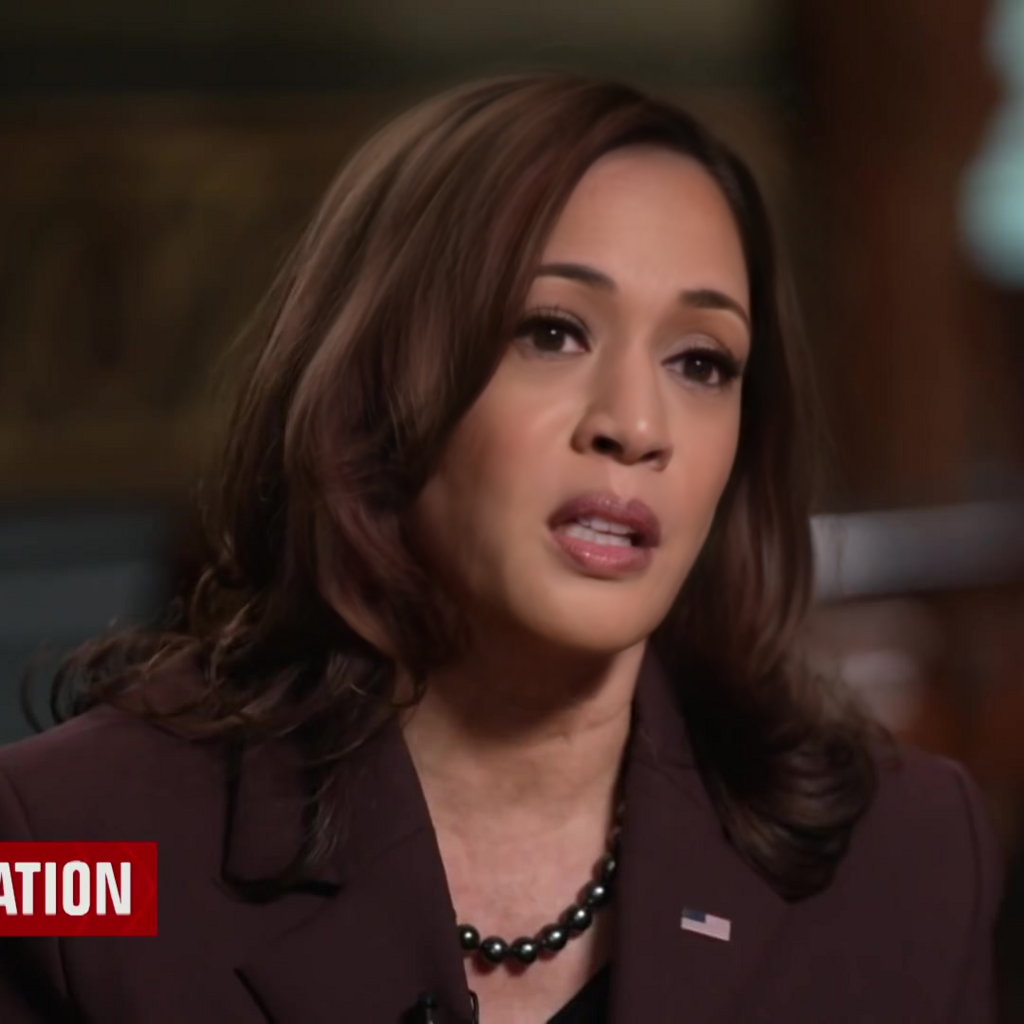} &
        \includegraphics[width=0.095\textwidth]{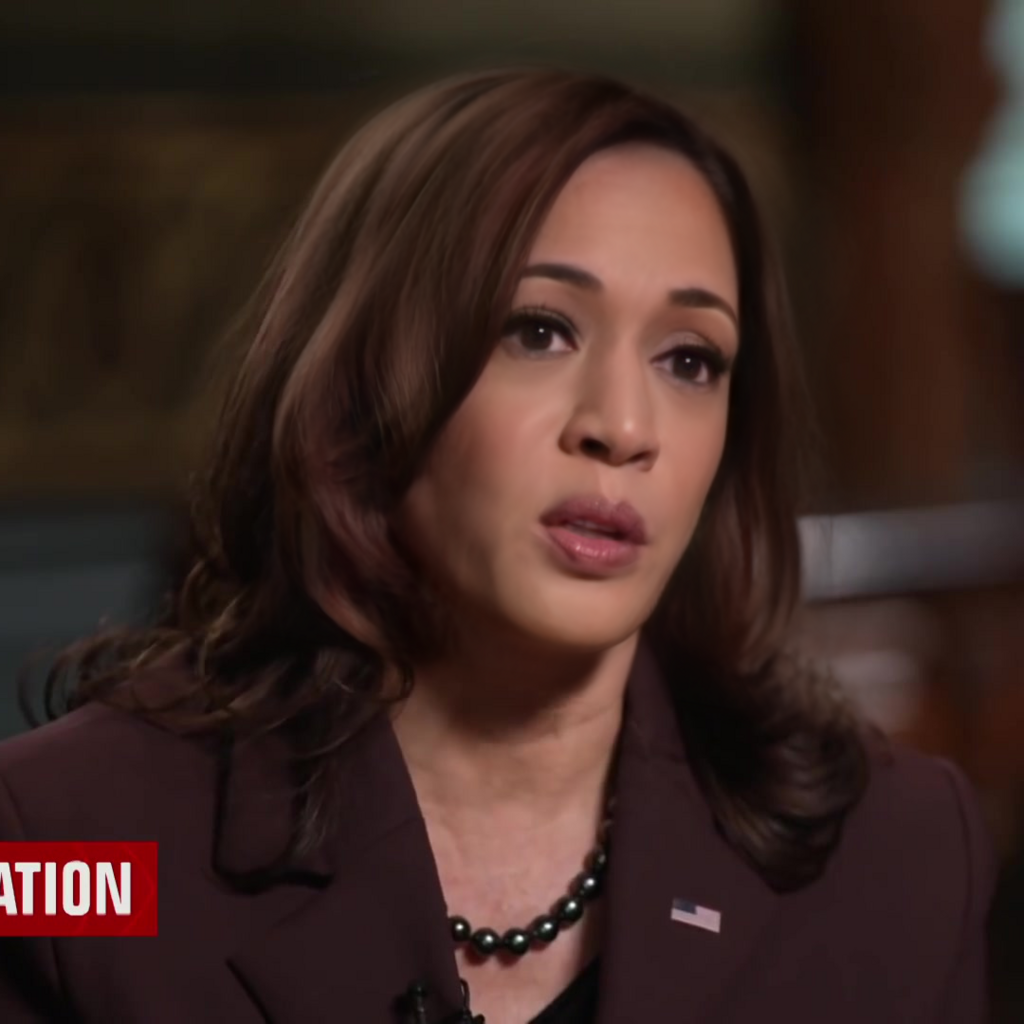} &
        \includegraphics[width=0.095\textwidth]{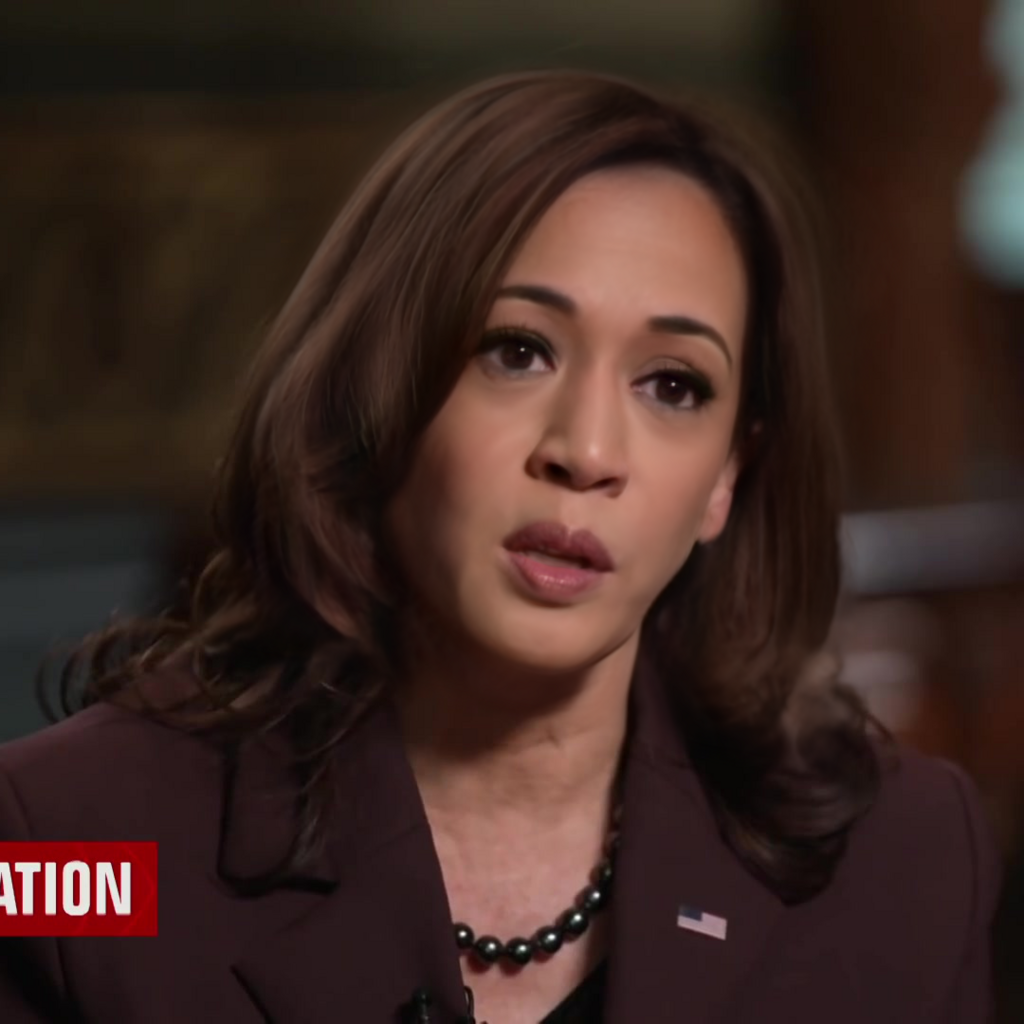} &
        \includegraphics[width=0.095\textwidth]{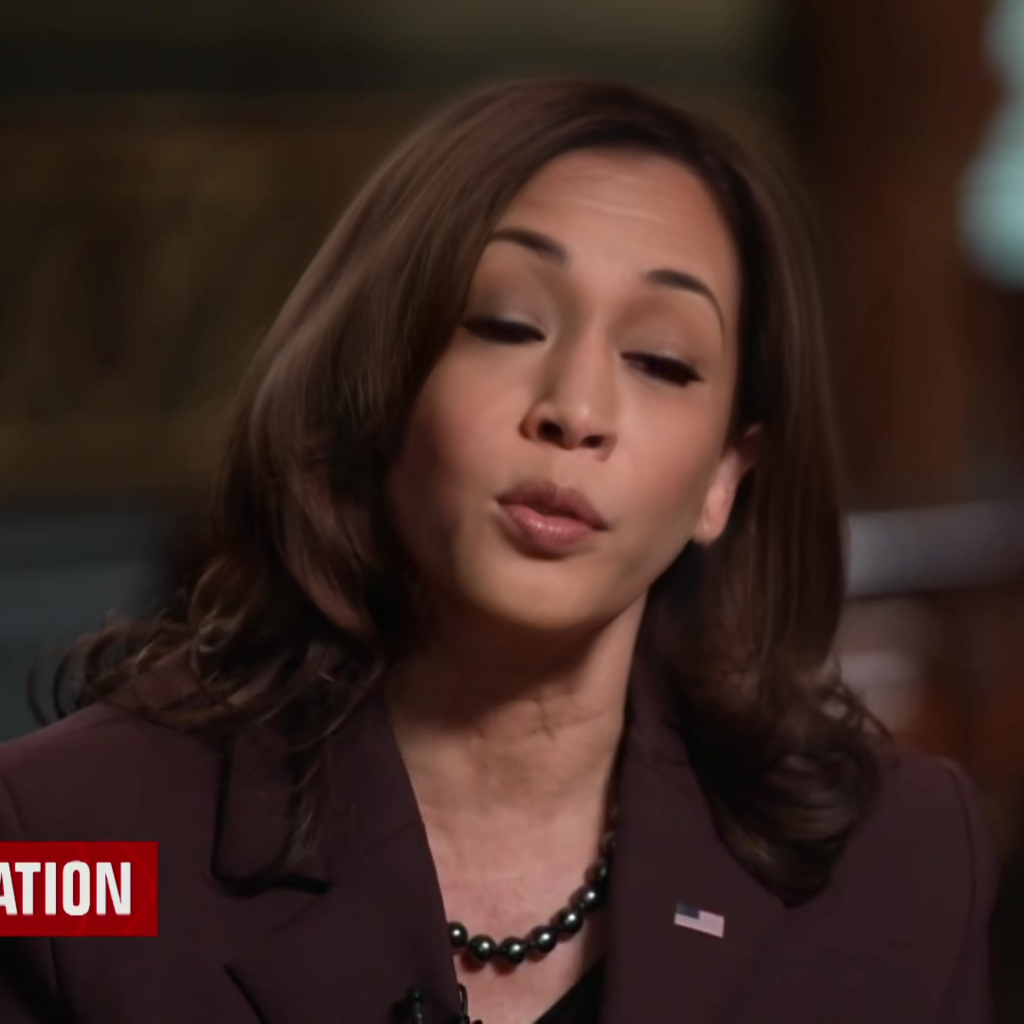} &
        \includegraphics[width=0.095\textwidth]{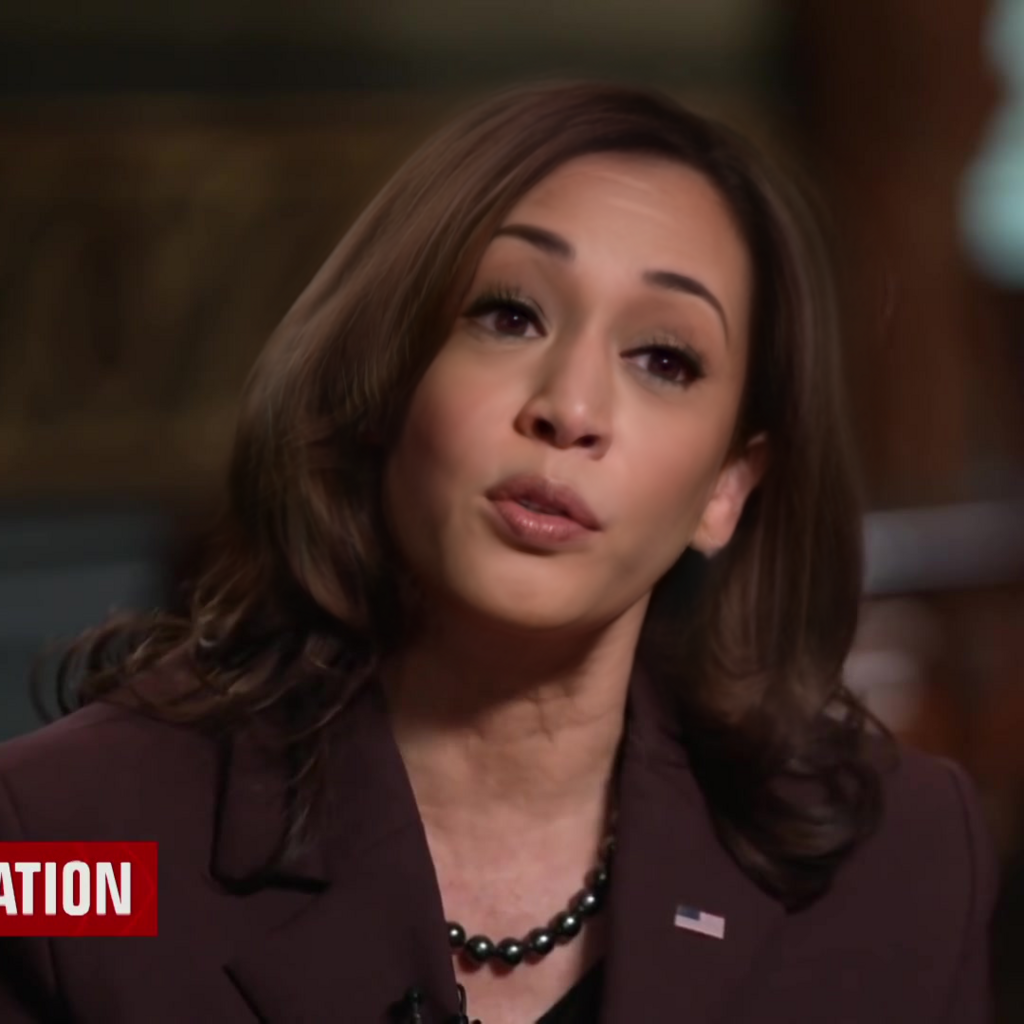} &
        \includegraphics[width=0.095\textwidth]{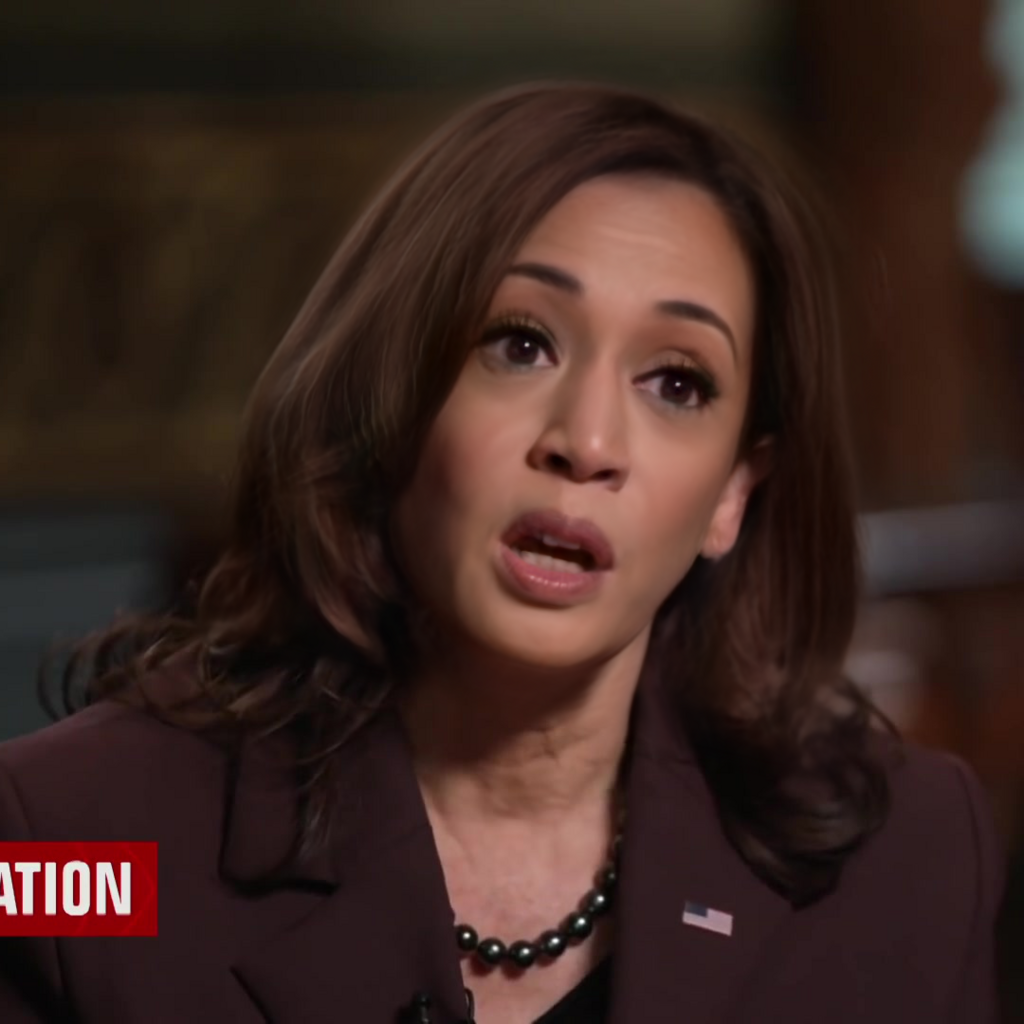} &
        \includegraphics[width=0.095\textwidth]{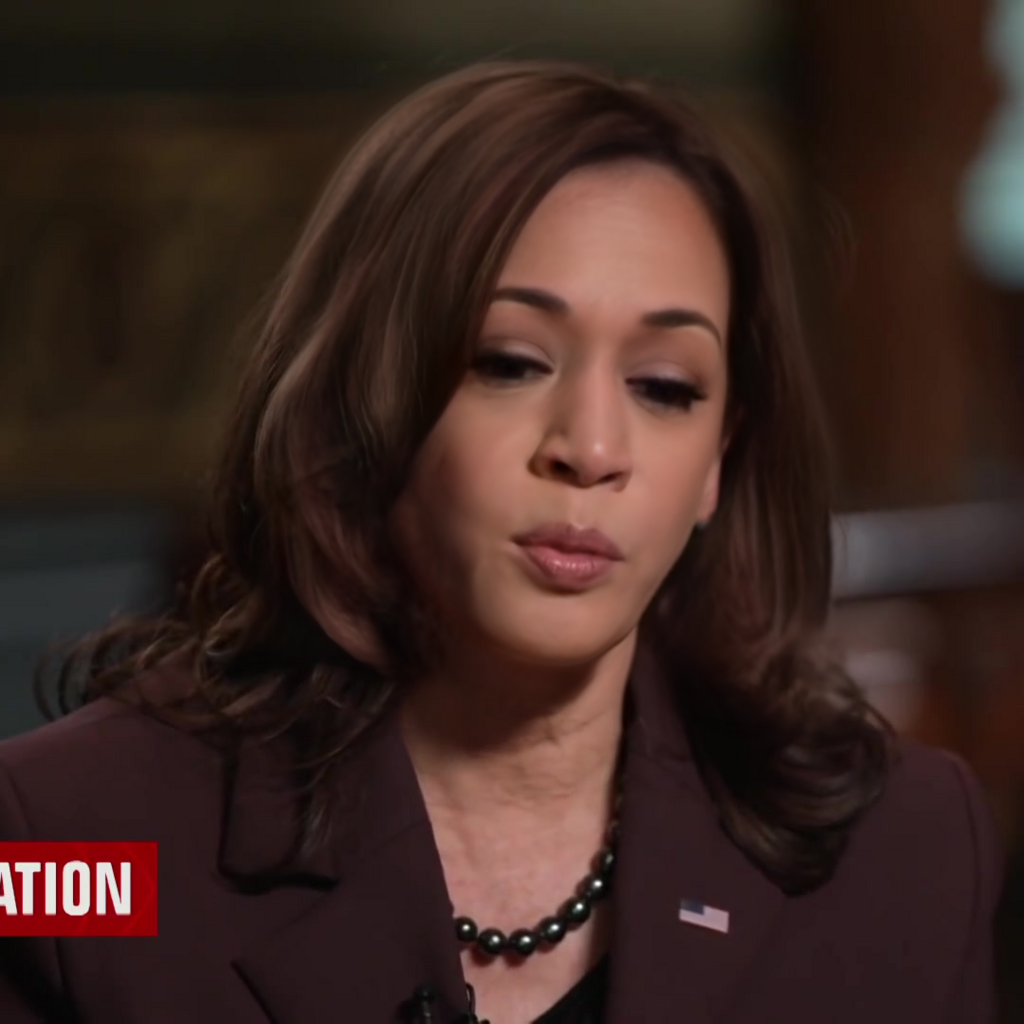} &
        \includegraphics[width=0.095\textwidth]{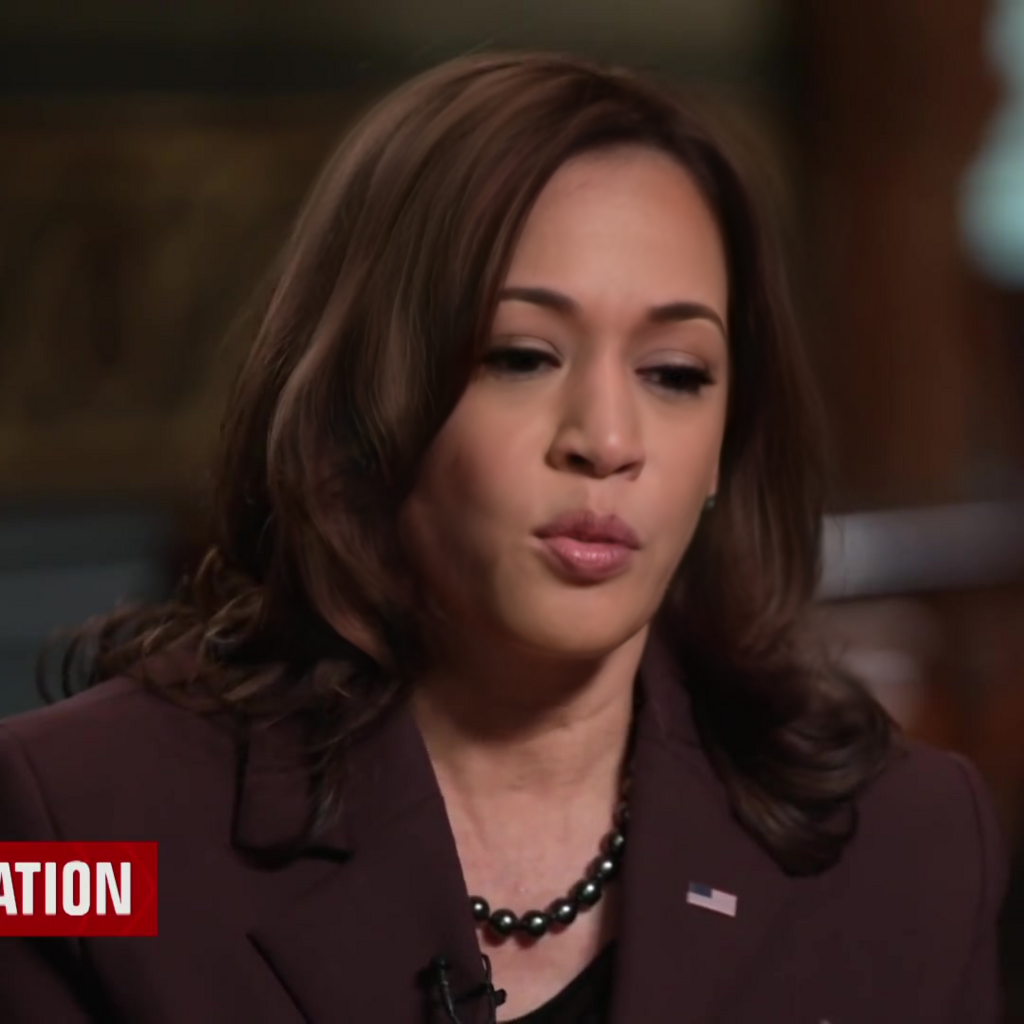} &
        \includegraphics[width=0.095\textwidth]{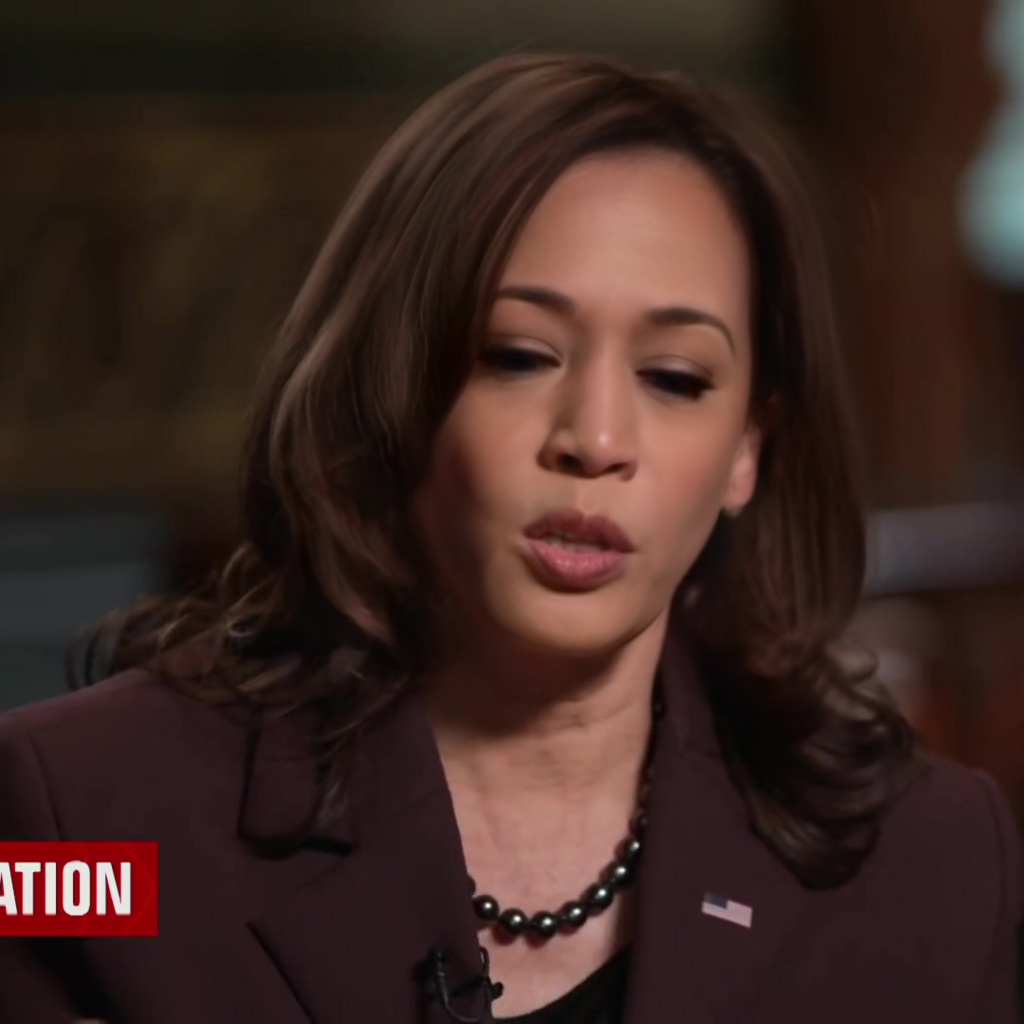} &
        \includegraphics[width=0.095\textwidth]{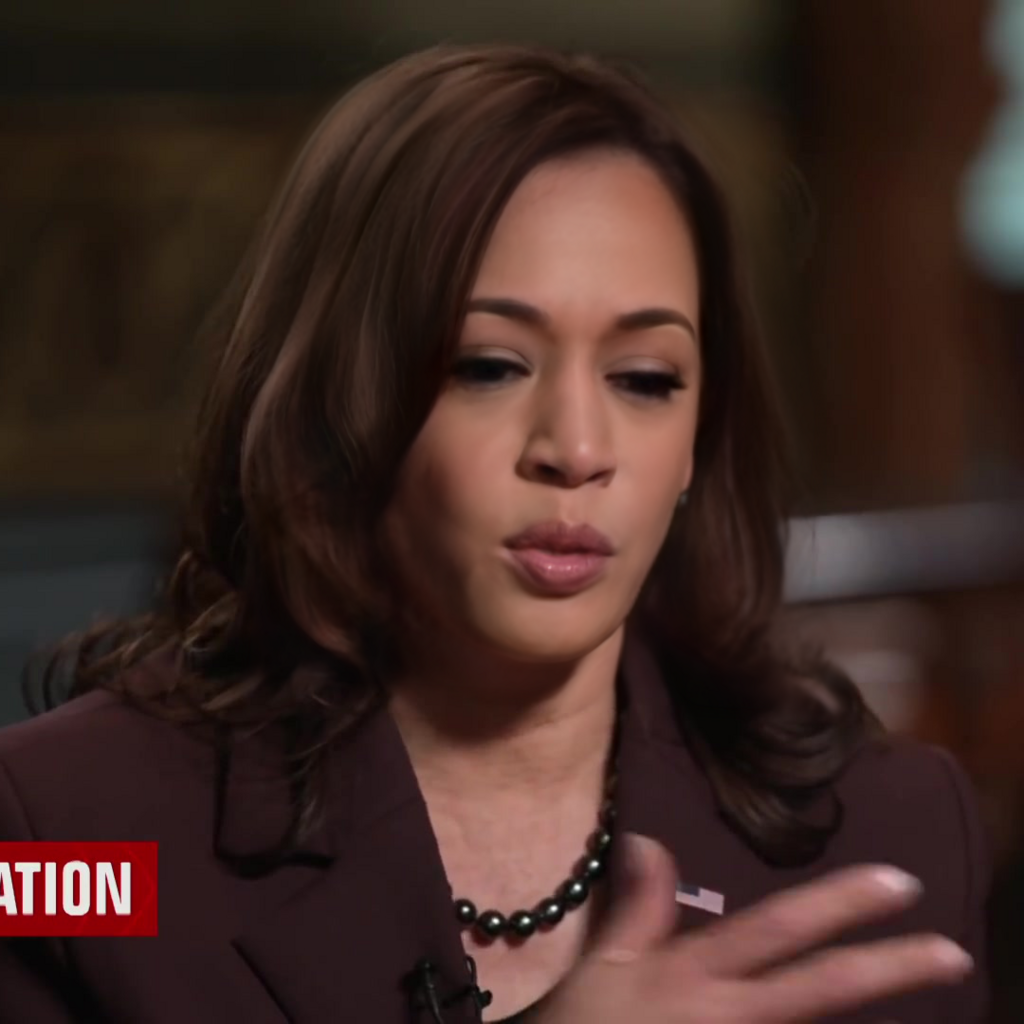} \\

        % \\

        %& Frame 1 & Frame 2 & Frame 3 & Frame 4 & Frame 1 & Frame 2 & Frame 3 & Frame 4
        
    \end{tabular}
    
    }
    \vspace{-0.225cm}
    \caption{Additional Video editing results using our proposed pipeline. For most modifications, our stitching framework can handle more challenging cases such as long hair.}
    %\vspace{-0.225cm}
    \label{fig:harris}
\end{figure*}

%% file: resources/figures/comparison.tex
\begin{figure}
\setlength{\tabcolsep}{0.5pt}
    \centering
    { \small 
\begin{tabular}{lcccc}
\rotatebox[origin=t]{90}{Original} &
\raisebox{-.42\totalheight}{\includegraphics[width=0.24\columnwidth]{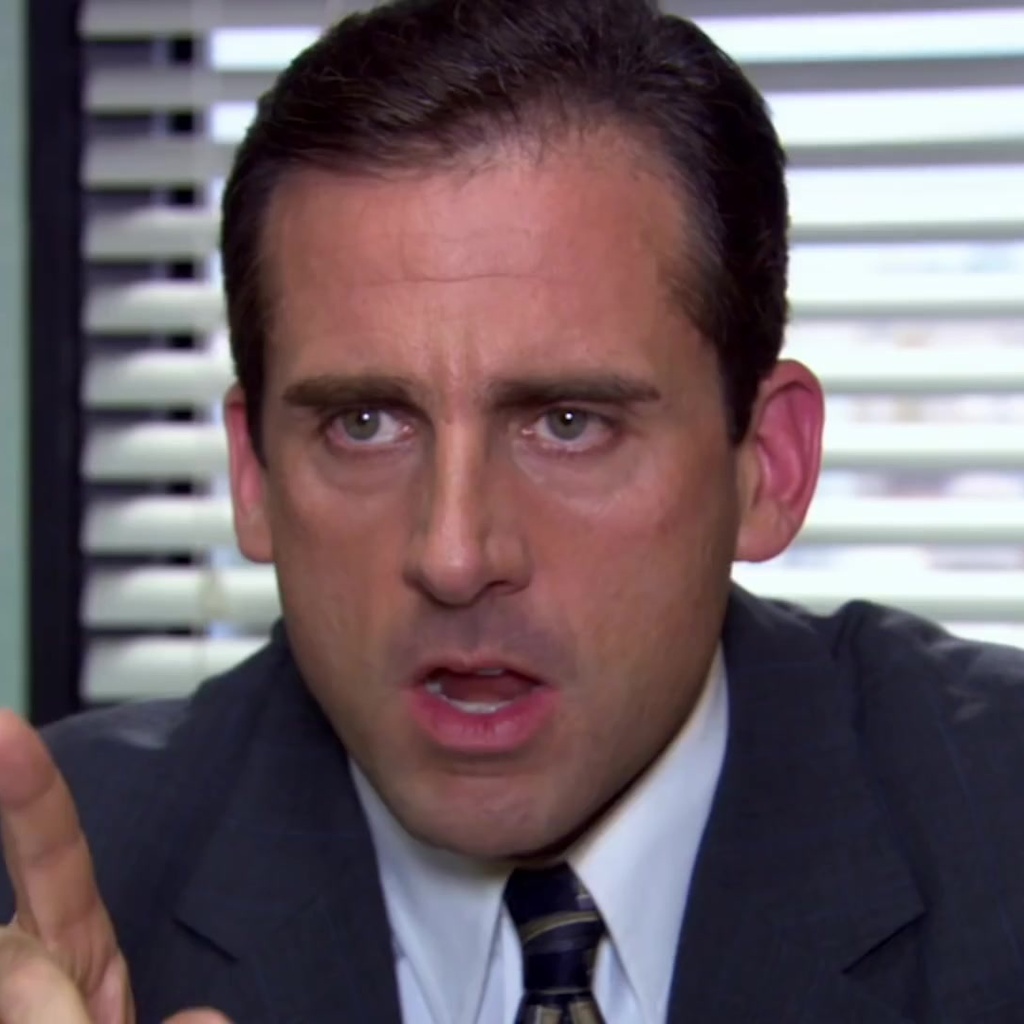}} &
\raisebox{-.42\totalheight}{\includegraphics[width=0.24\columnwidth]{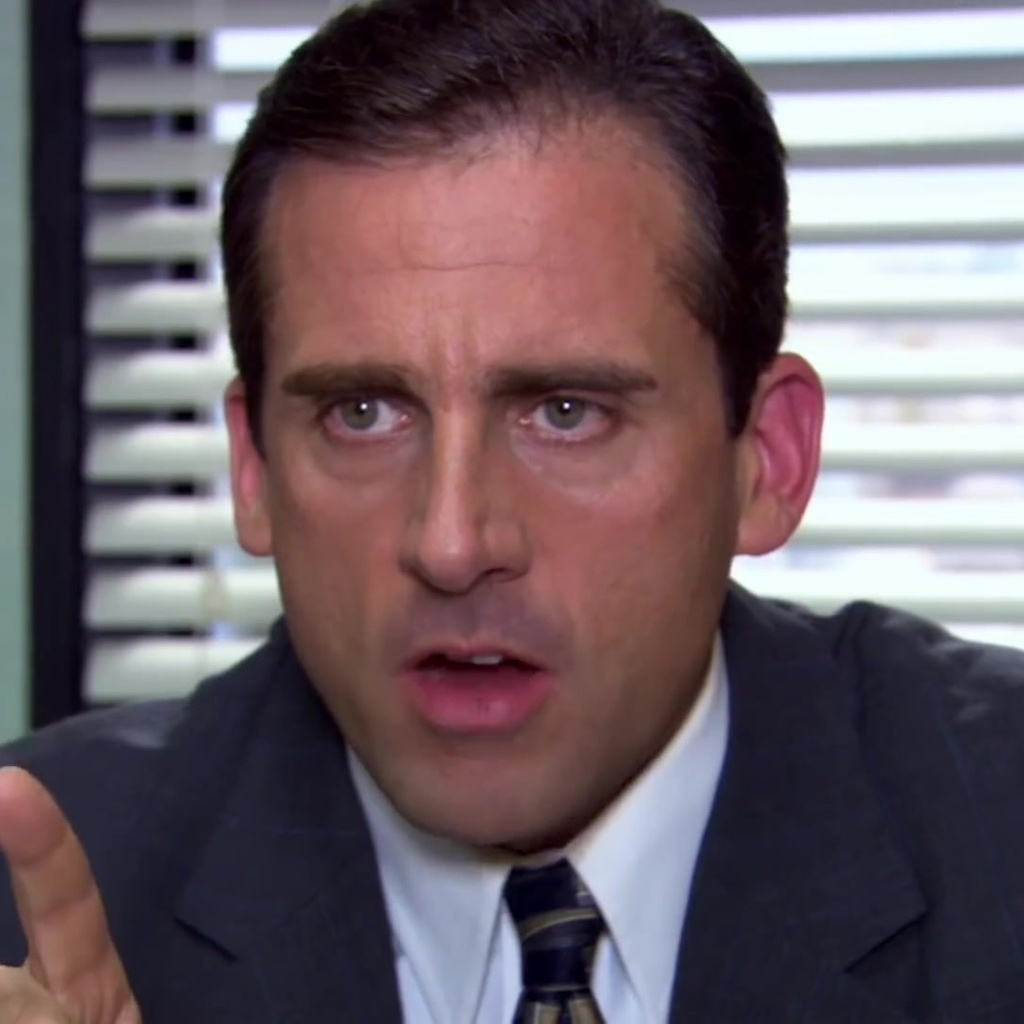}} &
\raisebox{-.42\totalheight}{\includegraphics[width=0.24\columnwidth]{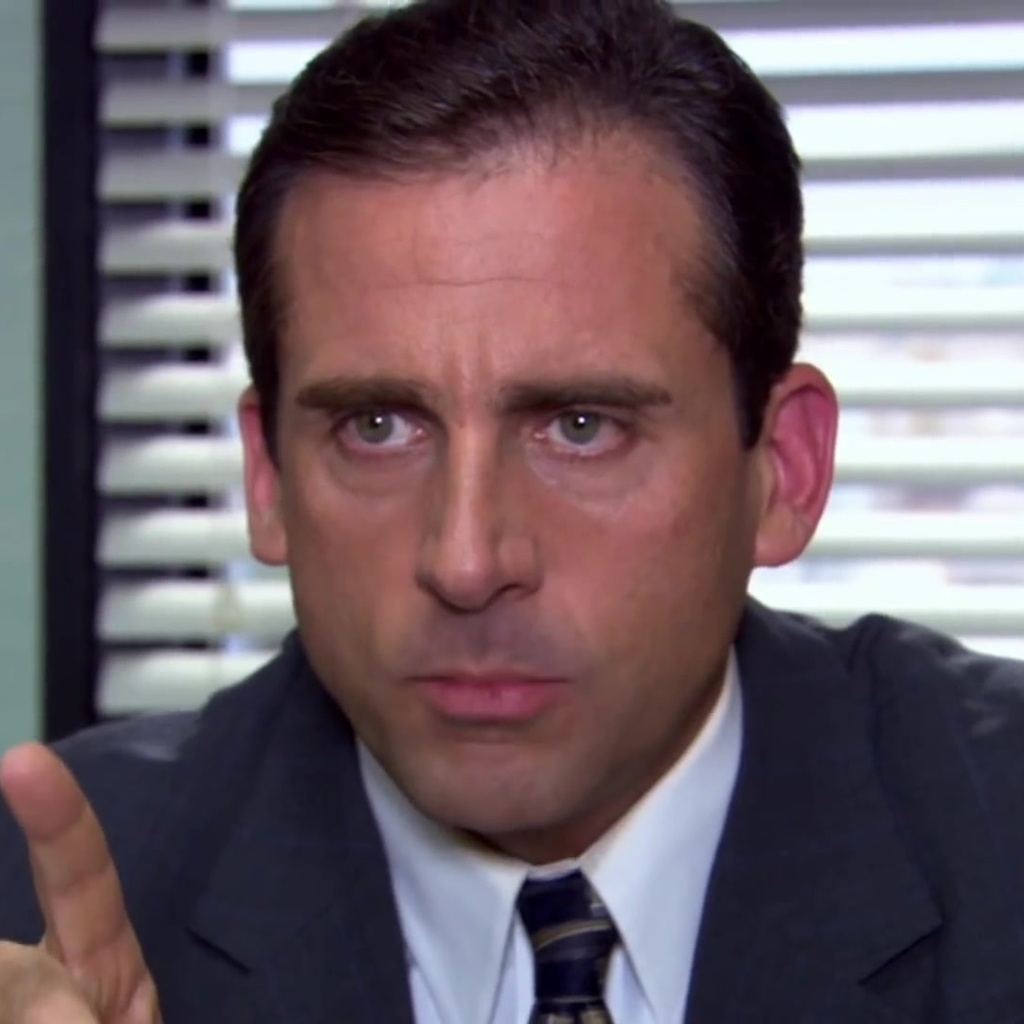}} & 
\raisebox{-.42\totalheight}{\includegraphics[width=0.24\columnwidth]{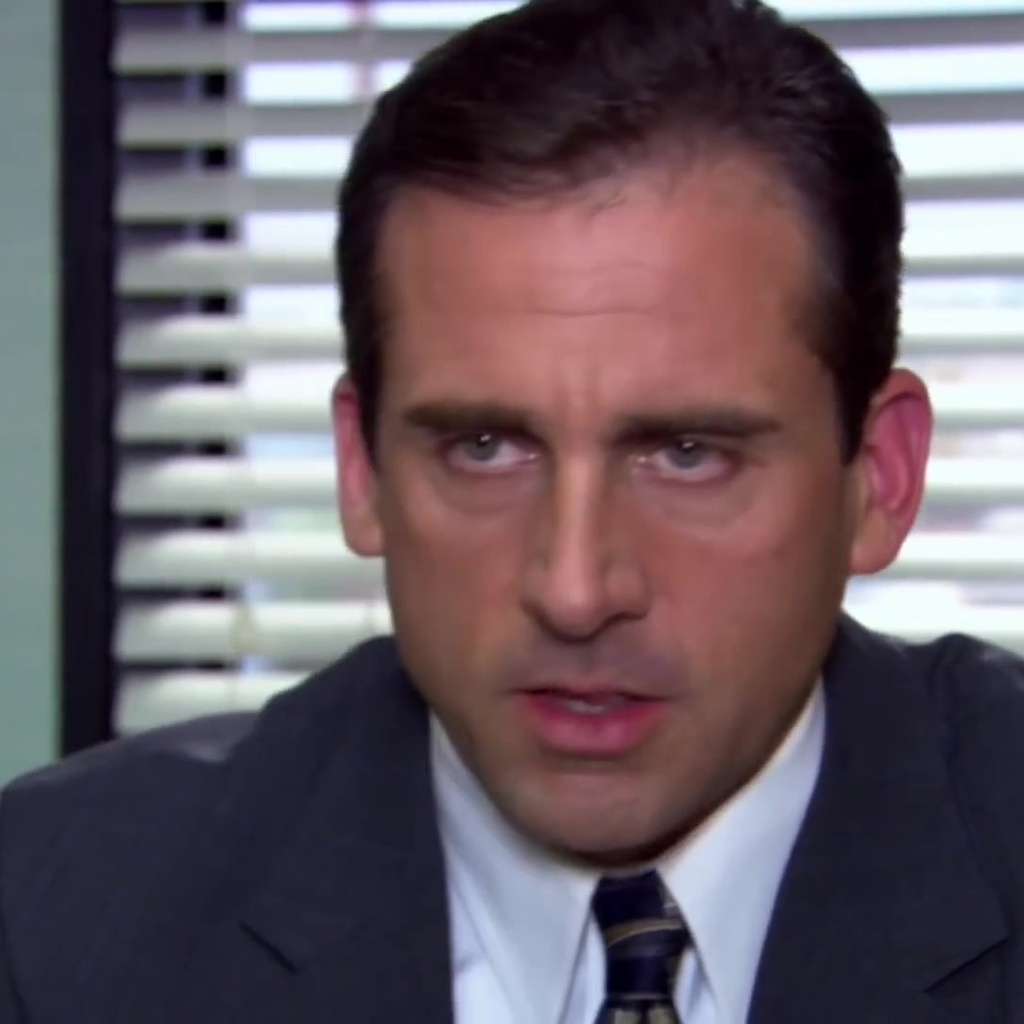}} \\
\noalign{\vskip .5mm}
\rotatebox[origin=t]{90}{Yao et al~\cite{yao2021latent}} &
\raisebox{-.42\totalheight}{\includegraphics[width=0.24\columnwidth]{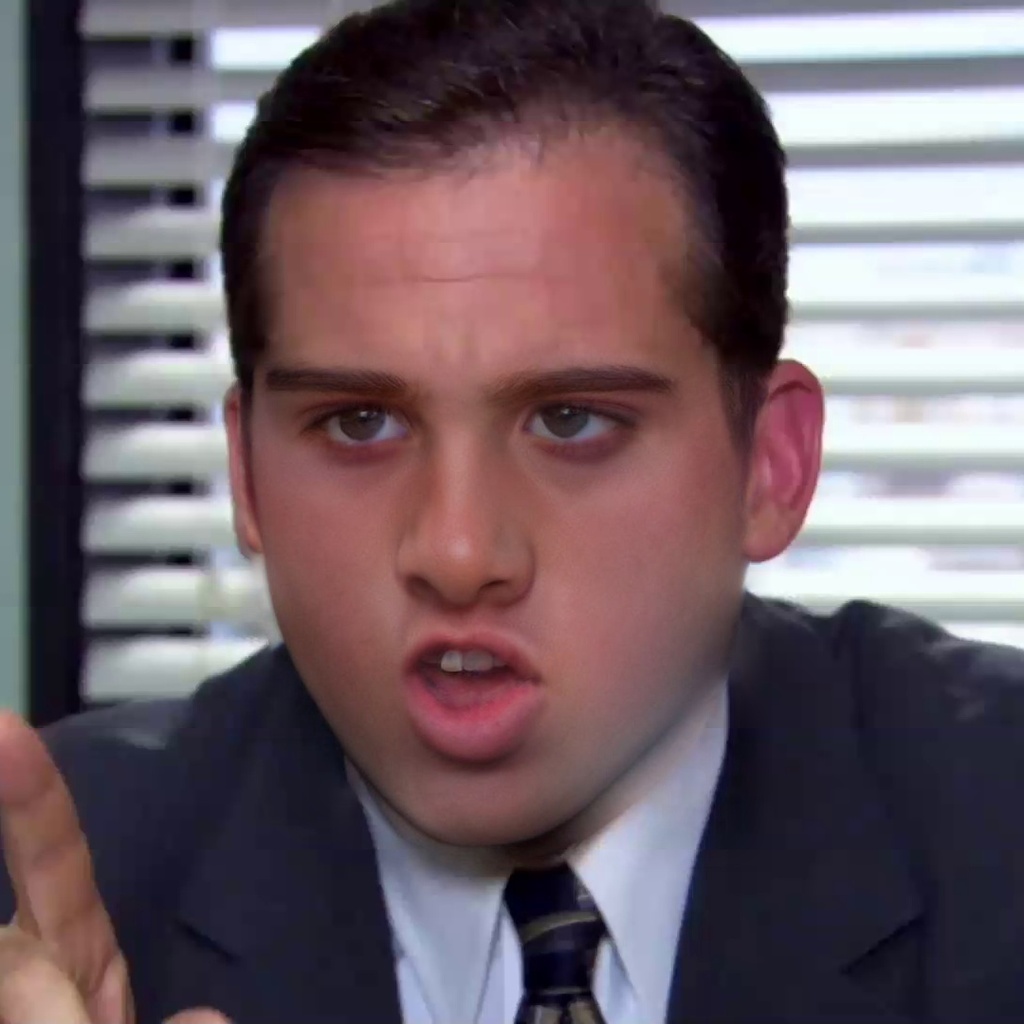}} &
\raisebox{-.42\totalheight}{\includegraphics[width=0.24\columnwidth]{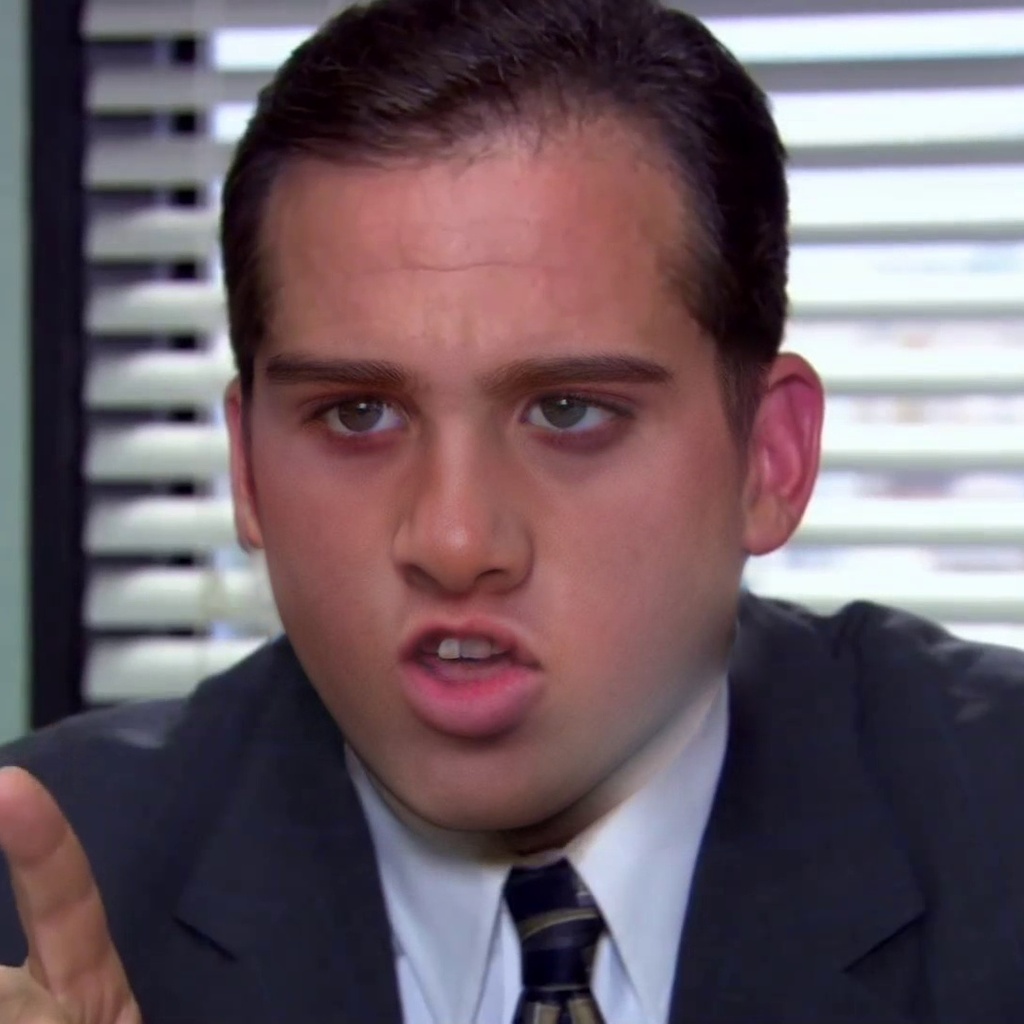}} &
\raisebox{-.42\totalheight}{\includegraphics[width=0.24\columnwidth]{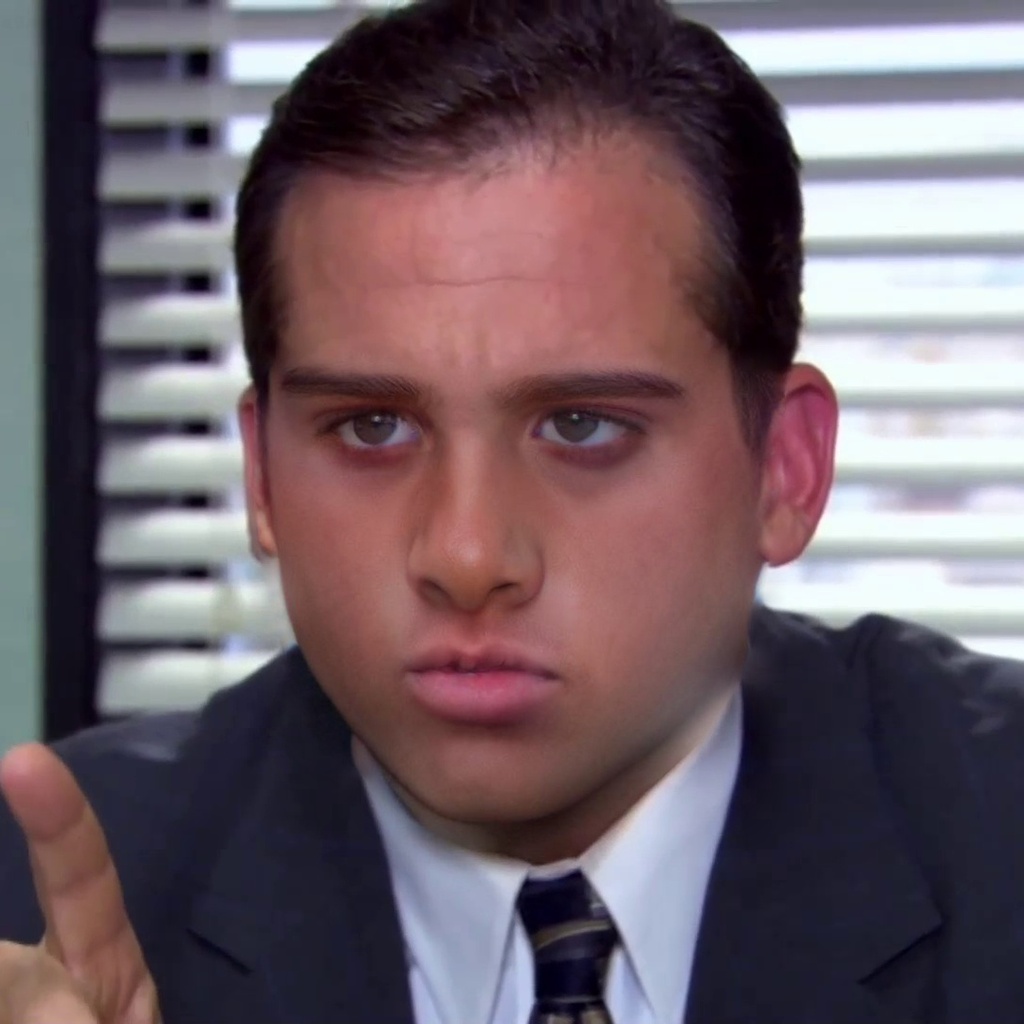}} &
\raisebox{-.42\totalheight}{\includegraphics[width=0.24\columnwidth]{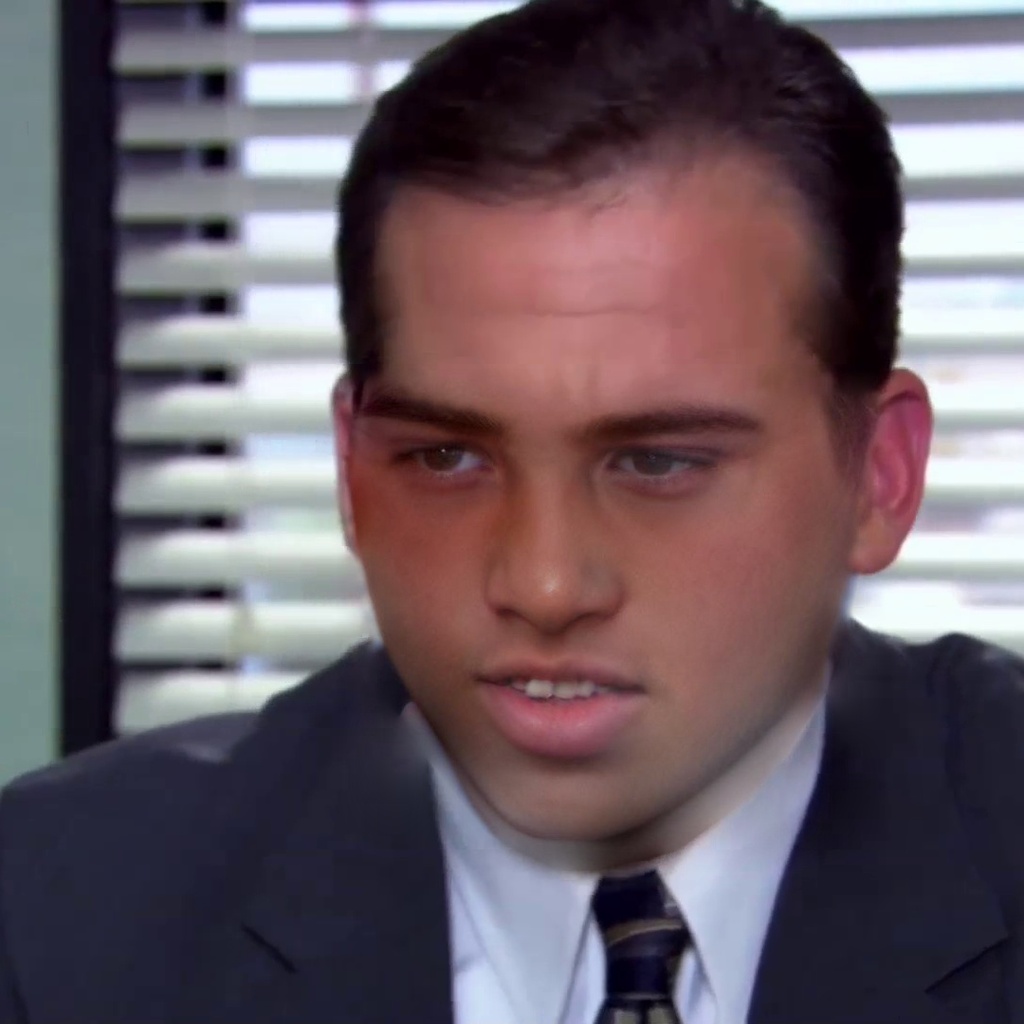}}\\
\noalign{\vskip .5mm}
\rotatebox[origin=t]{90}{PTI \shortcite{roich2021pivotal}} &
\raisebox{-.42\totalheight}{\includegraphics[width=0.24\columnwidth]{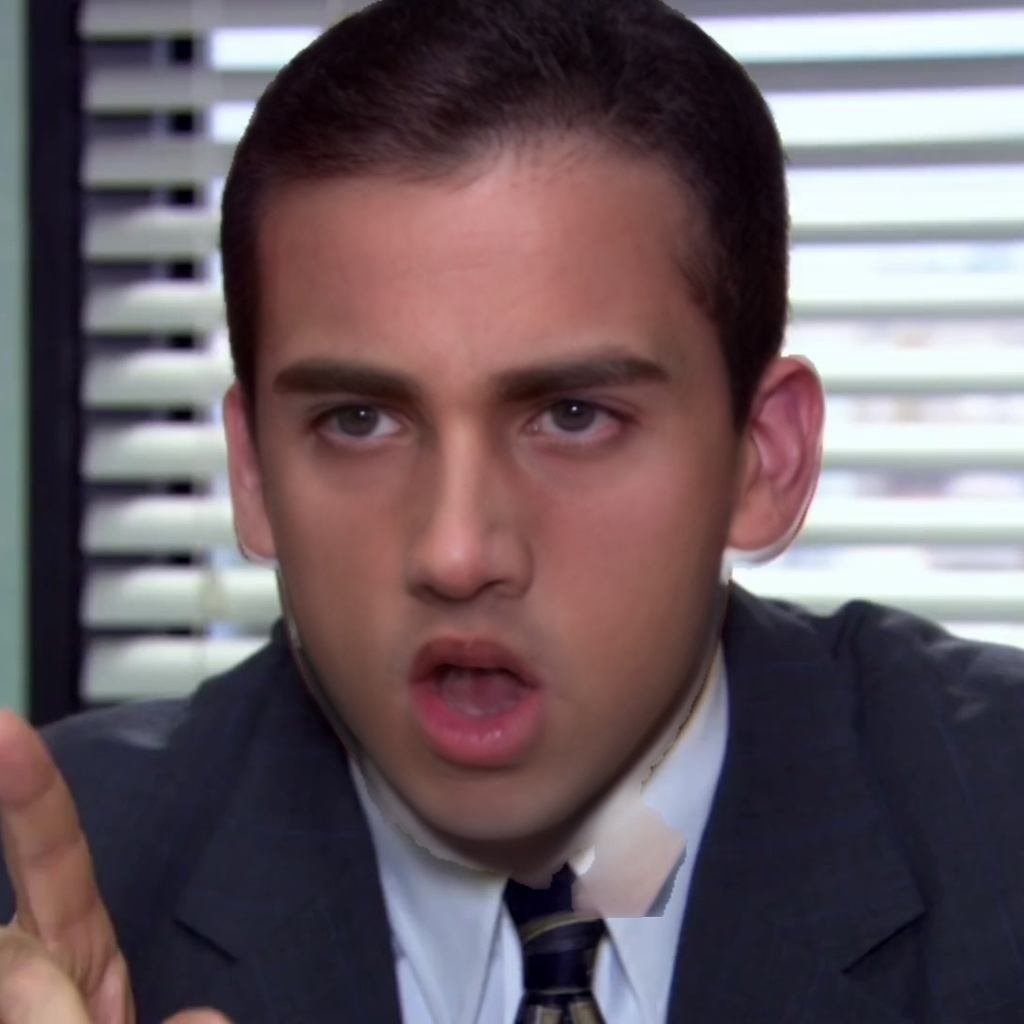}} &
\raisebox{-.42\totalheight}{\includegraphics[width=0.24\columnwidth]{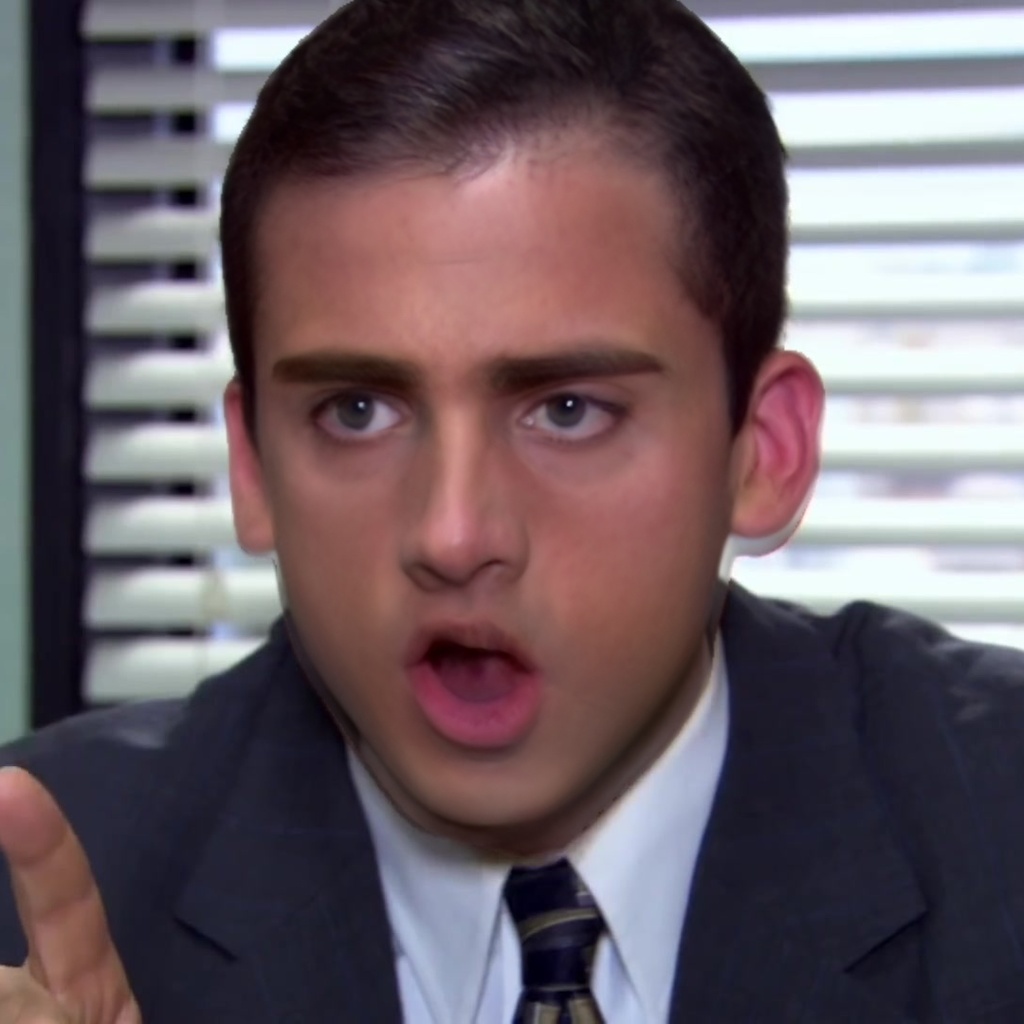}} &
\raisebox{-.42\totalheight}{\includegraphics[width=0.24\columnwidth]{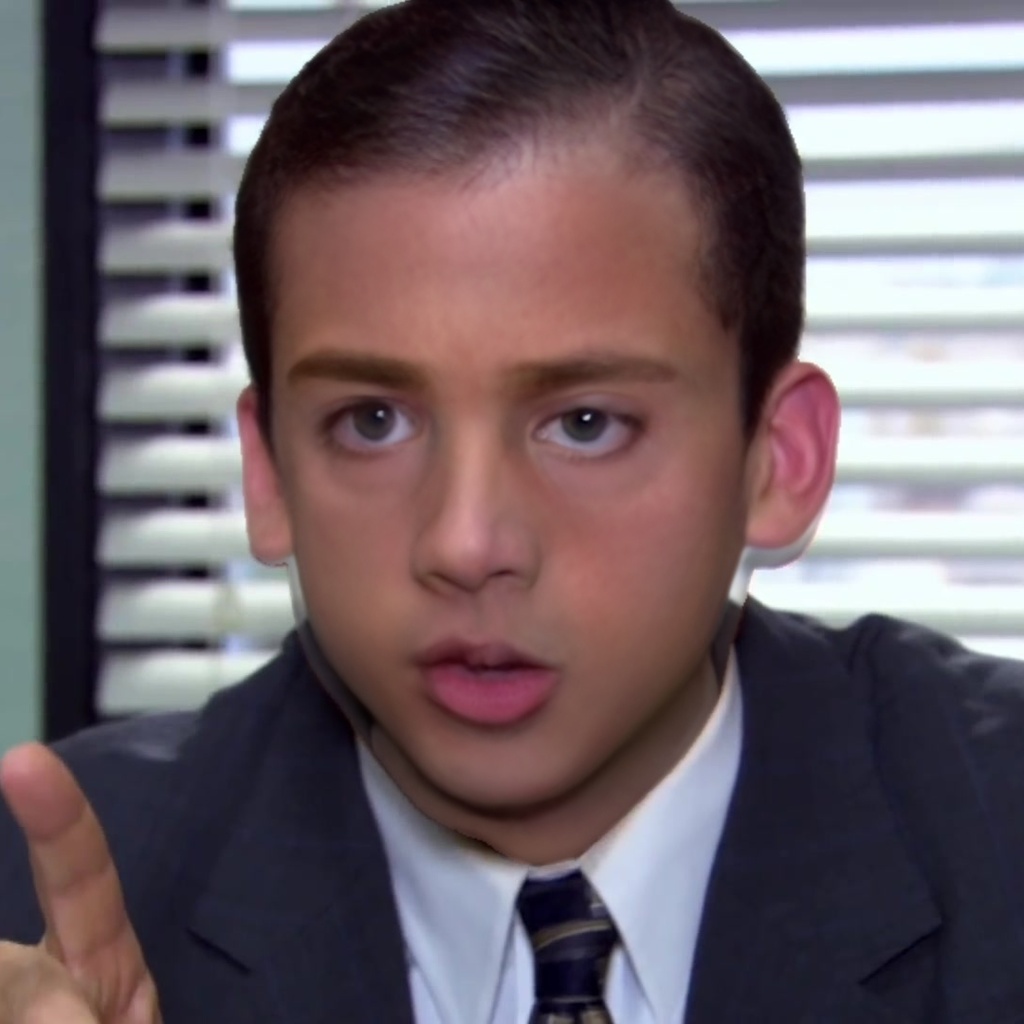}} &
\raisebox{-.42\totalheight}{\includegraphics[width=0.24\columnwidth]{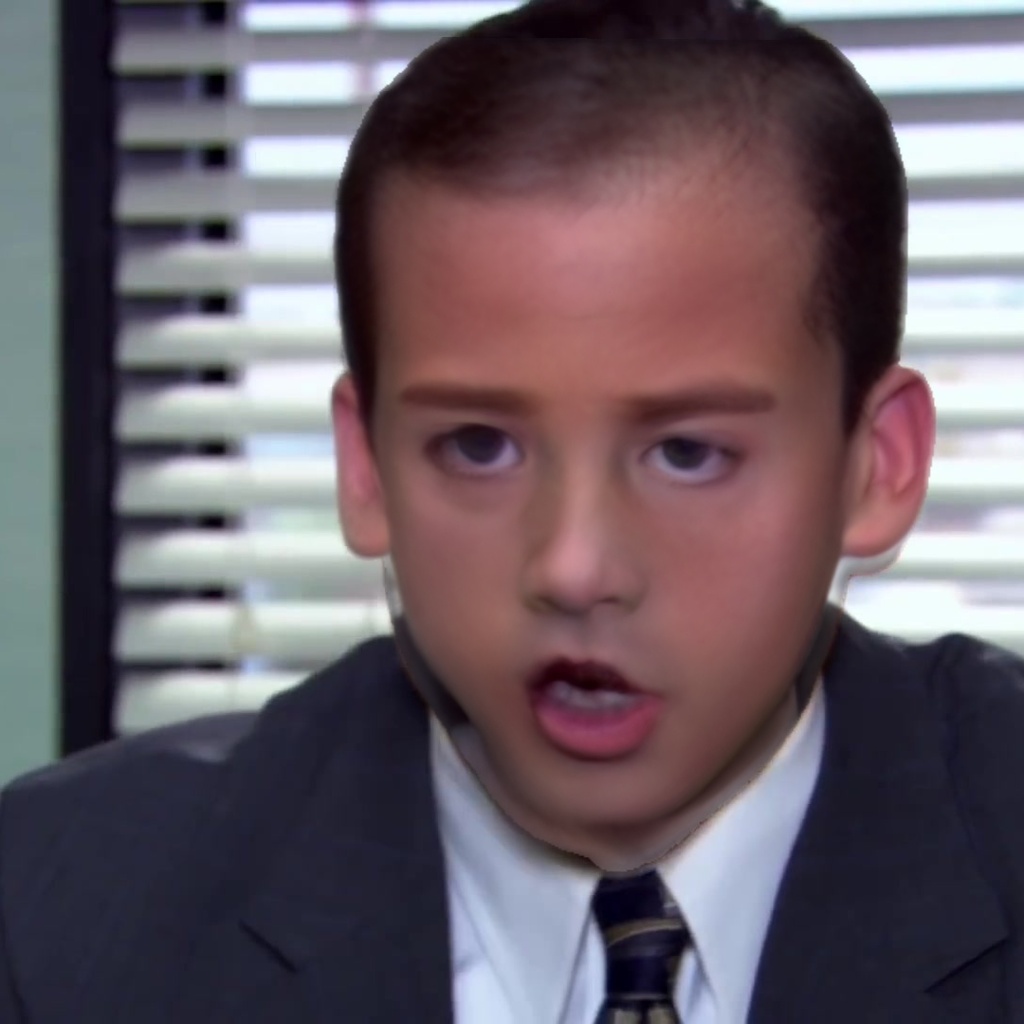}}\\
\noalign{\vskip .5mm}
\rotatebox[origin=t]{90}{Ours} &
\raisebox{-.42\totalheight}{\includegraphics[width=0.24\columnwidth]{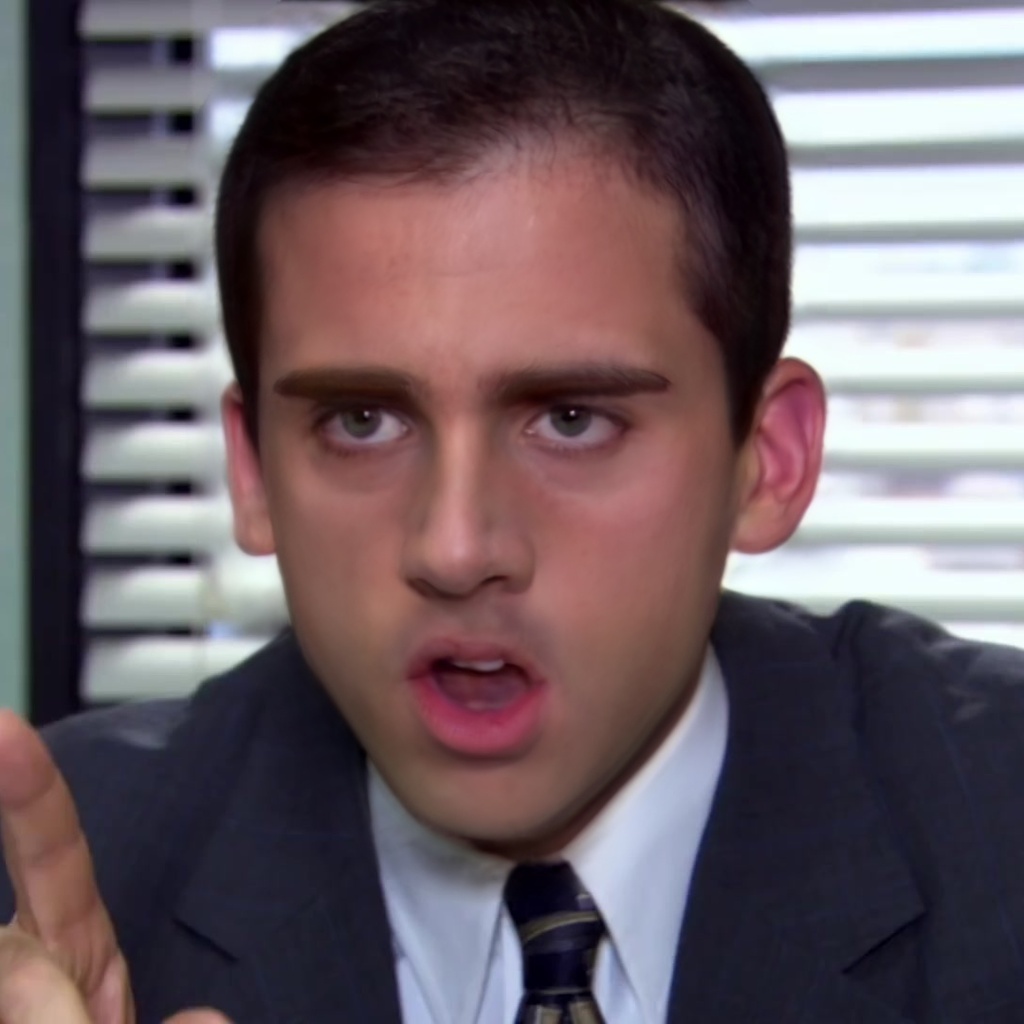}} &
\raisebox{-.42\totalheight}{\includegraphics[width=0.24\columnwidth]{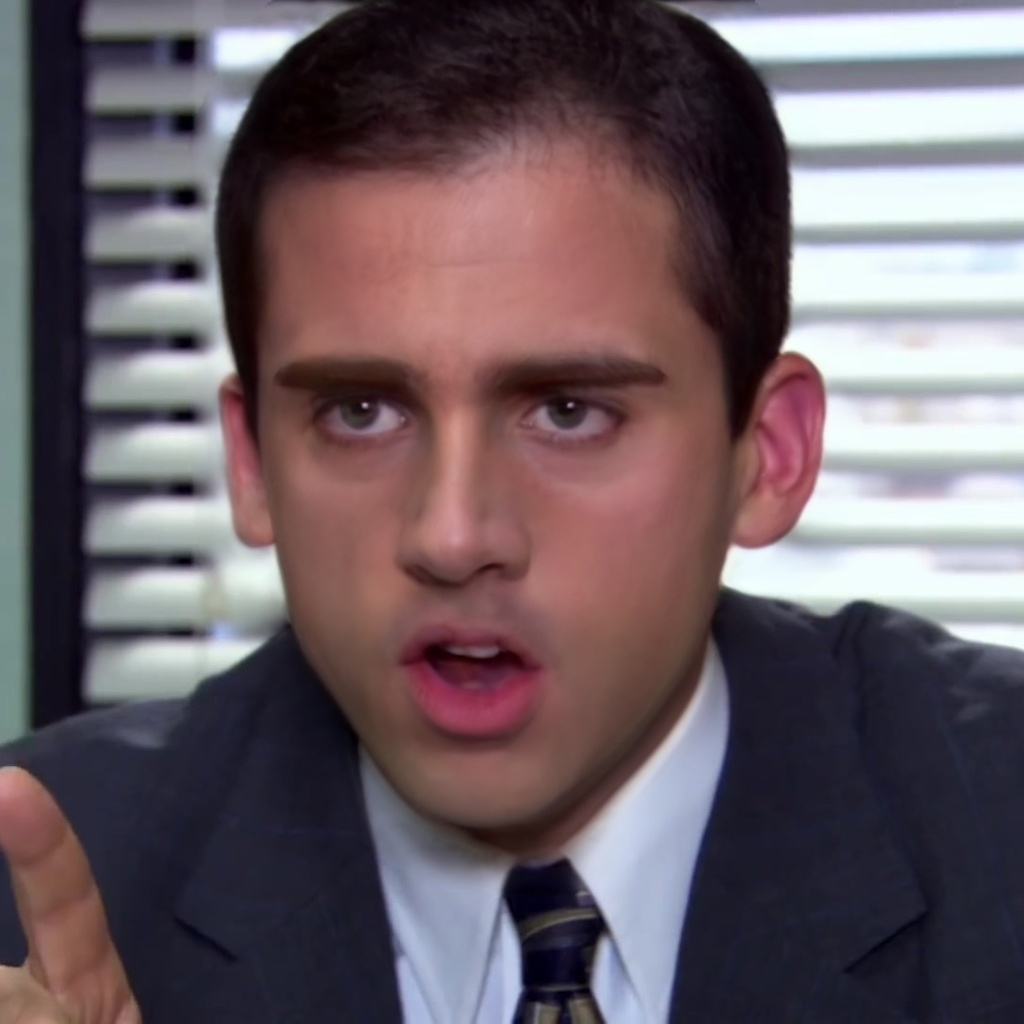}} &
\raisebox{-.42\totalheight}{\includegraphics[width=0.24\columnwidth]{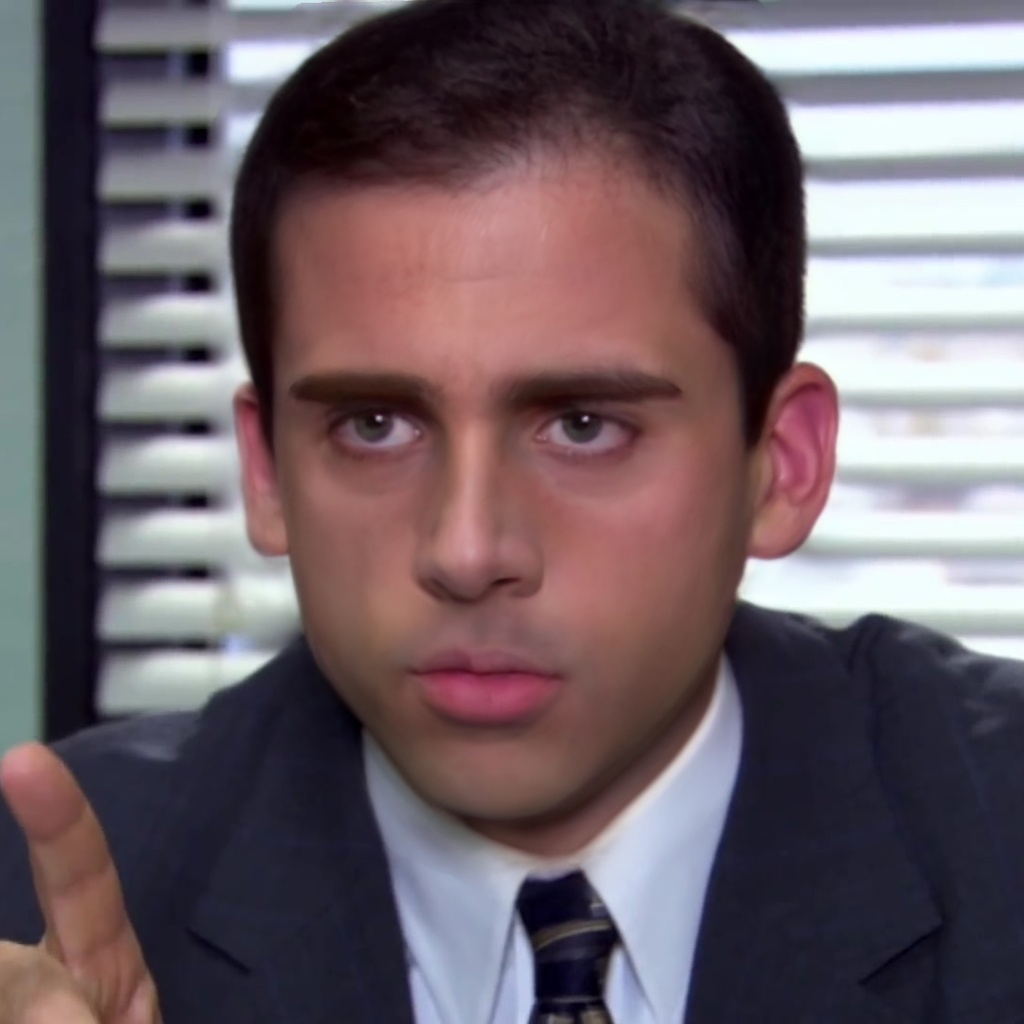}} &
\raisebox{-.42\totalheight}{\includegraphics[width=0.24\columnwidth]{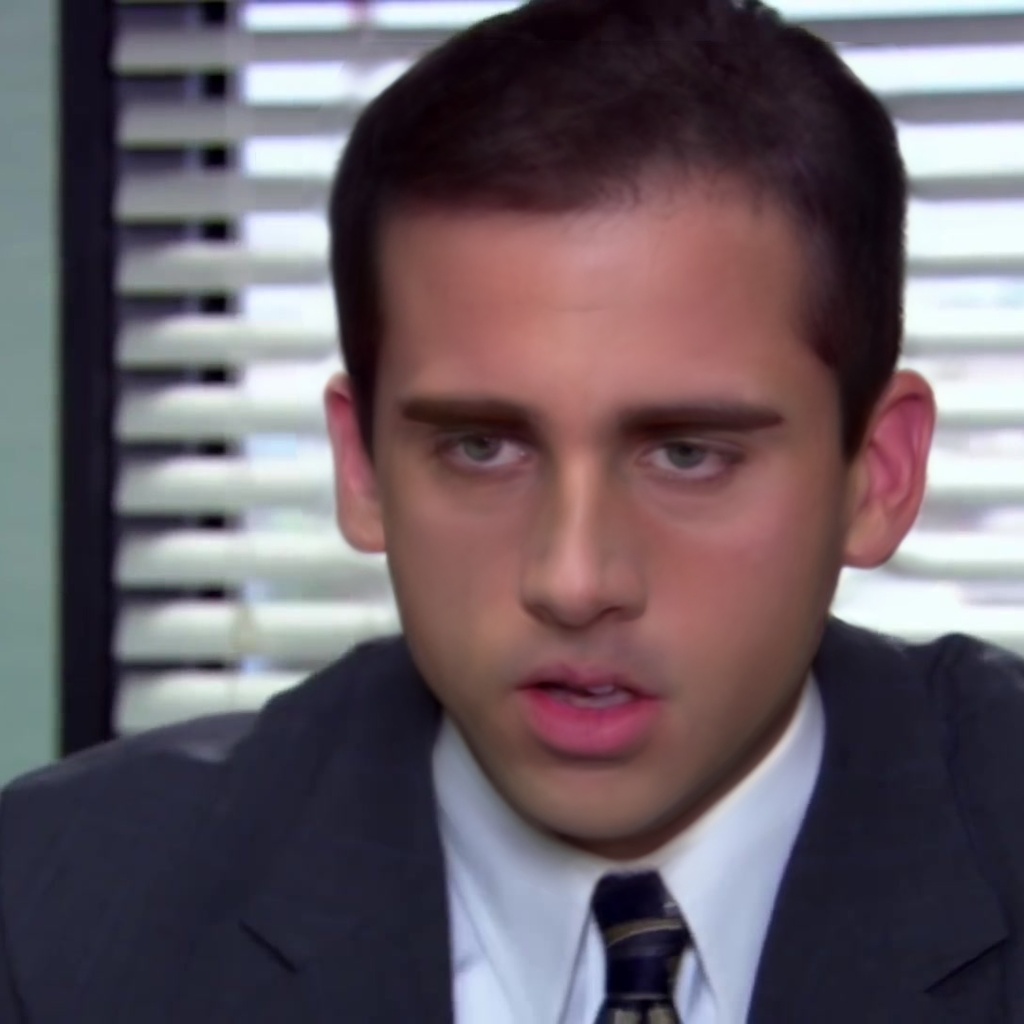}}\\
\end{tabular}
}

\caption{Visual comparison to alternative editing pipelines. Our method retains a higher degree of temporal consistency, produces realistic editing, and successfully mitigates blending-induced artifacts.}
\label{fig:comparison}
\end{figure}

%% file: resources/figures/ood.tex
\begin{figure}
    \centering
    \setlength{\belowcaptionskip}{-6pt}
    \setlength{\tabcolsep}{0.5pt}
    {
    \begin{tabular}{ccccc}
        
        \vspace{-0.0615cm}
        
        \raisebox{0.15in}{\rotatebox{90}{Original}} &
        \includegraphics[width=0.11\textwidth]{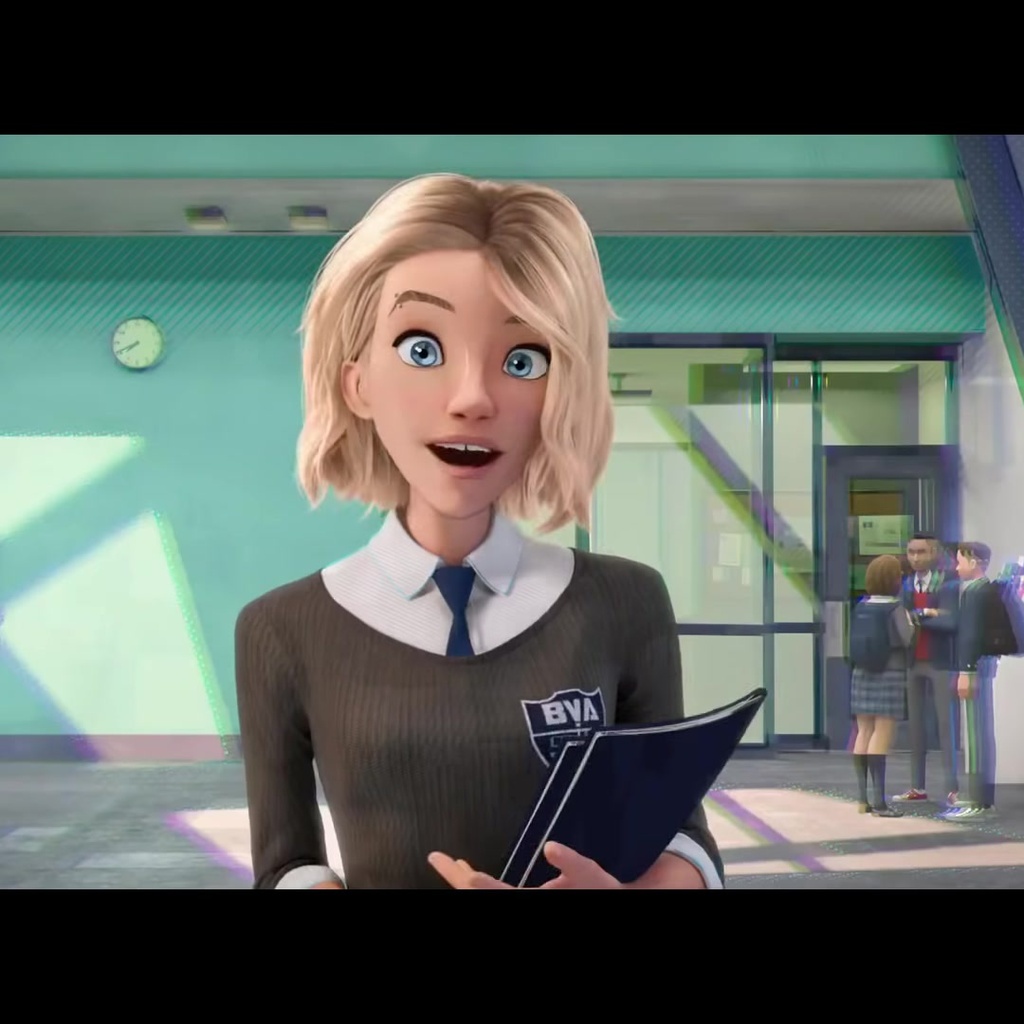} &
         \includegraphics[width=0.11\textwidth]{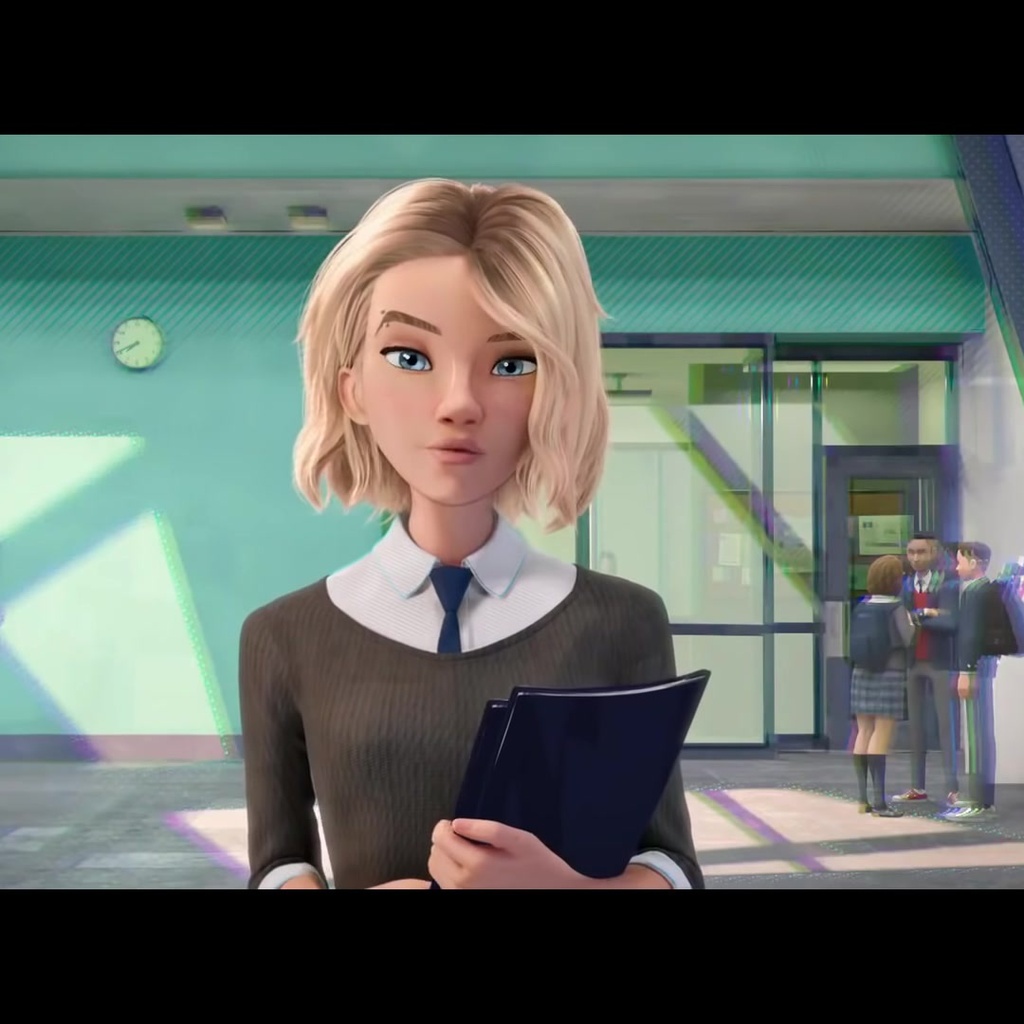} &
        \includegraphics[width=0.11\textwidth]{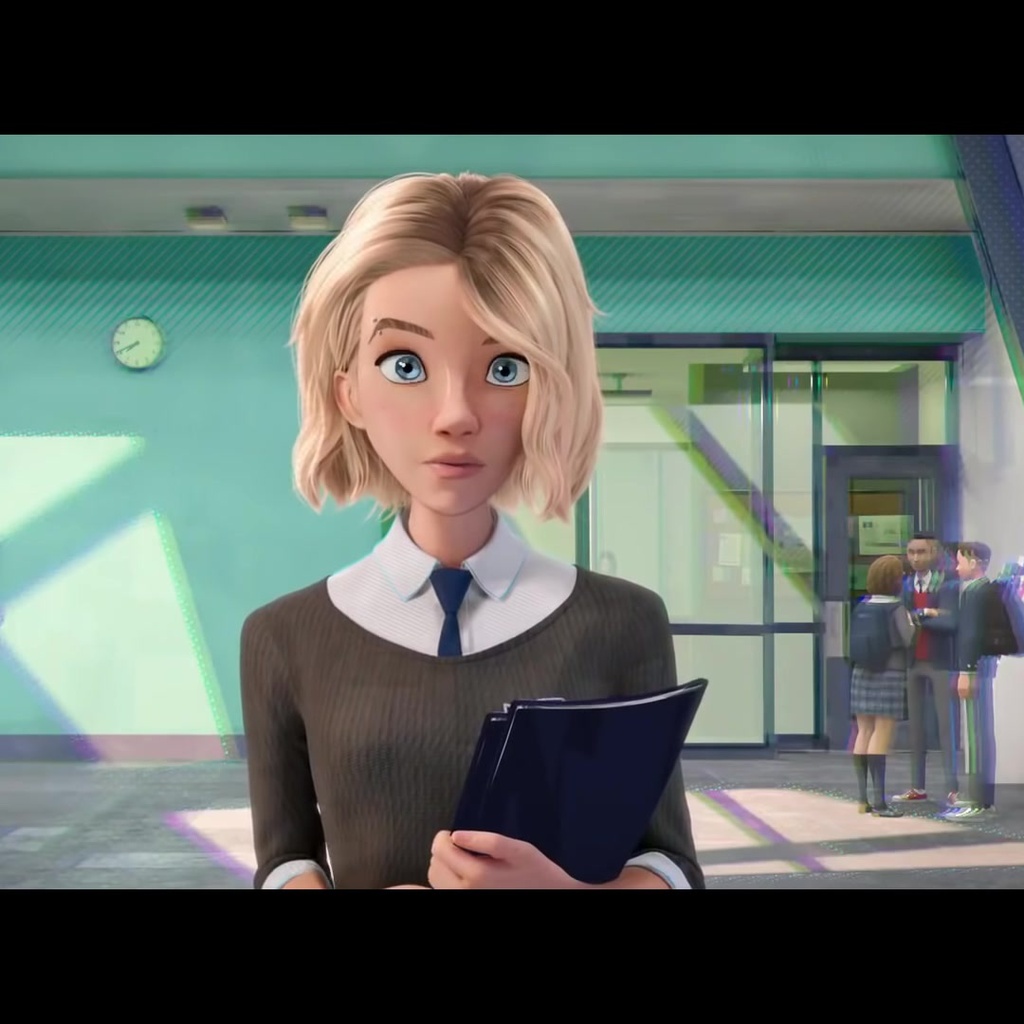} &                \includegraphics[width=0.11\textwidth]{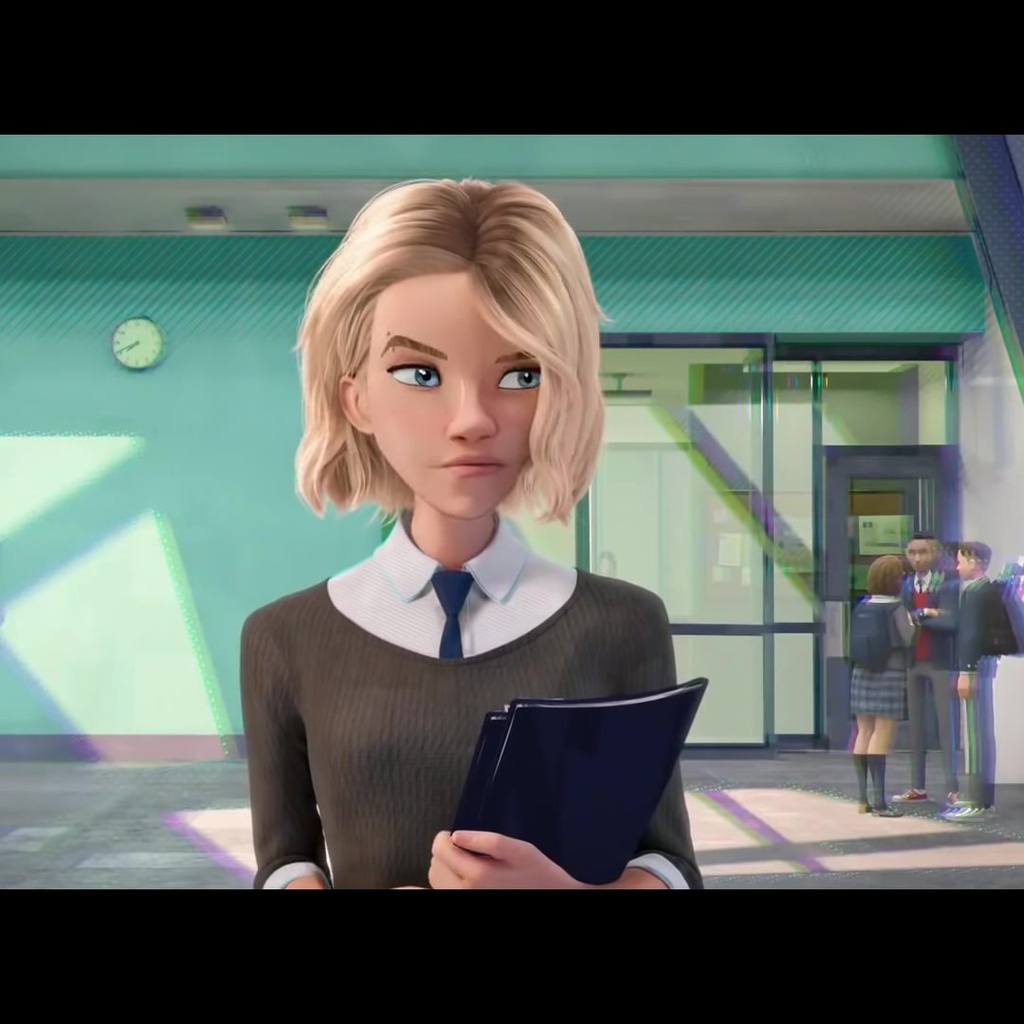} \\
        
        \raisebox{0.18in}{\rotatebox{90}{$+Smile$}} &
        \includegraphics[width=0.11\textwidth]{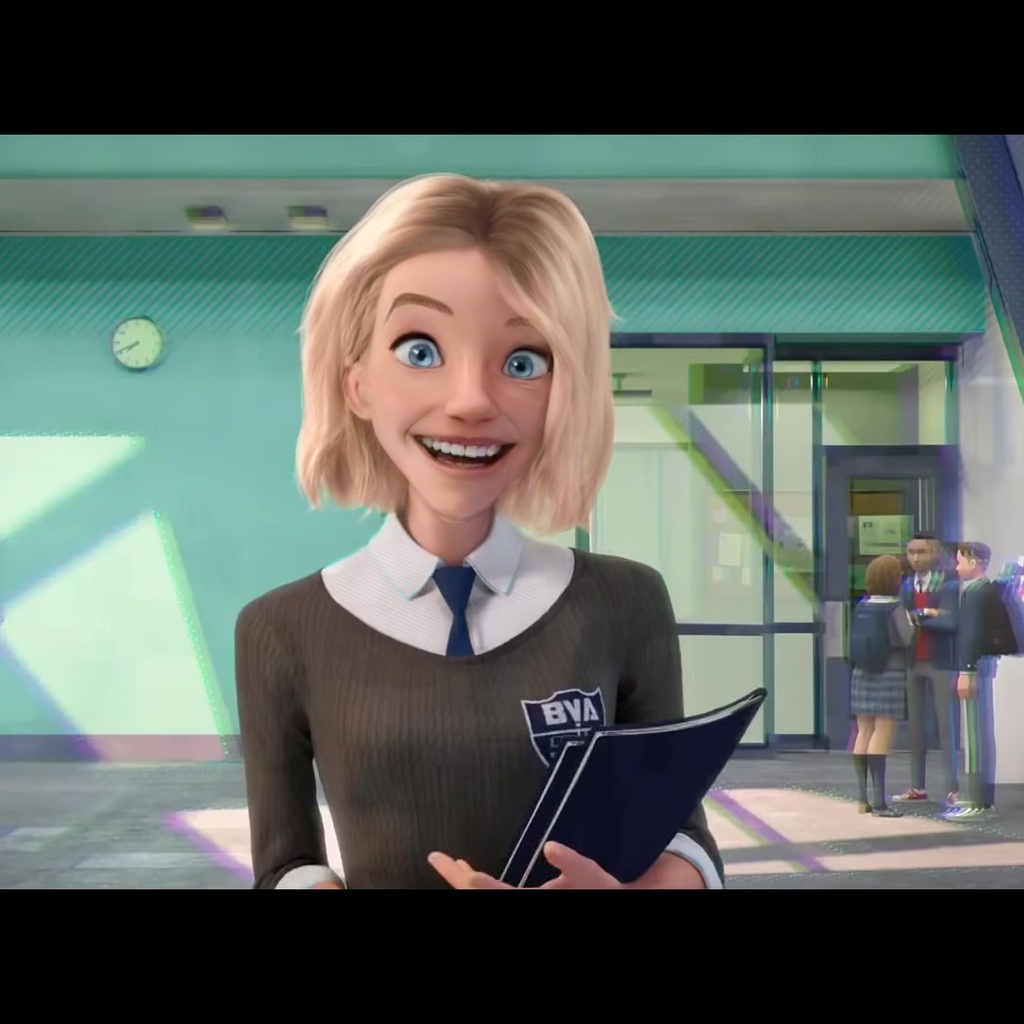} &
        \includegraphics[width=0.11\textwidth]{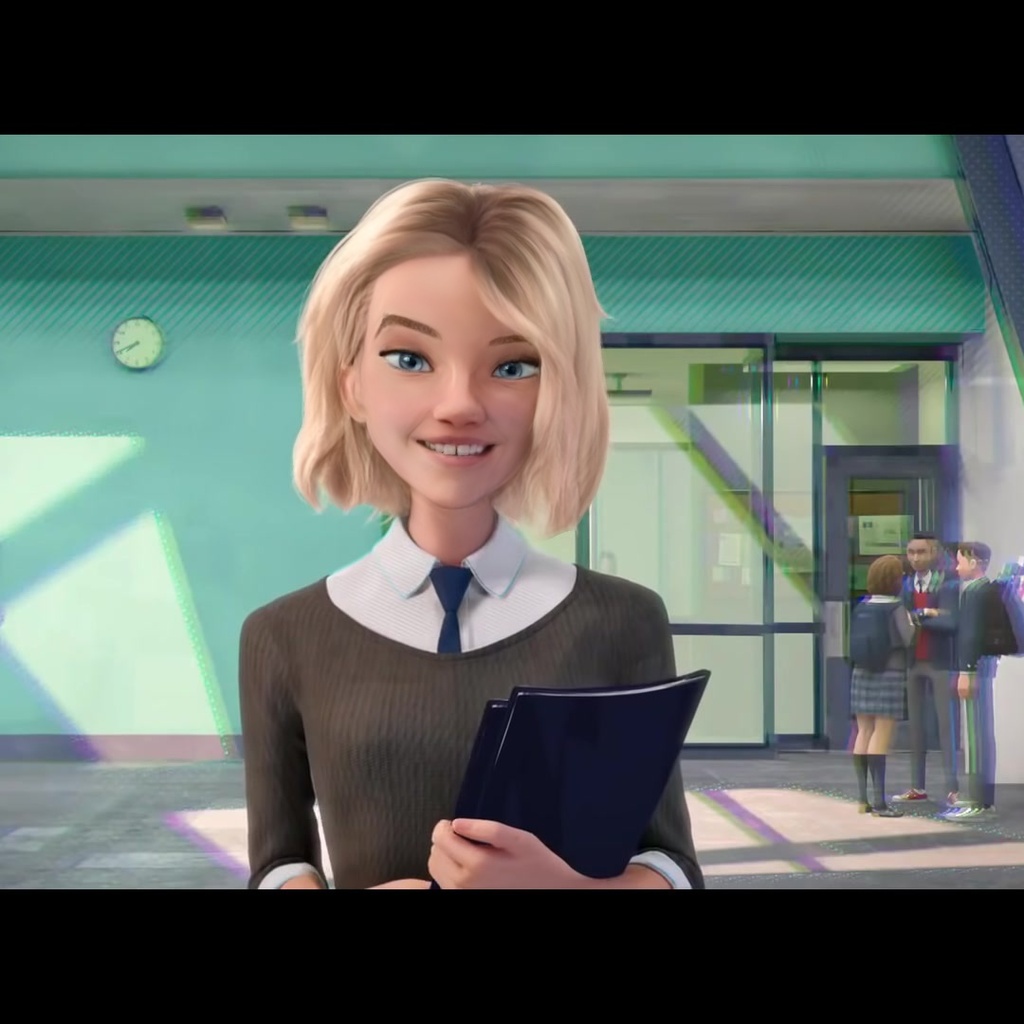} &
        \includegraphics[width=0.11\textwidth]{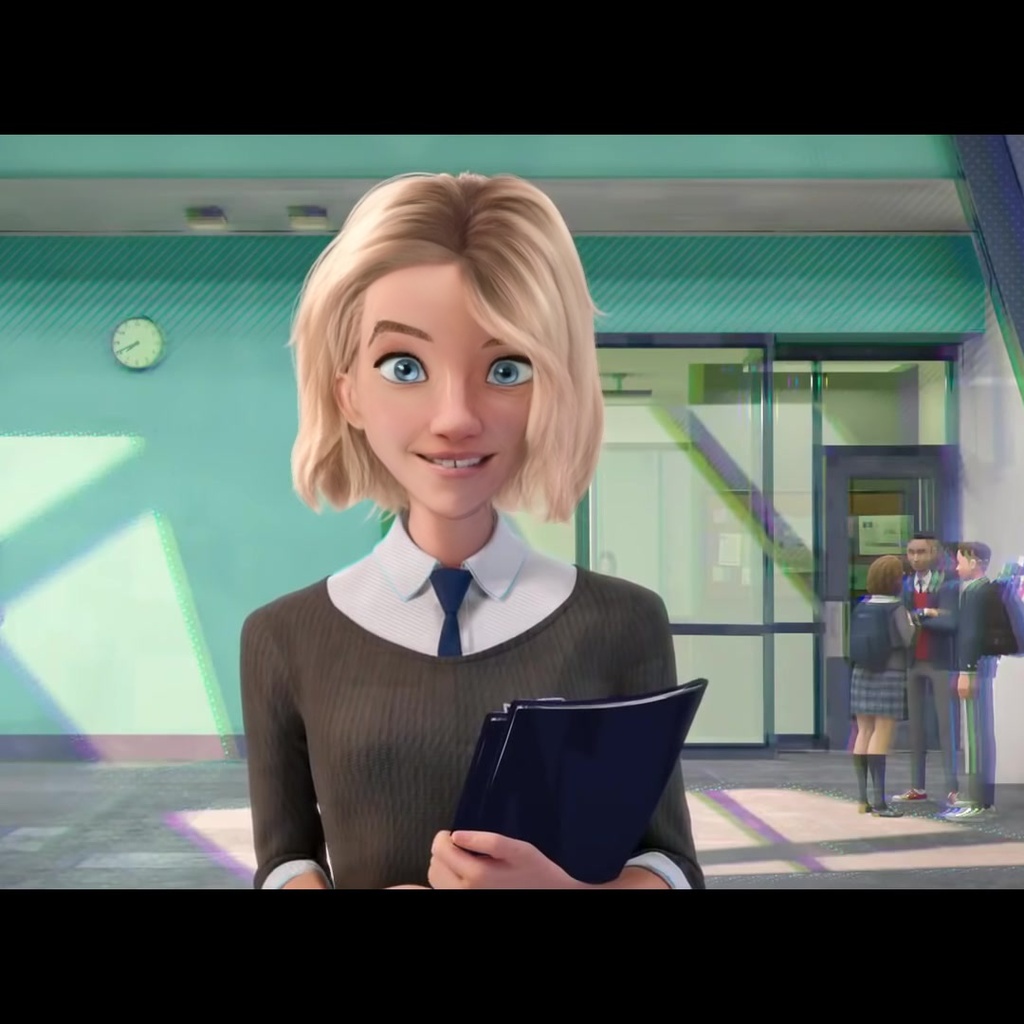} &
        \includegraphics[width=0.11\textwidth]{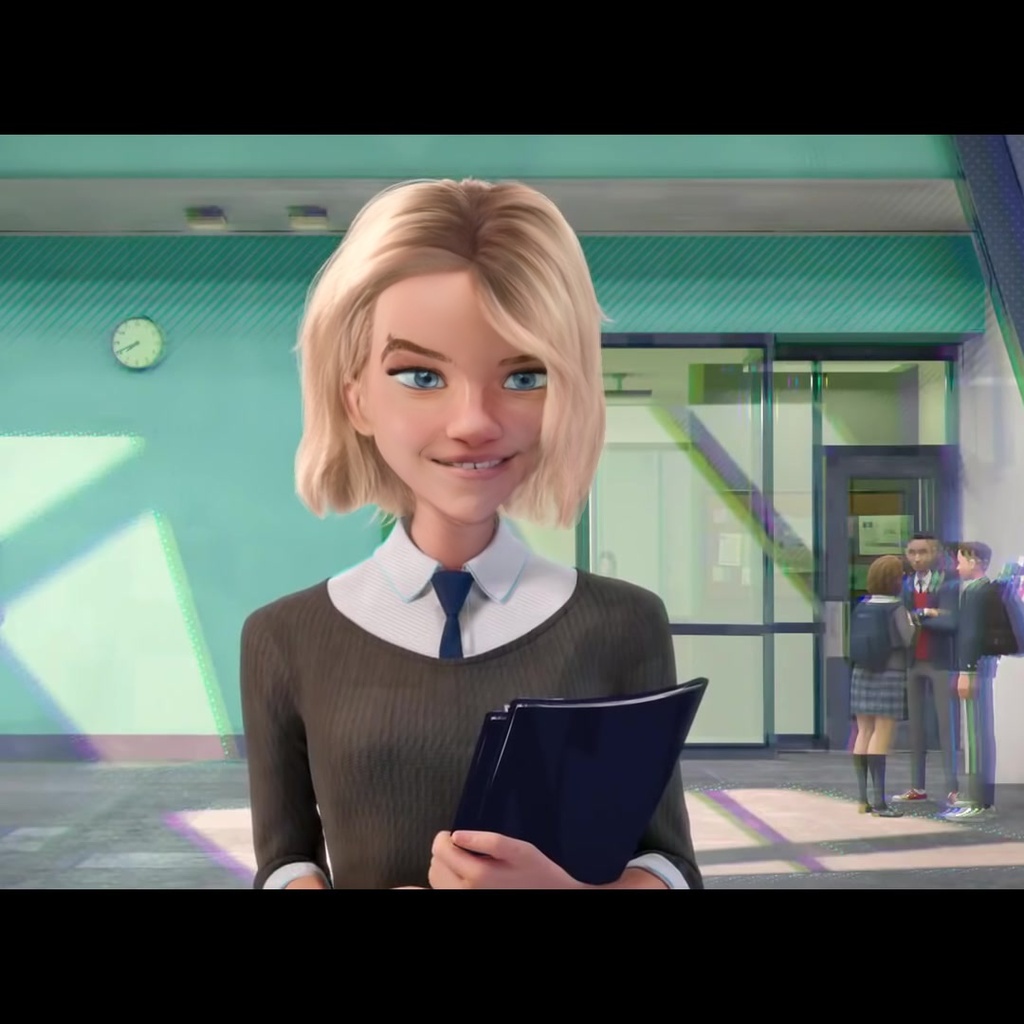} \\ \\
        
        \raisebox{0.15in}{\rotatebox{90}{Original}} &
        \includegraphics[width=0.11\textwidth]{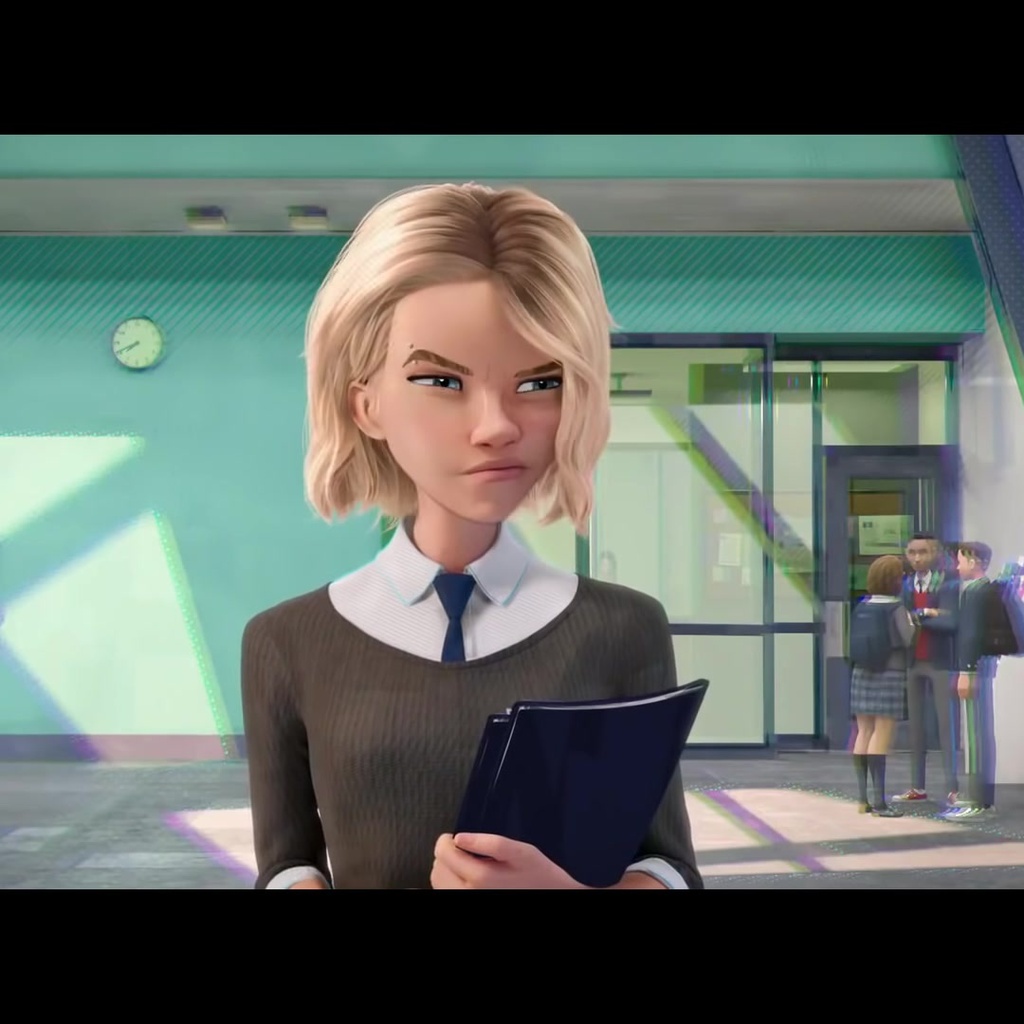} &
        \includegraphics[width=0.11\textwidth]{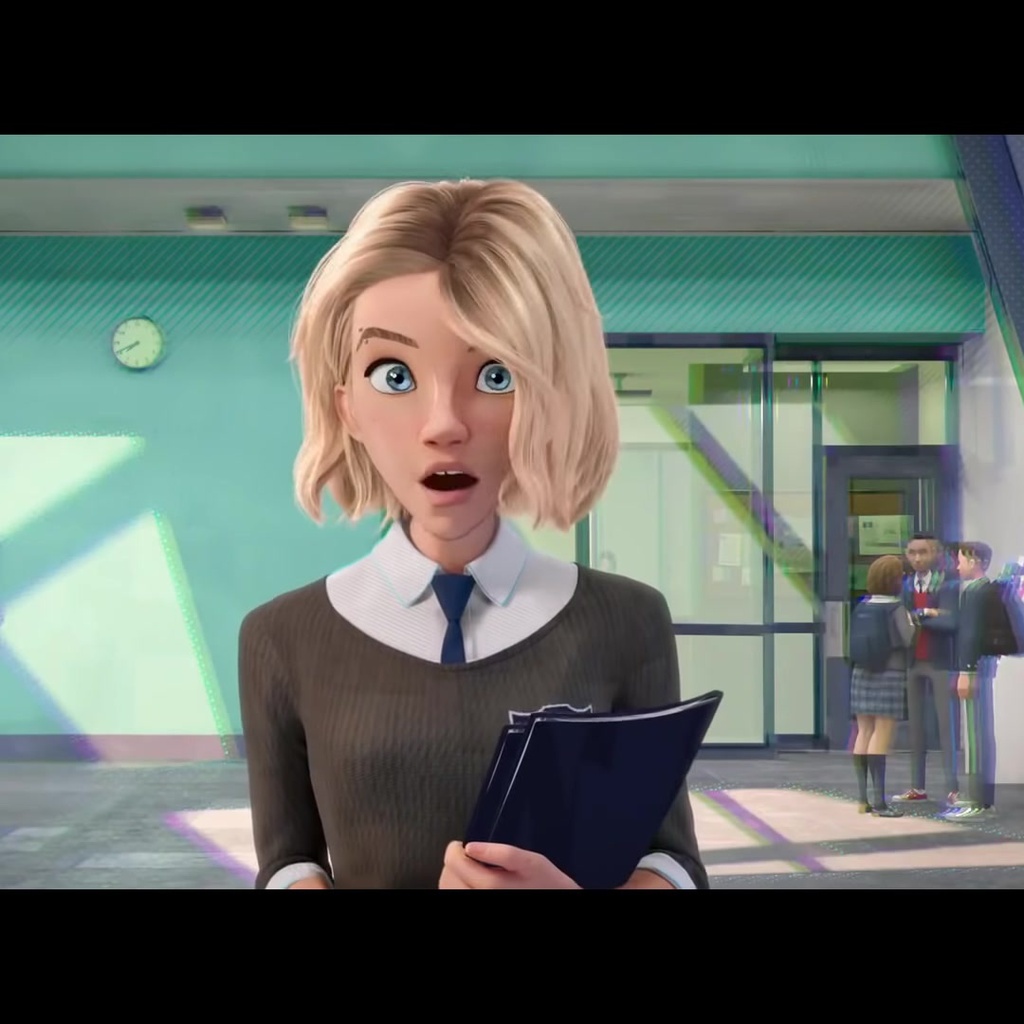} &
        \includegraphics[width=0.11\textwidth]{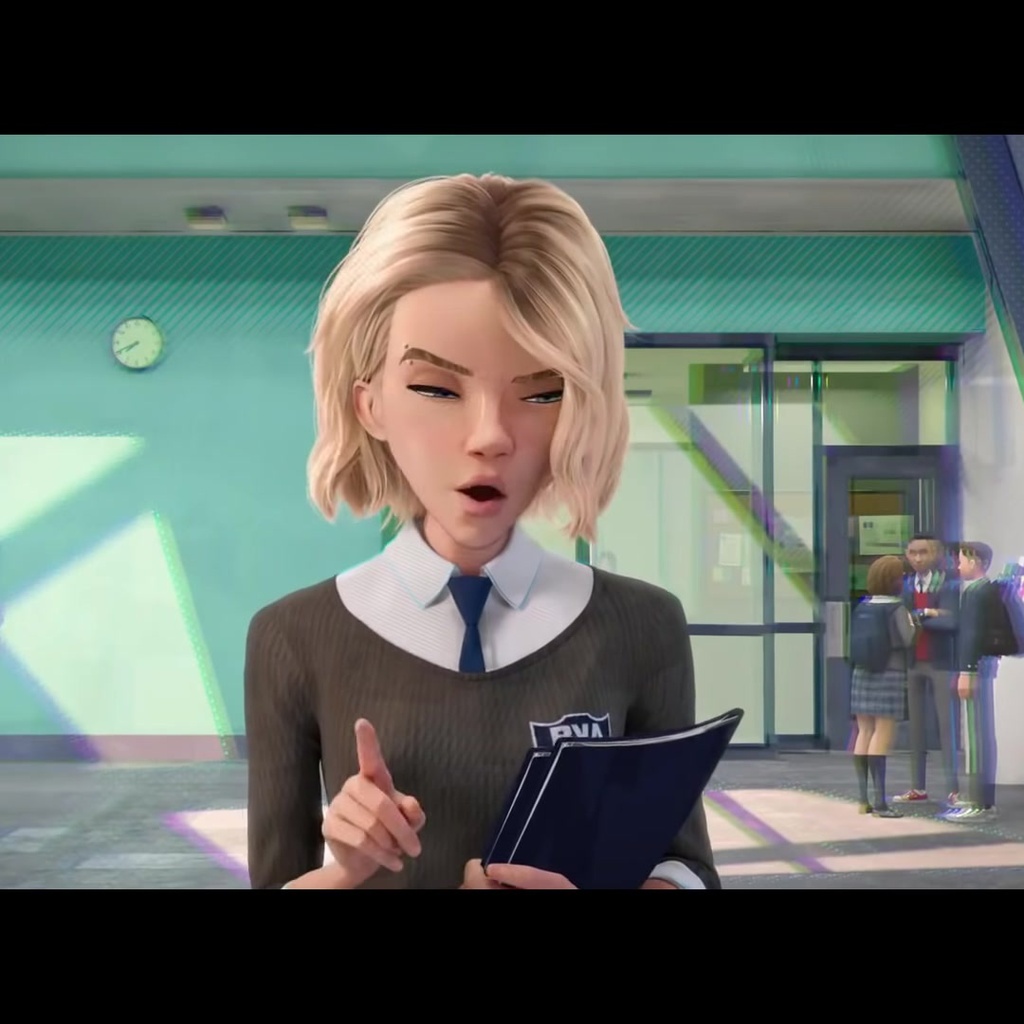} &
        \includegraphics[width=0.11\textwidth]{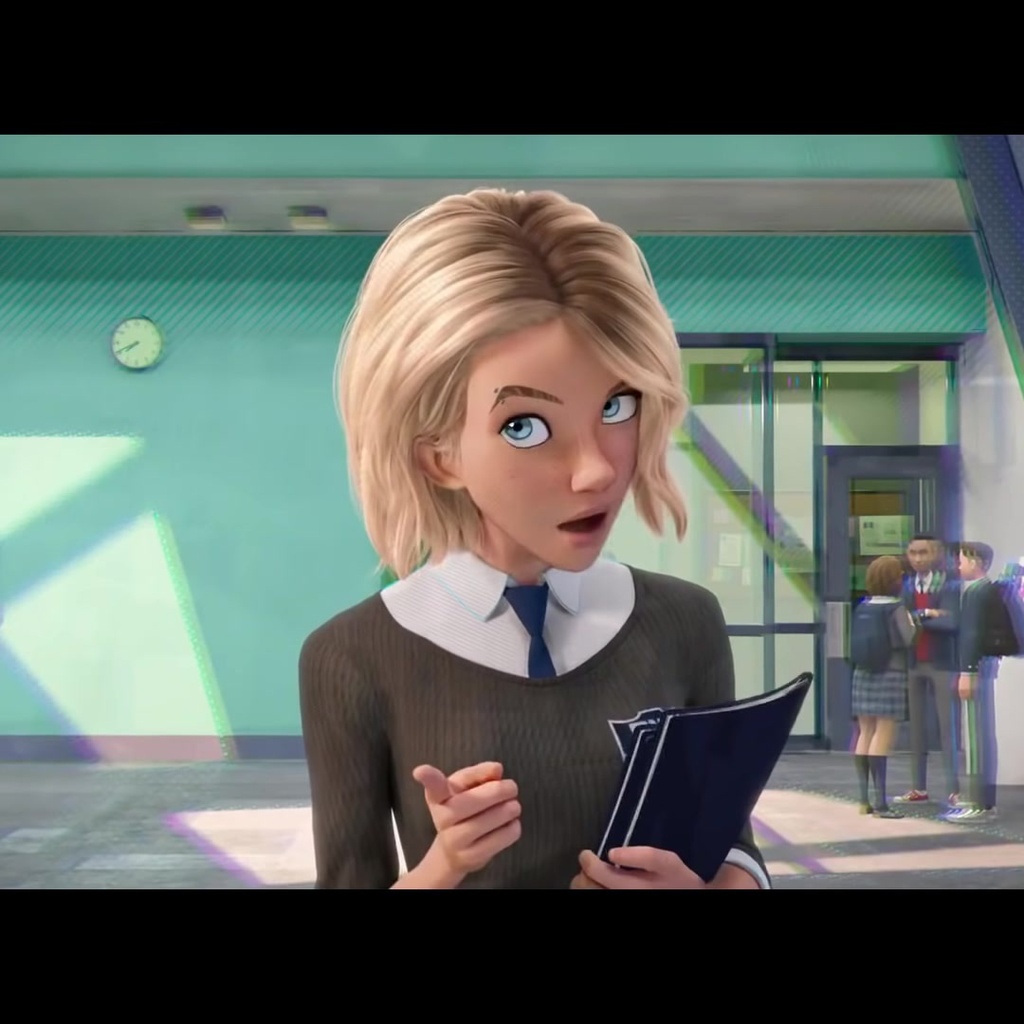} \\
        
        \raisebox{0.18in}{\rotatebox{90}{$+Smile$}} &
        \includegraphics[width=0.11\textwidth]{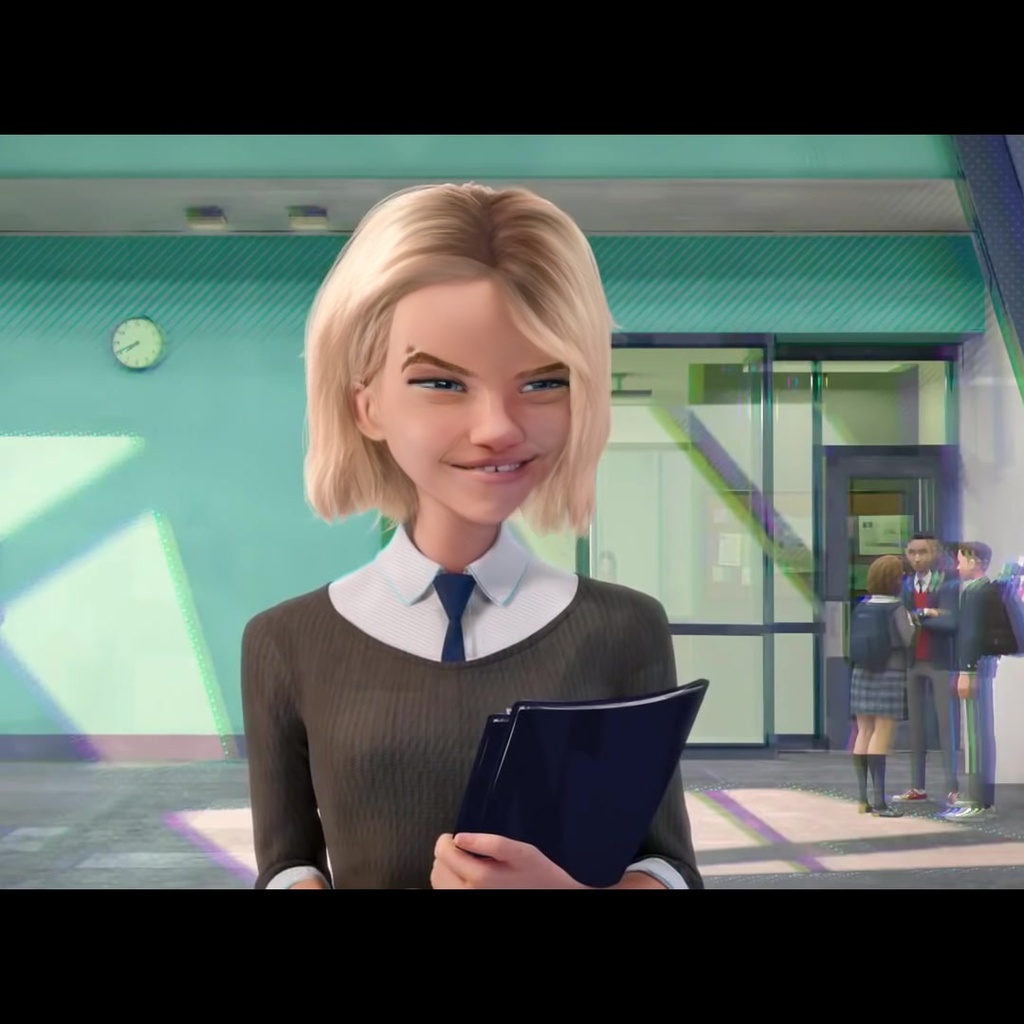} &
        \includegraphics[width=0.11\textwidth]{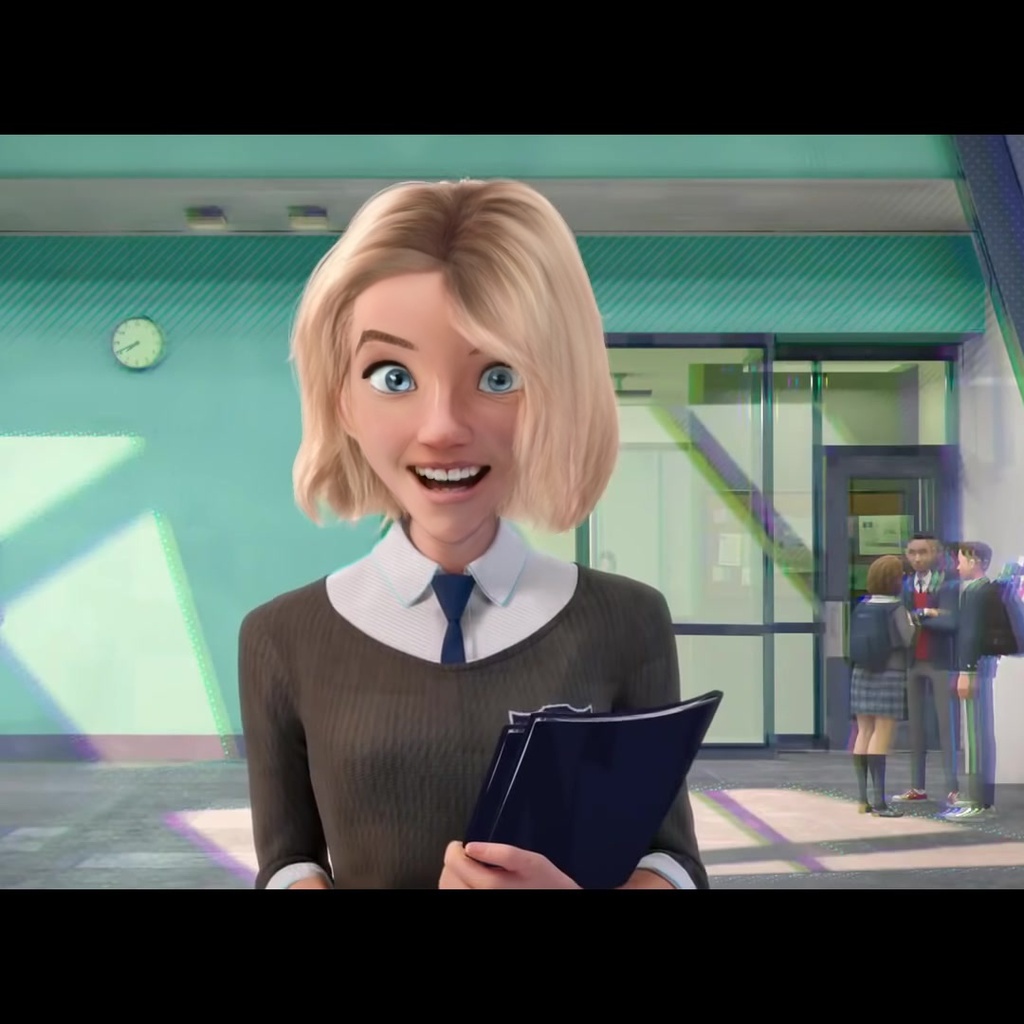} &
        \includegraphics[width=0.11\textwidth]{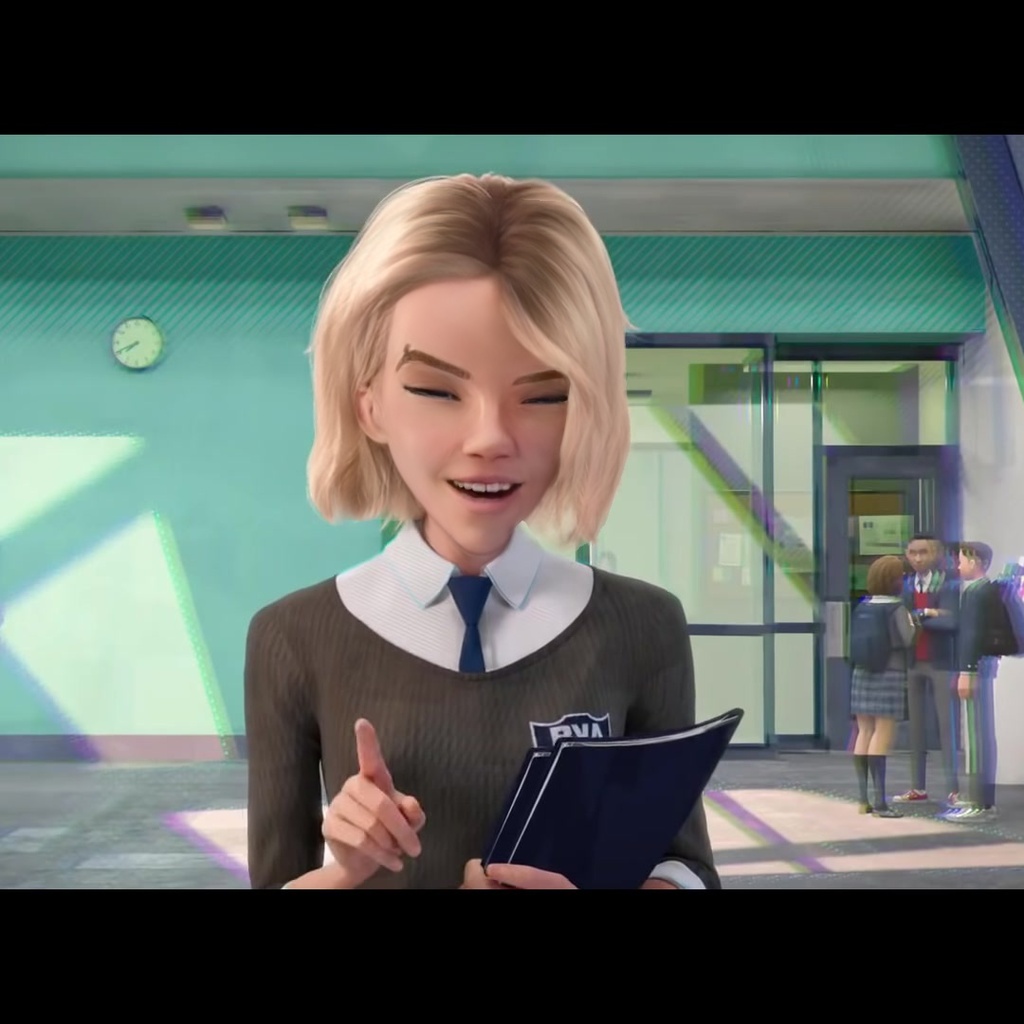} &
        \includegraphics[width=0.11\textwidth]{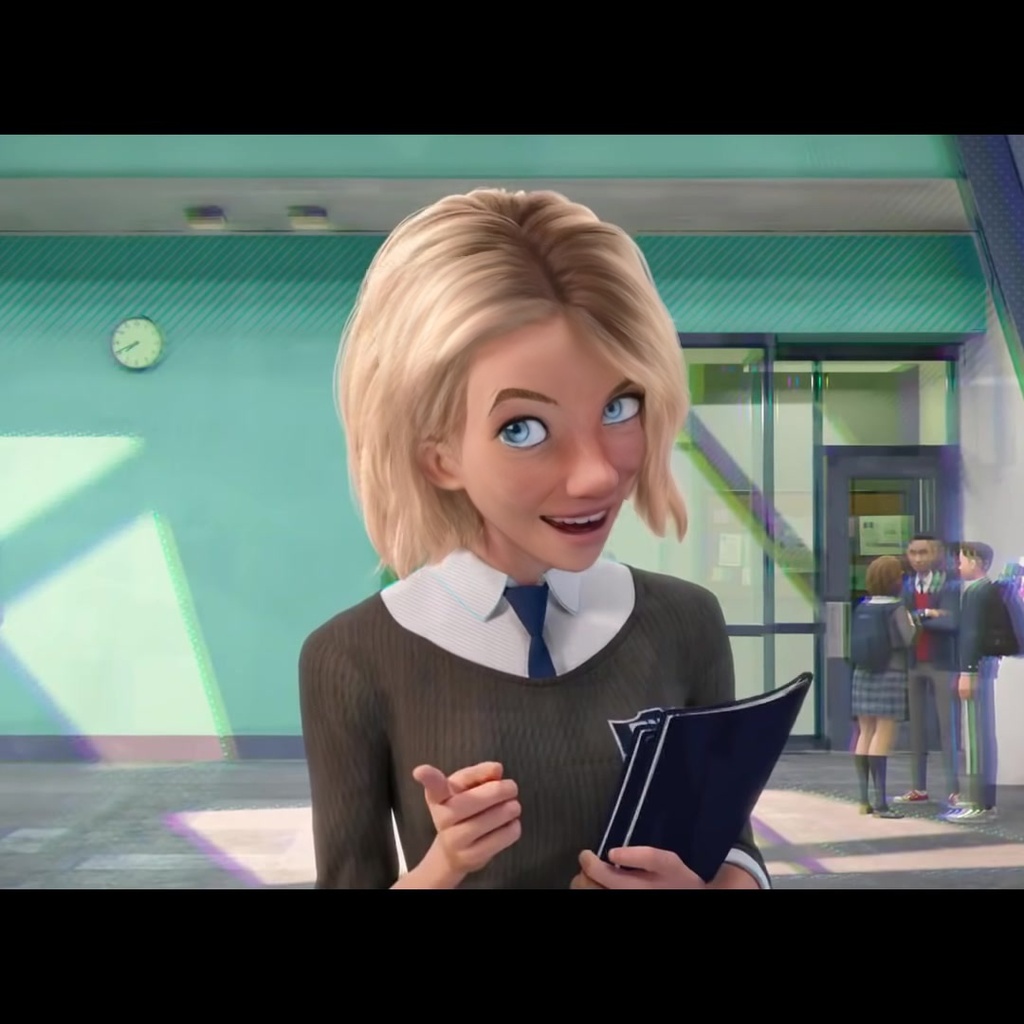} \\

        % \\

        %& Frame 1 & Frame 2 & Frame 3 & Frame 4 & Frame 1 & Frame 2 & Frame 3 & Frame 4
        
    \end{tabular}
    
    }
    \vspace{-0.225cm}
    \caption{Out-of-domain video editing. Our method can seamlessly adapt to other facial domains, and can handle challenging poses and expressions.}
    %\vspace{-0.225cm}
    \label{fig:ood}
\end{figure}

%% file: resources/tables/temporal_id.tex
\begin{table}[]
    \centering
\begin{tabular}{c | c | c }
        \toprule
        Model & TL-ID $\uparrow$ & TG-ID $\uparrow$ \\
        \midrule
        Latent Transformer & 0.976 & 0.811 \\
        PTI (optimization) & 0.933 & 0.901 \\
        Ours & \textbf{0.996} & \textbf{0.933} \\
        \bottomrule
\end{tabular}

    \caption{Temporal consistency metrics. Encoder based methods display improved identity preservation at the local (adjacent frame) level, but show considerable identity drift over time. PTI, preserves a greater degree of global identity, at the cost of local jitter from inconsistent pivots. Our pipeline outperforms the alternatives and achieves a local-identity preservation score which is nearly equal to the original video (1), demonstrating our ability to maintain a high degree of consistency.}
    \vspace{-0.35cm}
    \label{tb:temporal_id}
    
\end{table}

%% file: resources/figures/ablation.tex
\begin{figure}
\setlength{\tabcolsep}{0.5pt}
    \centering
    { \small 
\begin{tabular}{ccccc}
Original & w/o e4e & w/o PTI & w/o Stitching & Ours \\
\raisebox{-.32\totalheight}{\includegraphics[width=0.2\columnwidth]{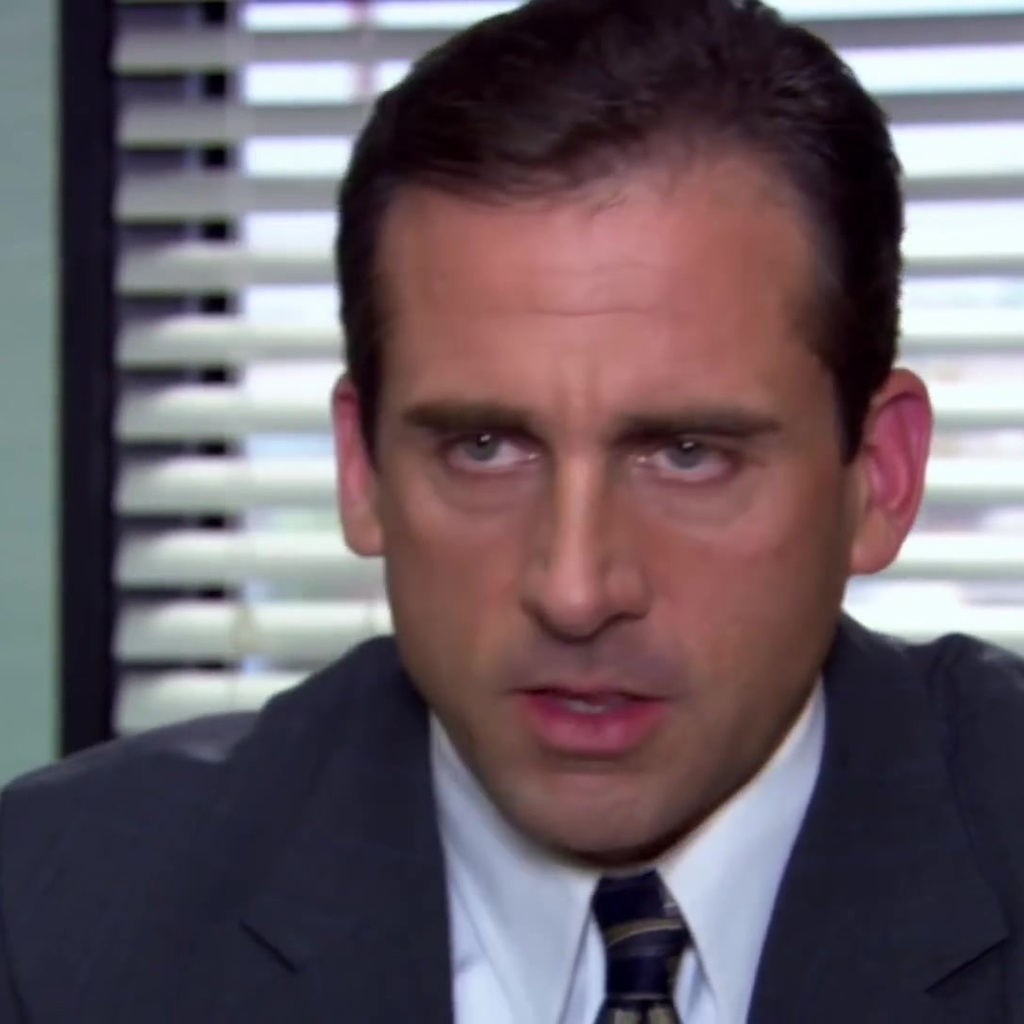}} &
\raisebox{-.32\totalheight}{\includegraphics[width=0.2\columnwidth]{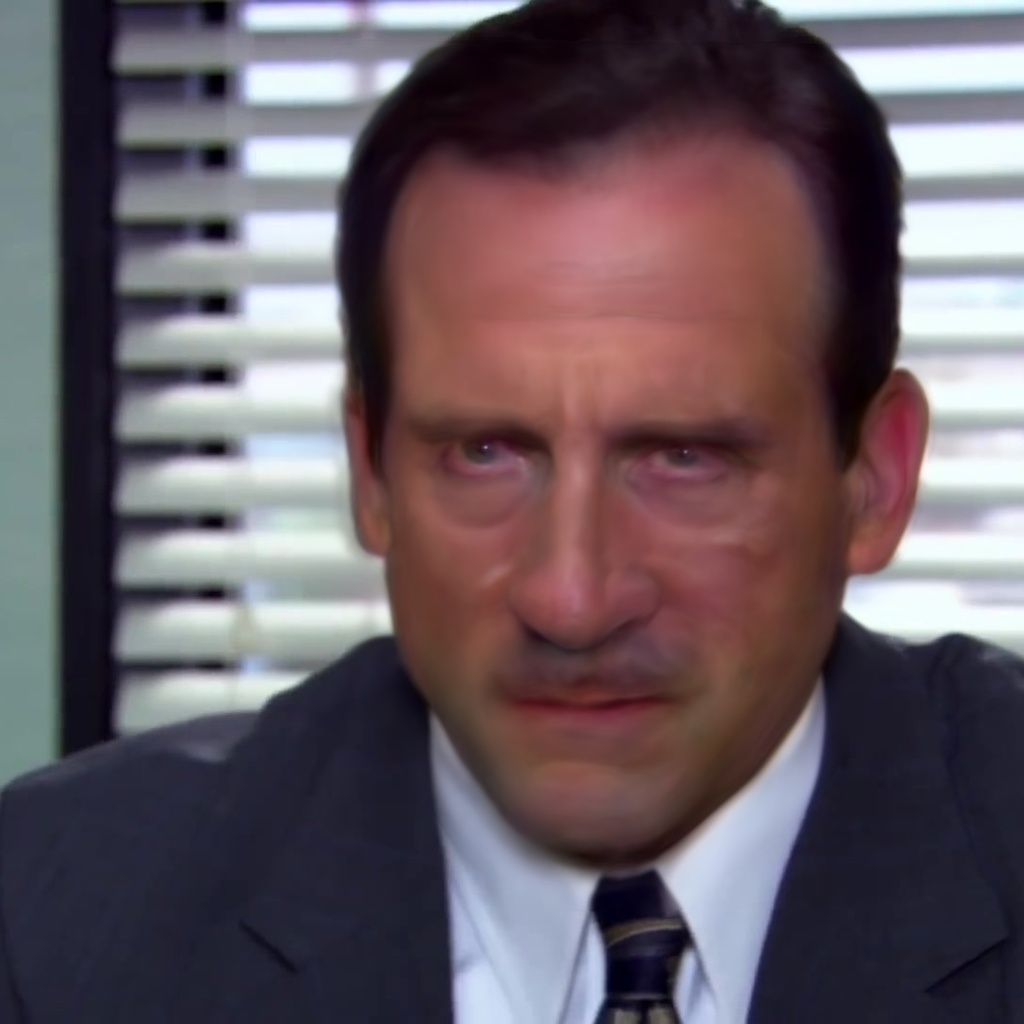}} &
\raisebox{-.32\totalheight}{\includegraphics[width=0.2\columnwidth]{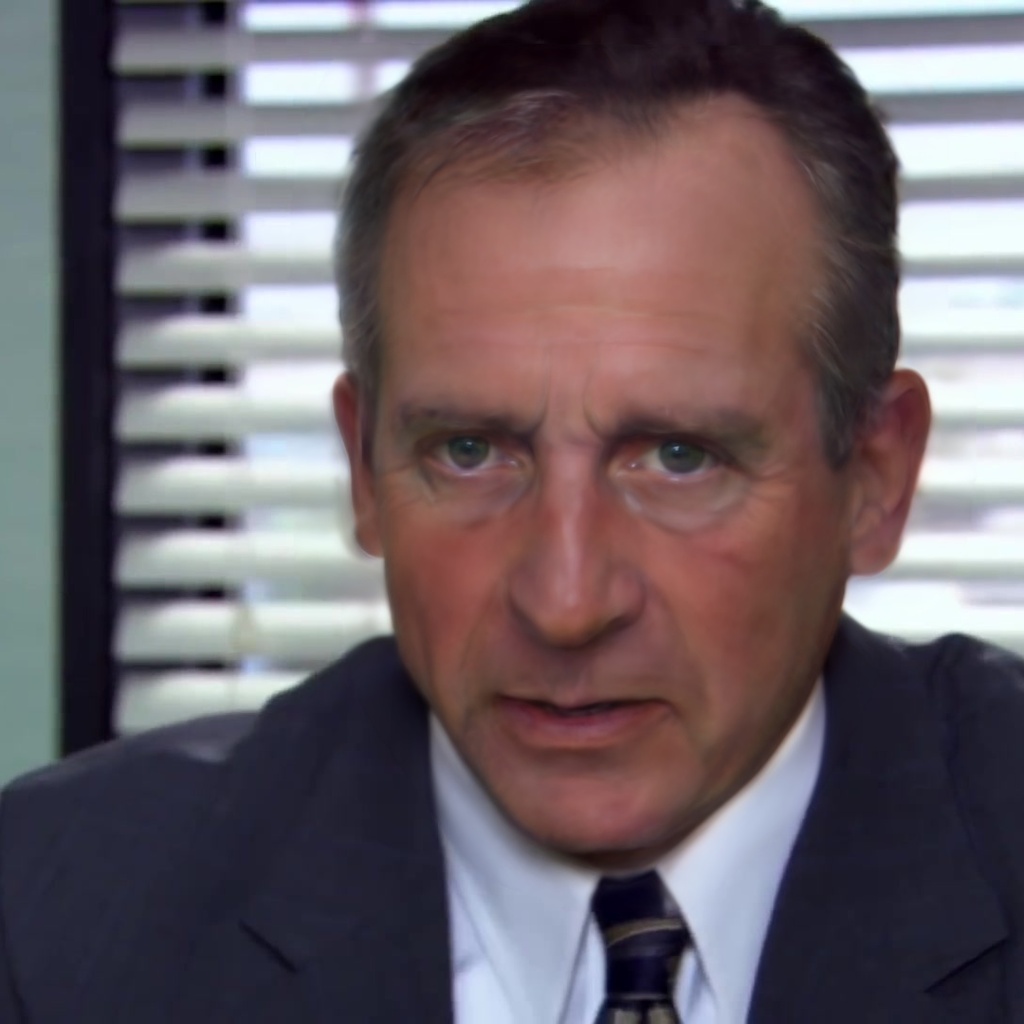}} &
\raisebox{-.32\totalheight}{\includegraphics[width=0.2\columnwidth]{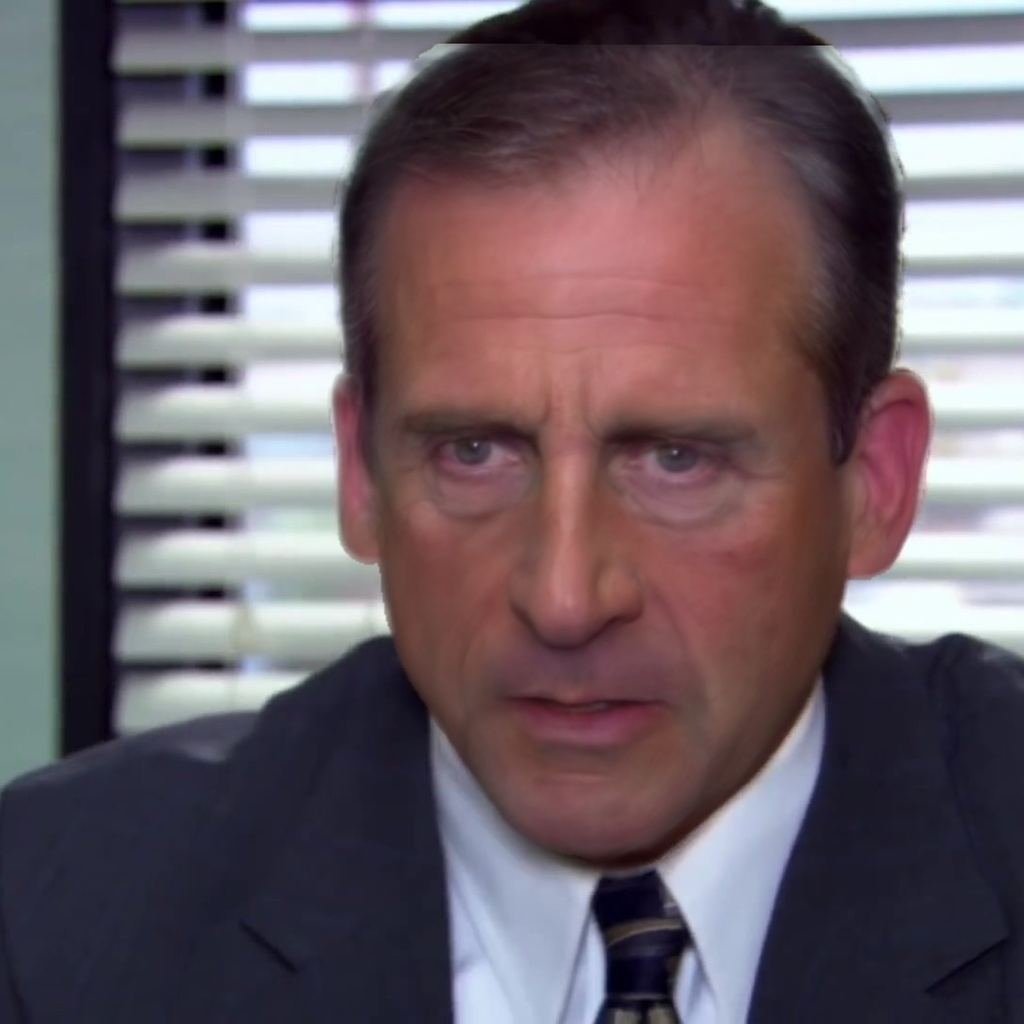}} &
\raisebox{-.32\totalheight}{\includegraphics[width=0.2\columnwidth]{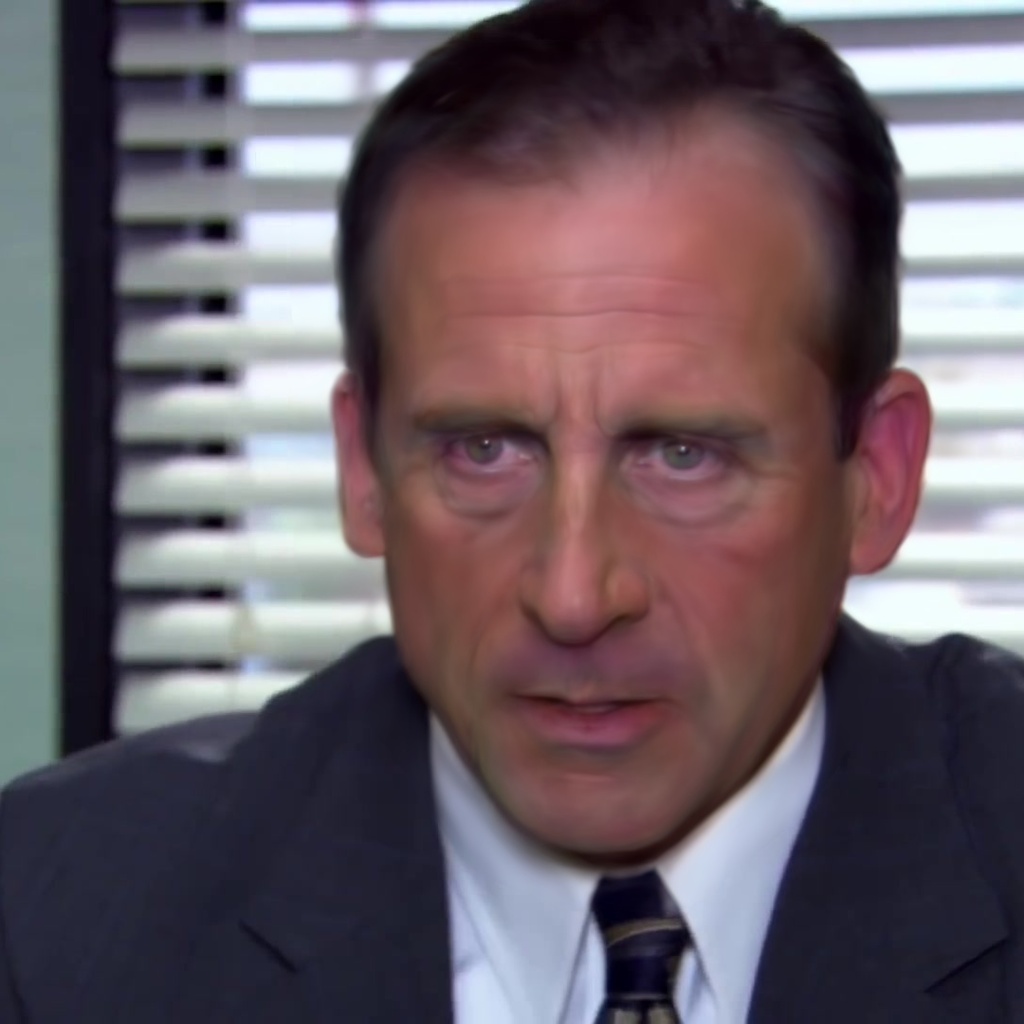}} \\

\raisebox{-.32\totalheight}{\includegraphics[width=0.2\columnwidth]{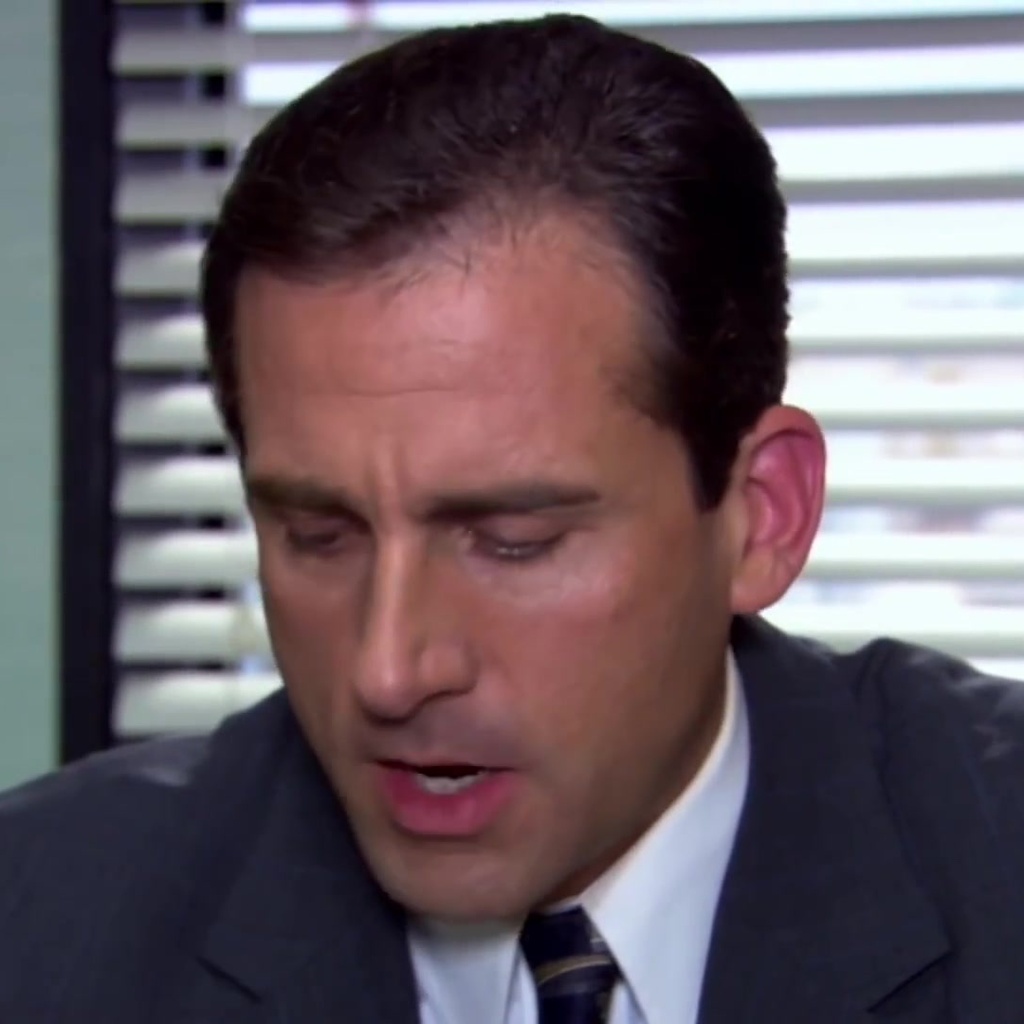}} &
\raisebox{-.32\totalheight}{\includegraphics[width=0.2\columnwidth]{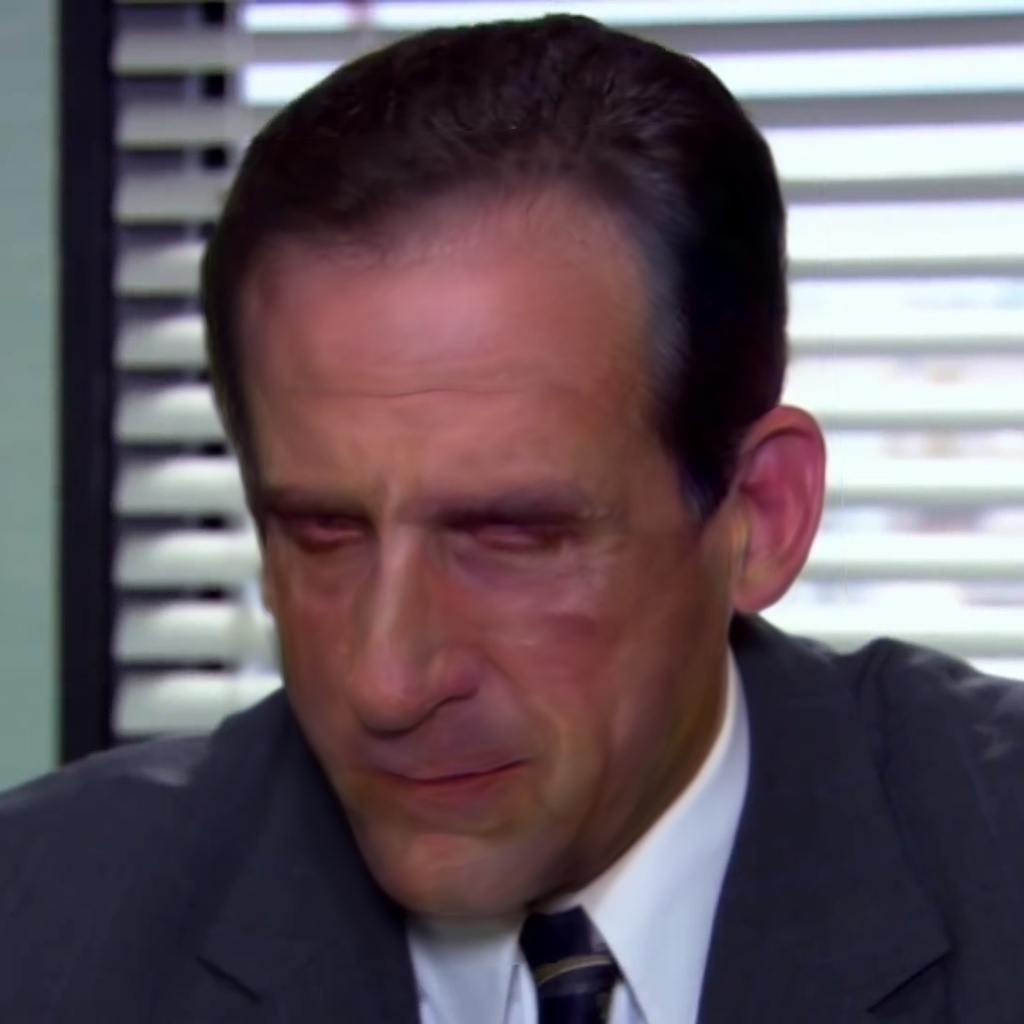}} &
\raisebox{-.32\totalheight}{\includegraphics[width=0.2\columnwidth]{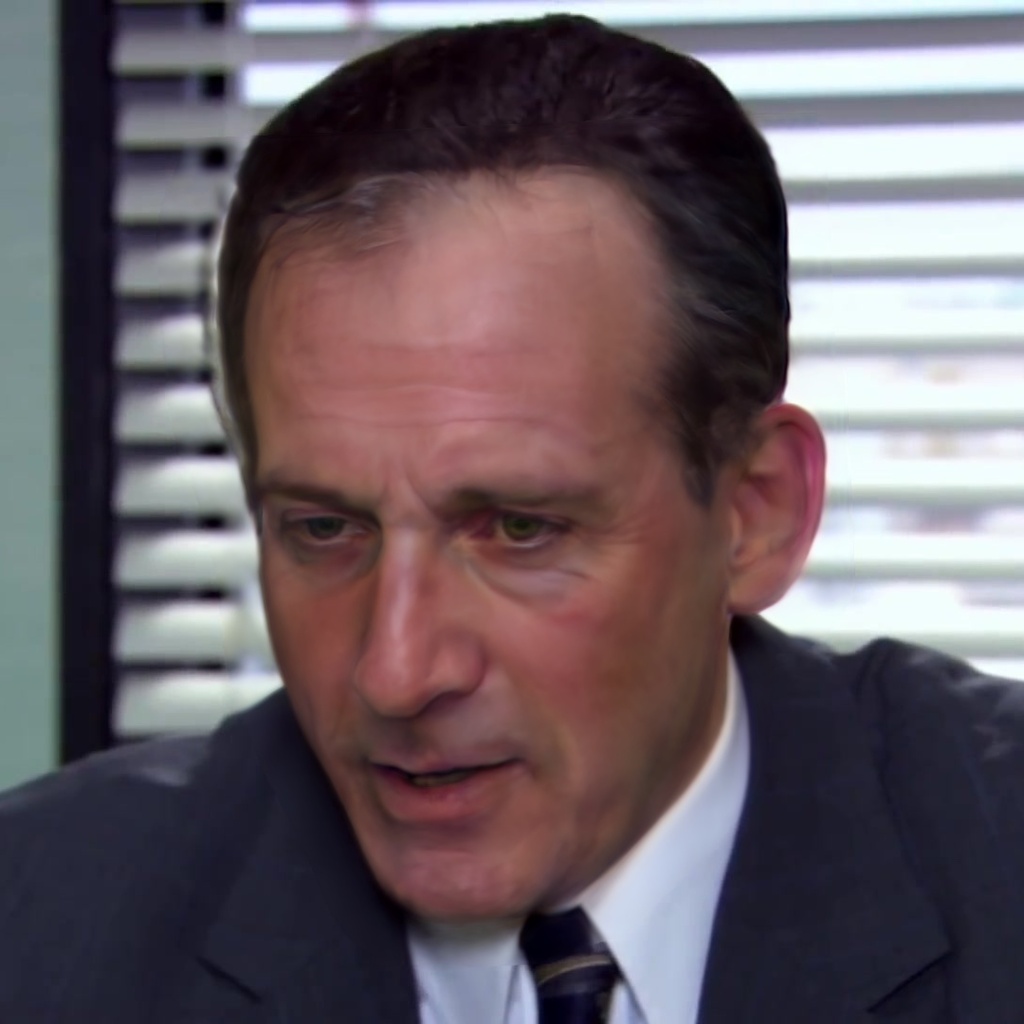}} &
\raisebox{-.32\totalheight}{\includegraphics[width=0.2\columnwidth]{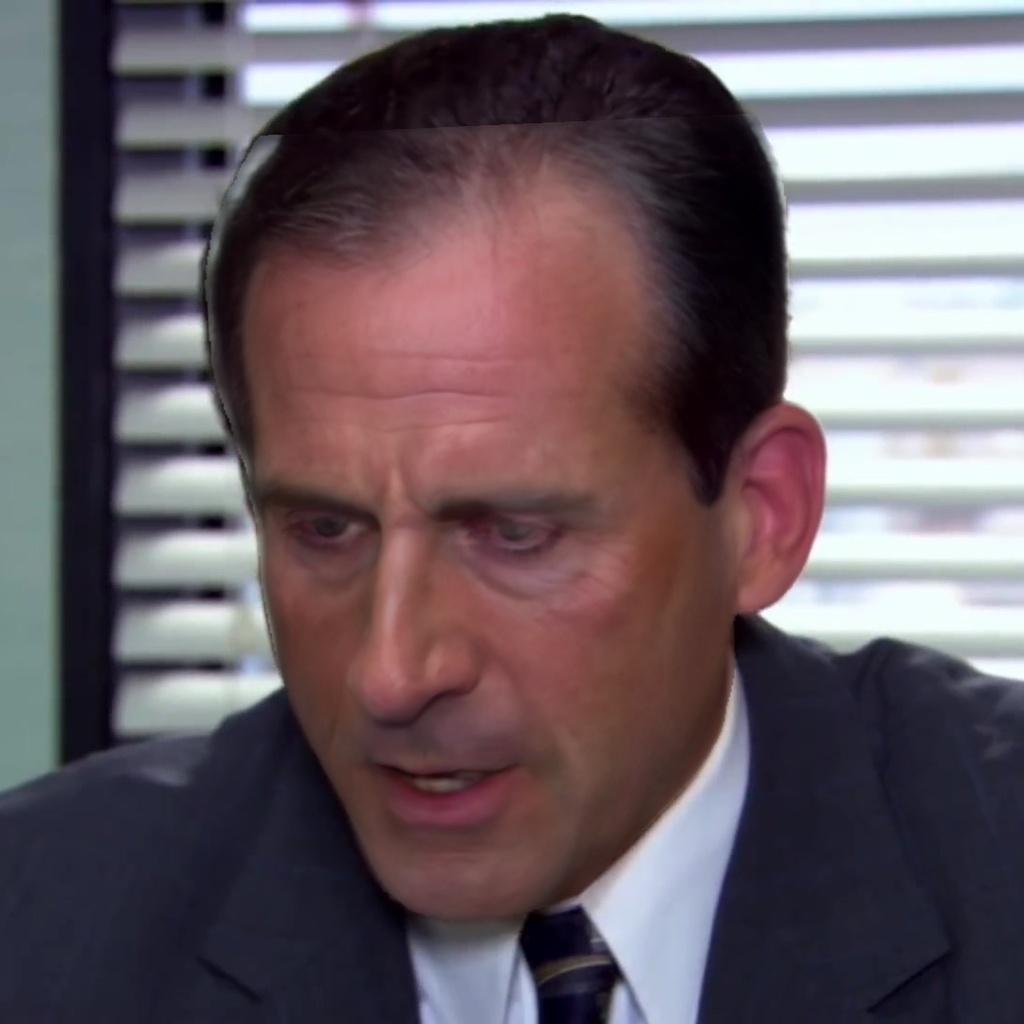}} &
\raisebox{-.32\totalheight}{\includegraphics[width=0.2\columnwidth]{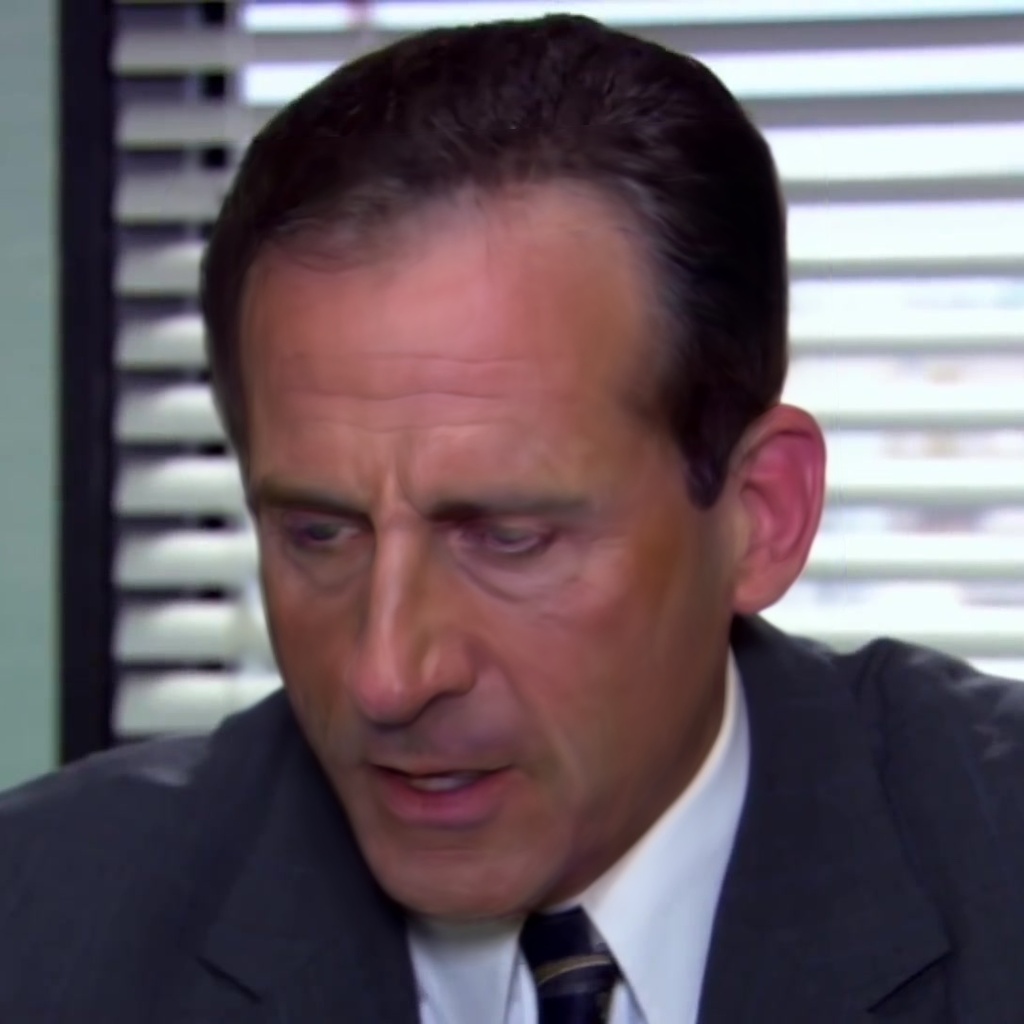}} \\

\raisebox{-.32\totalheight}{\includegraphics[width=0.2\columnwidth]{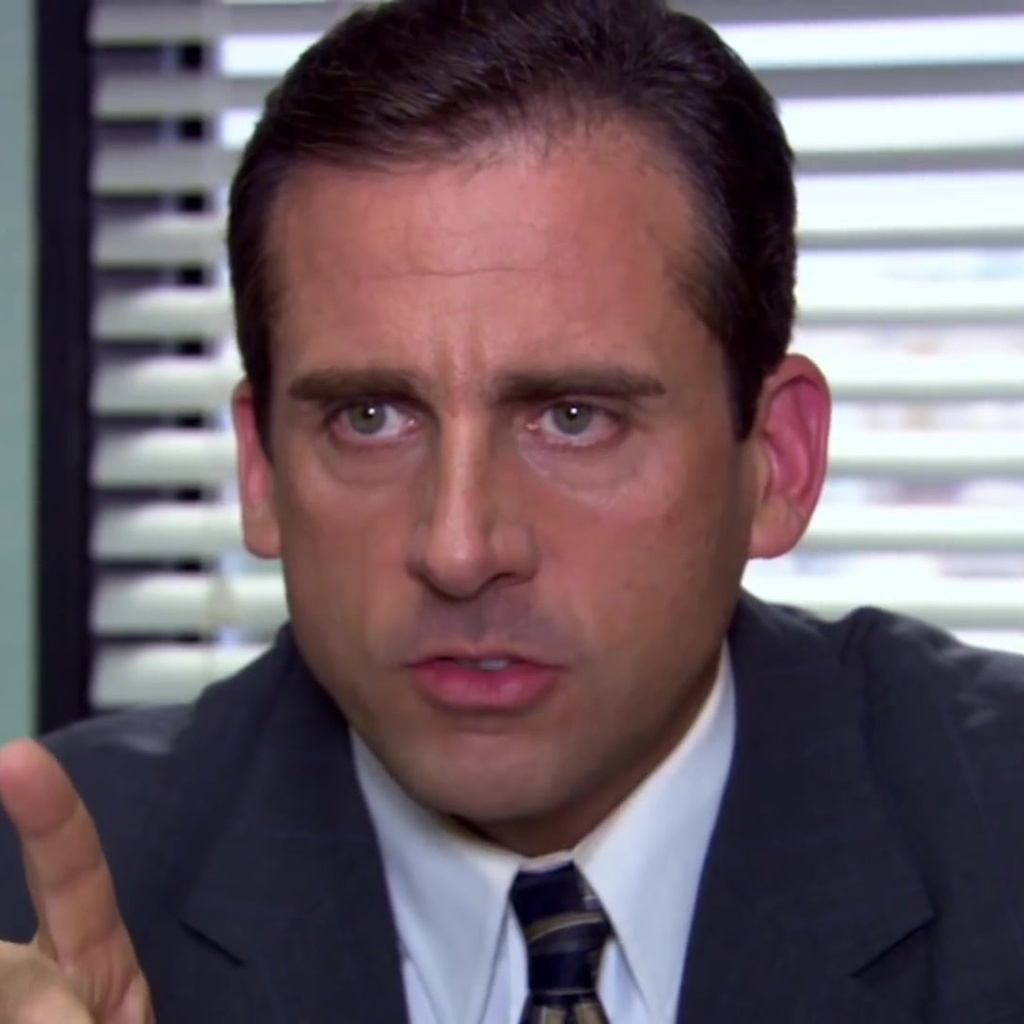}} &
\raisebox{-.32\totalheight}{\includegraphics[width=0.2\columnwidth]{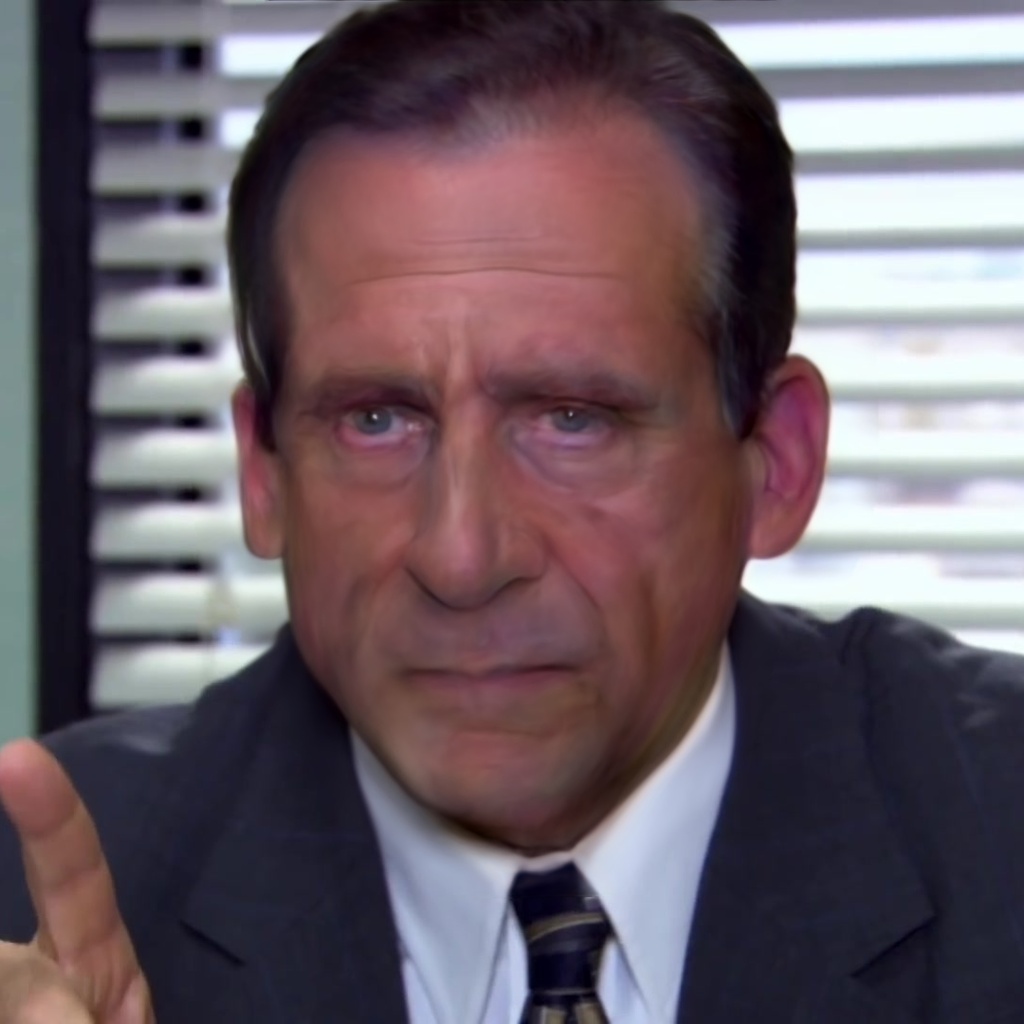}} &
\raisebox{-.32\totalheight}{\includegraphics[width=0.2\columnwidth]{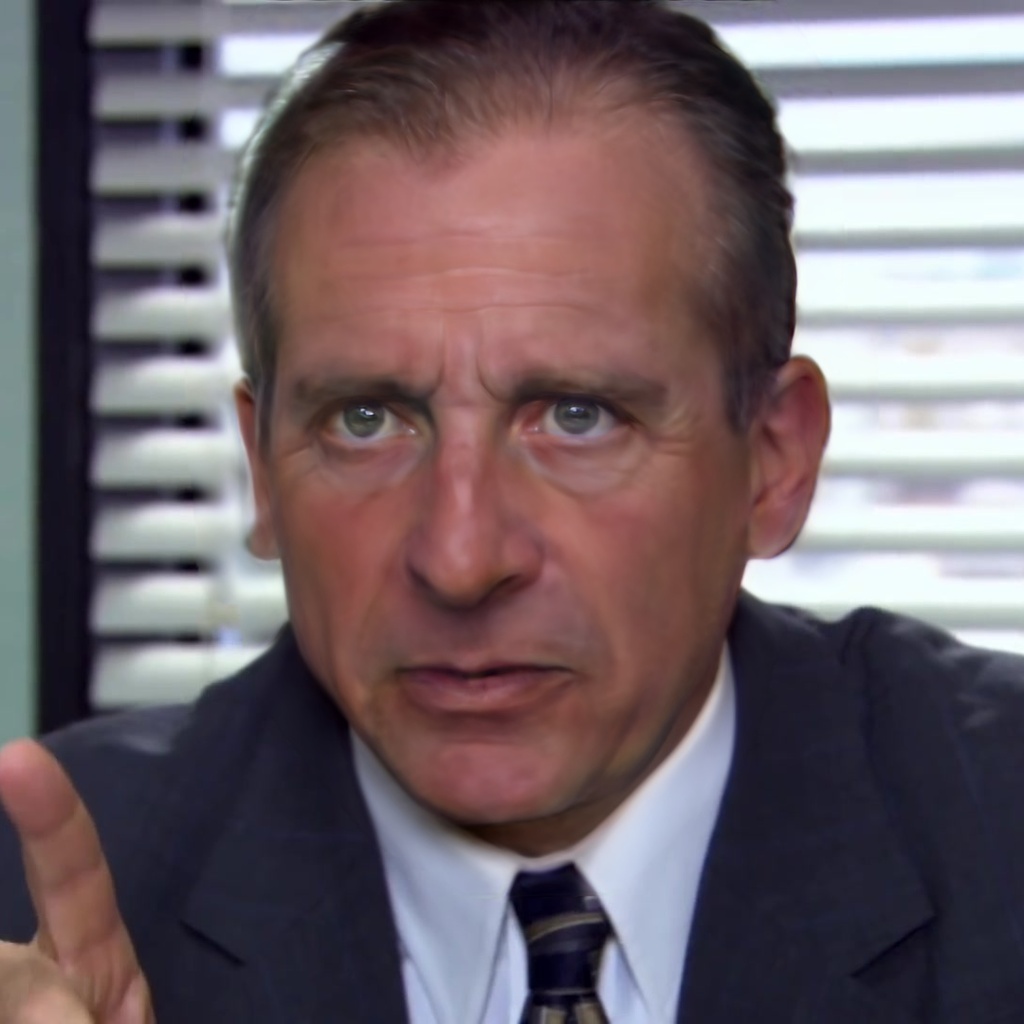}} &
\raisebox{-.32\totalheight}{\includegraphics[width=0.2\columnwidth]{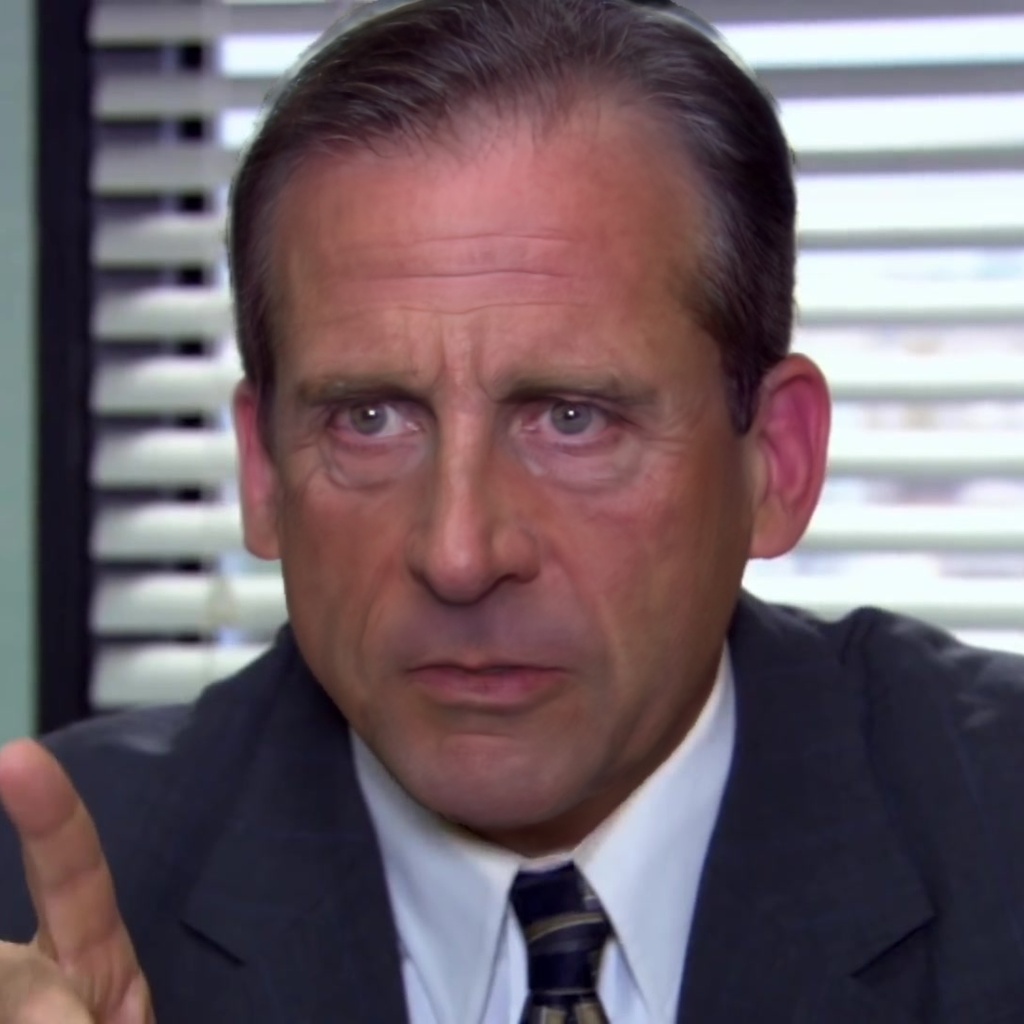}} &
\raisebox{-.32\totalheight}{\includegraphics[width=0.2\columnwidth]{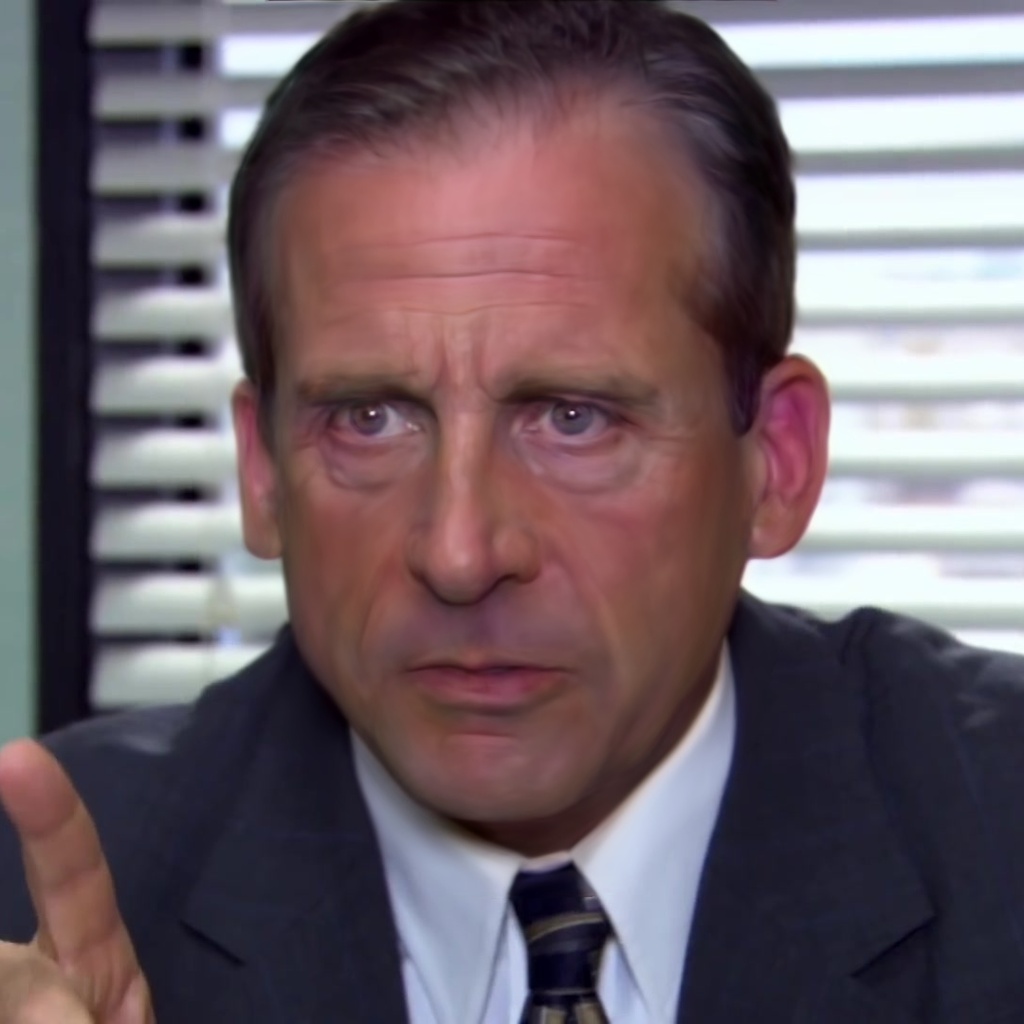}} \\

\raisebox{-.32\totalheight}{\includegraphics[width=0.2\columnwidth]{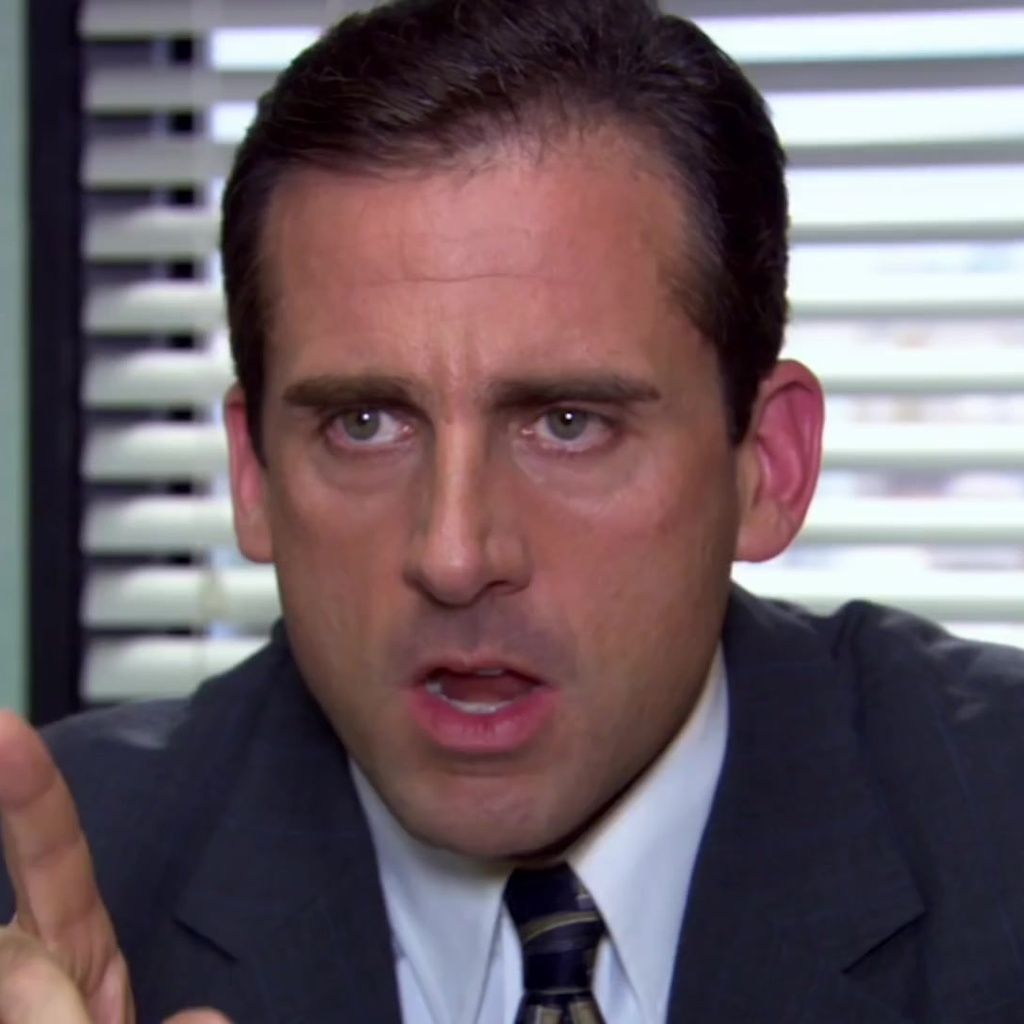}} &
\raisebox{-.32\totalheight}{\includegraphics[width=0.2\columnwidth]{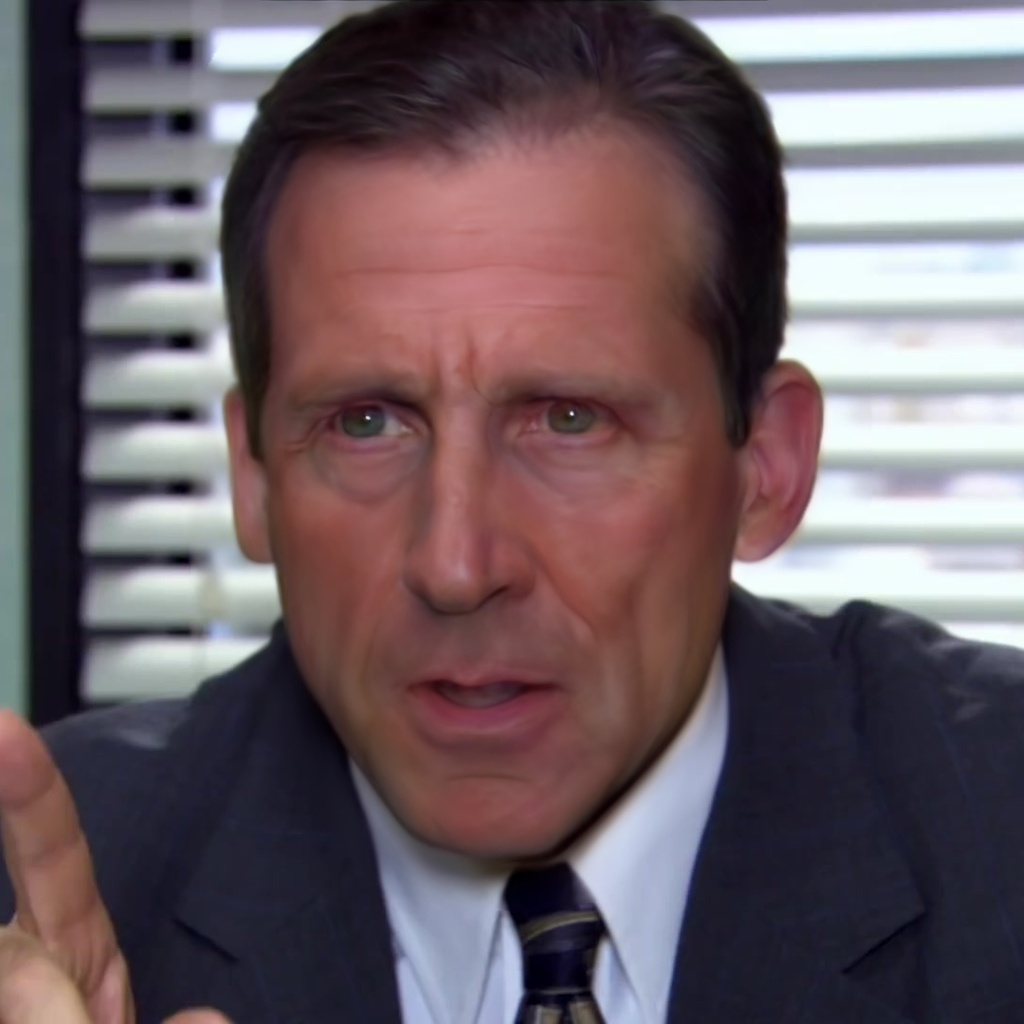}} &
\raisebox{-.32\totalheight}{\includegraphics[width=0.2\columnwidth]{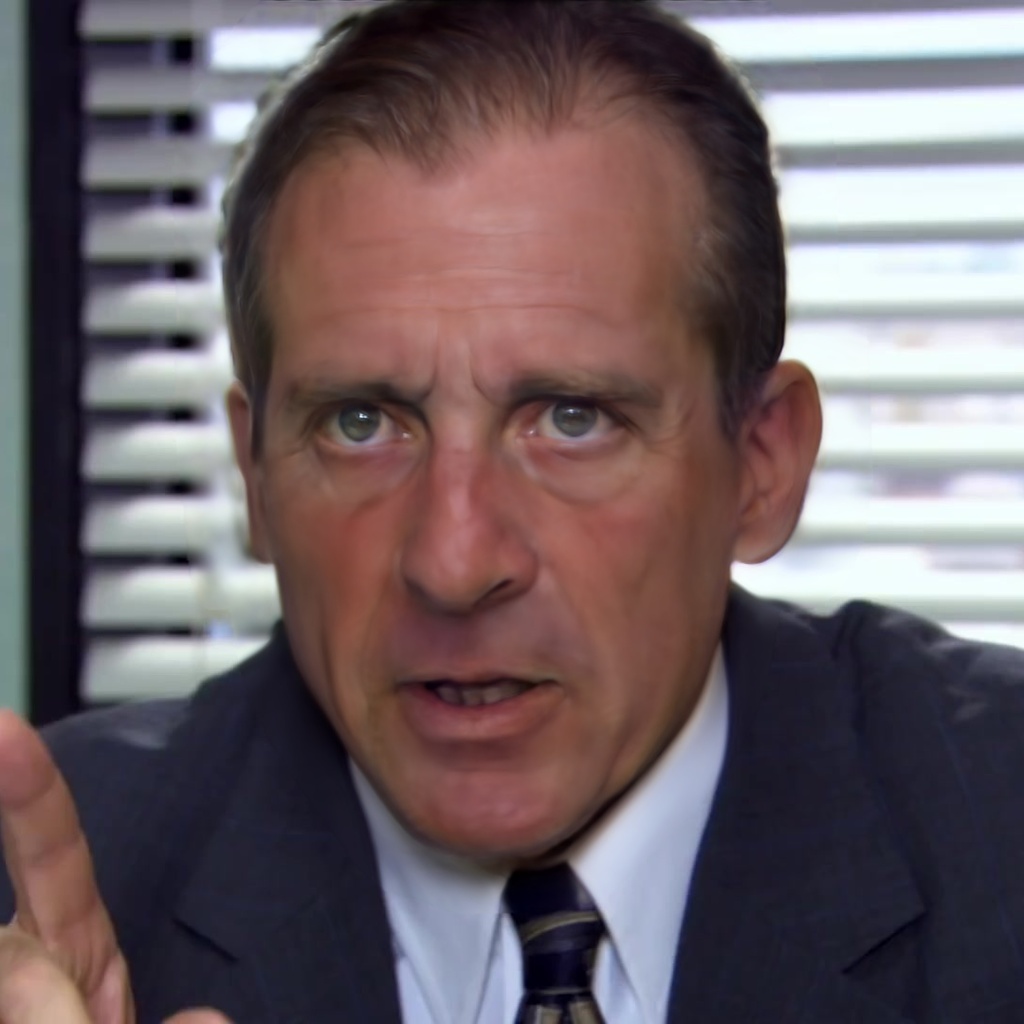}} &
\raisebox{-.32\totalheight}{\includegraphics[width=0.2\columnwidth]{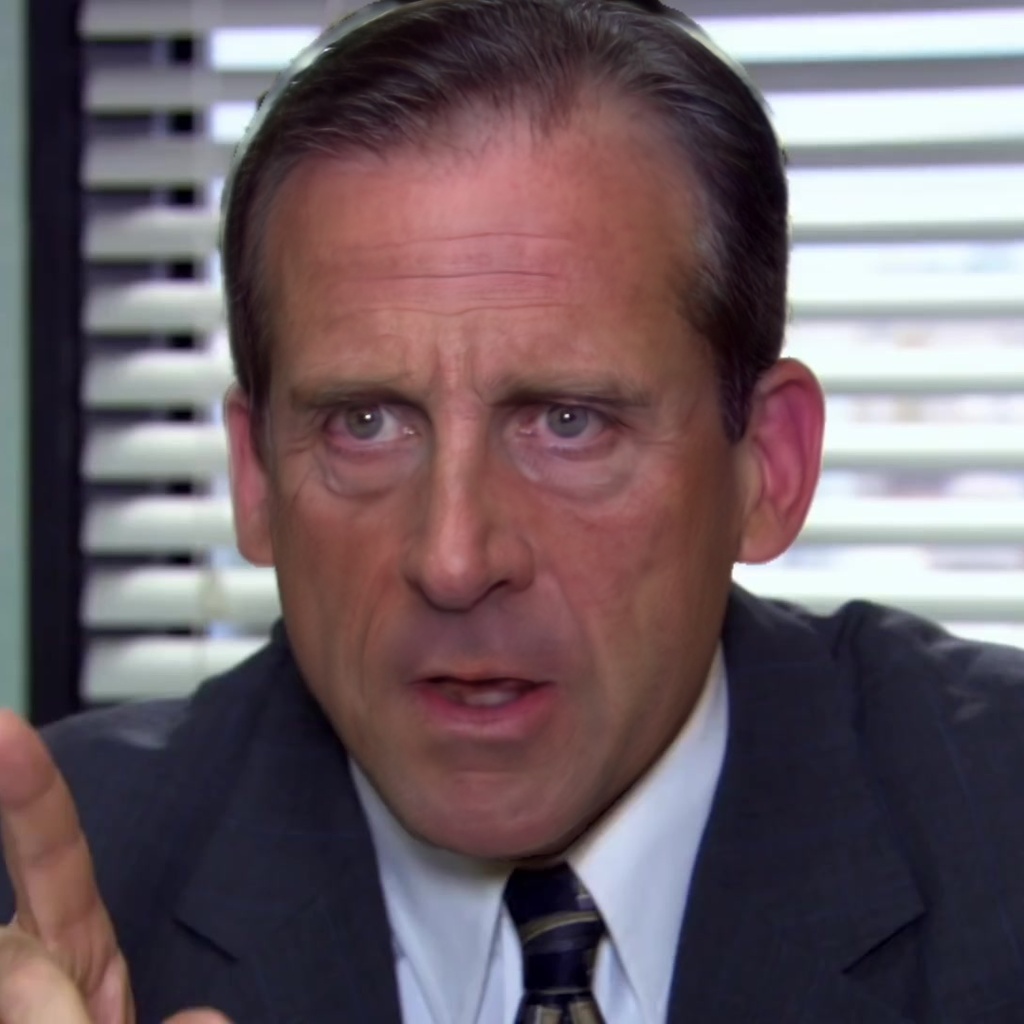}} &
\raisebox{-.32\totalheight}{\includegraphics[width=0.2\columnwidth]{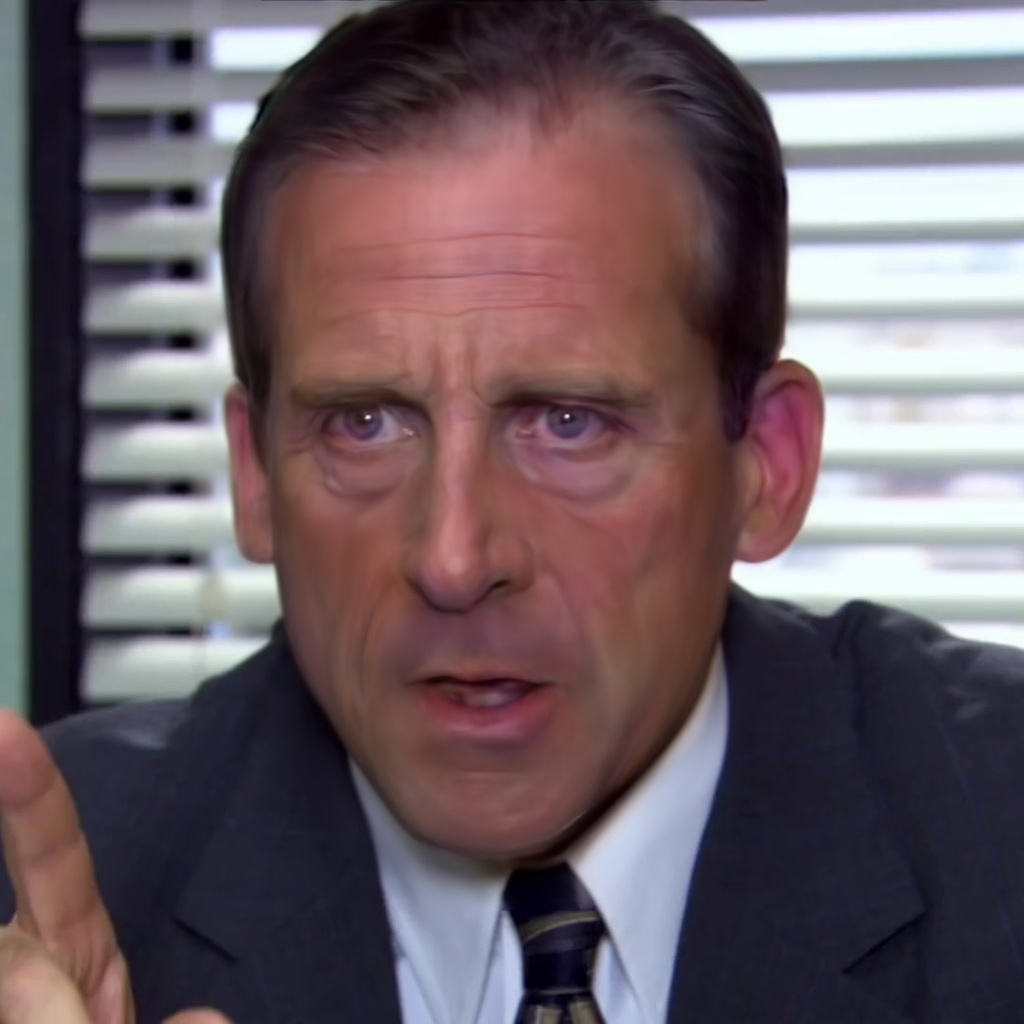}} \\

\end{tabular}
}

\caption{Qualitative demonstration of the importance of our pipeline components. Replacing the encoder with an optimization step results in poor editing consistency. Without PTI, identity drifts over time, and stitching performance deteriorates. Replacing stitching with a mask-based blending scheme results in visual artifacts, such as sharp transitions in hair regions. Our full pipeline successfully avoids these pitfalls and generates a consistent video.}
\label{fig:ablation}
\end{figure}

%% file: conclusion.tex
\section{Discussion}

We presented a novel approach for semantic editing of facial videos. %without requiring additional temporal supervision. 
By employing smooth and consistent tools, we demonstrated that standard StyleGAN editing techniques can be readily applied to in-the-wild videos, without compromising temporal coherency.

While our method works well in many practical scenarios, it still faces some limitations.
Particularly, the StyleGAN alignment process is prone to leaving portions of hair (\eg, pigtails) outside the cropped region. These `external' regions do not undergo any semantic manipulations, and may result in jarring transitions when attempting to modify attributes such as hair length or color. 
A further limitation arises in the form of the `texture sticking' effect investigated in StyleGAN3~\cite{aliasfreeKarras2021}. The use of per-frame optimizations, rather than latent-space interpolations, significantly reduces this effect. However, in some cases it is still visible. We hope that as inversion and editing tools for StyleGAN3 emerge, they could be joined with our approach in order to obtain sticking-free results.

Perhaps surprisingly, our framework produces coherent videos without requiring complex machinery designed to directly enforce temporal coherence.
% Perhaps surprisingly, our framework even outperforms more complex methods, which are designed to directly enforce temporal coherence. 
These results indicate that spectral and inductive biases can play a crucial role in maintaining coherency, yielding significant advantages over attempts to brute-force consistency through loss terms.
In addition, we highlight the challenge of stitching an edited crop to the video and propose a designated tuning scheme which can avoid the pitfalls associated with current Poisson-blending approaches.
Looking forward, we hope that our approach can be improved with temporally-aware goals that are meant to supplement it, rather than serve as substitutions. For example, it may be possible to fine-tune the inversion encoder on the input video, to motivate greater consistency in the generated codes. 
% \ron{Moreover, a release of a high-quality video dataset might lead to new directions, utilizing the power of additional supervision.}